\documentclass[10pt,twocolumn,letterpaper]{article}

\usepackage[pagenumbers]{cvpr}
\usepackage{url}
\usepackage{graphicx}

\usepackage{booktabs}       
\usepackage{amsfonts}       
\usepackage{nicefrac}       
\usepackage{xcolor}         

\usepackage{amsmath}
\usepackage{amssymb}
\usepackage{graphicx}
\usepackage{multirow}
\usepackage{bm}
\usepackage{array}
\usepackage{makecell}
\usepackage{adjustbox}
\usepackage[table]{xcolor}
\usepackage{pifont}
\usepackage{algorithm}
\usepackage{algpseudocode}
\usepackage{wrapfig}
\usepackage{cuted}
\usepackage{capt-of}
\usepackage{soul}

\definecolor{inrow}{RGB}{237,244,253}
\definecolor{baserow}{RGB}{248,243,231}

\usepackage{amsmath,amssymb,amsthm}
\theoremstyle{definition}

\theoremstyle{plain}

\definecolor{cvprblue}{rgb}{0.21,0.49,0.74}
\usepackage[pagebackref,breaklinks,colorlinks,allcolors=cvprblue]{hyperref}

\title{Change the Base, Change the Game: Rethinking Pan-sharpening}

\author{Jeonghyeok Do \footnotemark[1]\\
[0.3em]
KAIST\\
{\tt\small ehwjdgur0913@kaist.ac.kr}
\and
Geunhyuk Youk \footnotemark[1]\\
[0.3em]
KAIST\\
{\tt\small rmsgurkjg@kaist.ac.kr}
\and
Munchurl Kim \footnotemark[2]\\
[0.3em]
KAIST\\
{\tt\small mkimee@kaist.ac.kr}
\and
\small{\url{https://kaist-viclab.github.io/PanBase_site}}
}
\begin{document}
\maketitle

\begin{strip}
\centering
    \vspace{-1.6cm}
    \includegraphics[width=\linewidth]{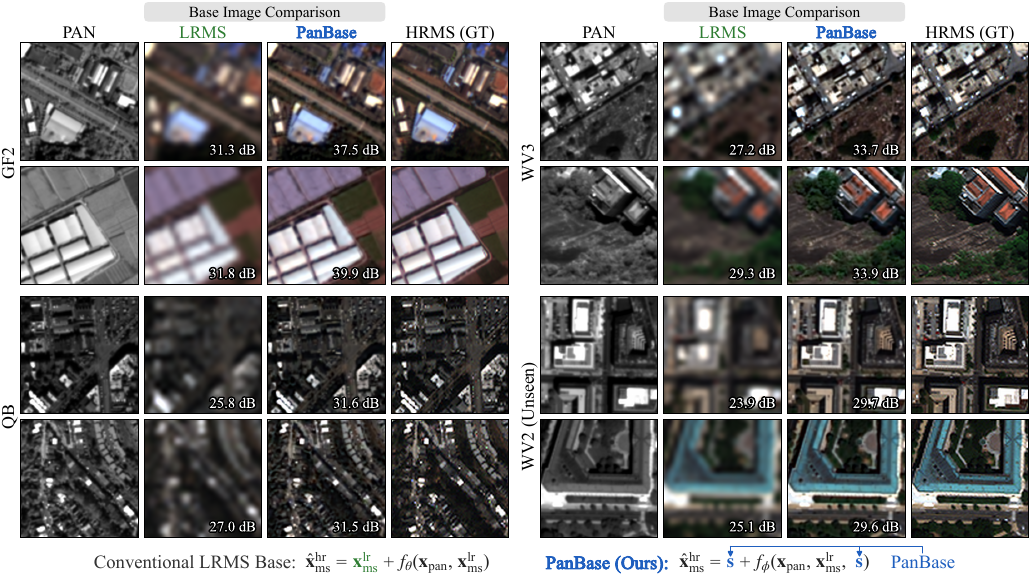}
    \vspace{-0.7cm}
    \captionof{figure}{\textbf{Why start with a blurry base? Same backbone, a sharper base.}
    Examples from PanCollection span GaoFen-2 (GF2), QuickBird (QB),
    WorldView-3 (WV3), and WorldView-2 (WV2).
    The reported PSNR values compare interpolated LRMS and PanBase
    against the HRMS ground truth ($\mathbf{x}_{\mathrm{ms}}^{\mathrm{hr}}$)
    before any neural refinement.
    PanBase replaces the interpolated LRMS in the global skip and also
    serves as an additional backbone input. The final HRMS prediction
    ($\hat{\mathbf{x}}_{\mathrm{ms}}^{\mathrm{hr}}$) is obtained by adding
    the backbone's predicted residual to the base.
    For unseen WV2, PanBase reuses the coefficients calibrated on WV3
    without refitting.}
    \label{fig:teaser}
\end{strip}

{
  \renewcommand{\thefootnote}{\fnsymbol{footnote}}
  \footnotetext[1]{These authors contributed equally.}
  \footnotetext[2]{Corresponding author.}
}

\begin{abstract}
Many pan-sharpening (PS) networks reconstruct high-resolution multispectral (HRMS) images by adding learned residuals to $4\times$-interpolated low-resolution multispectral (LRMS) inputs, incorporating panchromatic (PAN) detail only through these residual corrections.
This raises a simple question: Why start with a blurry base when sharp spatial detail is already available?
We propose \textbf{PanBase}, a sharper multispectral starting point designed for more effective residual learning.
Guided by physical sensor characteristics, PanBase combines PAN spatial detail with LRMS spectral information while enforcing consistency with the LRMS observations under sensor-specific blur.
We construct this base without neural network training, estimating only nine coefficients per spectral band through closed-form calibration.
Our PanBase can be easily integrated into existing backbones with a global residual skip through two simple changes: (i) replacing the skip base with PanBase and (ii) concatenating it as an additional input.
Using each backbone's original training recipe and matched training budgets, PanBase improves most reduced- and full-resolution metrics on WorldView-3, QuickBird, and GaoFen-2, while achieving 4.4$\times$ faster training convergence on average.
In zero-shot evaluation on WorldView-2, standalone PanBase
achieves competitive reconstruction without target-sensor
fitting or neural refinement.
\end{abstract}    
\section{Introduction}
\label{sec:intro}

Residual learning is widely used in image restoration, including
super-resolution~\cite{kim2016vdsr}, denoising~\cite{zhang2017dncnn},
and deblurring~\cite{cho2021mimounet}.
In this approach, a global skip connection carries an initial estimate,
or \textbf{base}, derived from the available observations, such as
low-resolution, noisy, or blurred images.
The network adds its predicted residual to this base to reconstruct the target.
A base that captures more of the target structure leaves less for the
network to recover, which can accelerate convergence and improve
reconstruction quality.

Pan-sharpening (PS) offers a natural opportunity to build a more
informative base because it reconstructs a high-resolution multispectral
(HRMS) image from two complementary observations: a panchromatic (PAN)
image with fine spatial detail and a low-resolution multispectral (LRMS)
image with band-specific spectral information.
(i) Starting from PAN, PS resembles colorization: PAN provides the
spatial structure, and LRMS guides the recovery of the spectral bands.
(ii) Starting from LRMS, PS resembles $4\times$ super-resolution: LRMS
provides the spectral information, and PAN guides the recovery of fine
spatial detail.
In either view, one observation provides the starting point and the
other guides the recovery of what is missing.
This raises a natural question: \textit{Why not build a better base from
both observations before training the residual network?}

Recent PS methods have improved residual prediction through
adaptive convolutions~\cite{duan2024canconv,wang2025arconv},
modality-aware attention~\cite{do2025pancrafter}, state-space
models~\cite{peng2024fusionmamba}, and diffusion~\cite{kim2025uknowdiffpan}.
Yet many residual networks still use interpolated LRMS as the base in
their global skip connection~\cite{yang2017pannet}. Although the residual
branch receives PAN, the base does not directly incorporate its fine
spatial detail. This leaves an opportunity to improve the starting point
itself using detail that is already available in PAN.

In this paper, we propose \textbf{PanBase}, a sharper multispectral base
constructed from PAN and LRMS before backbone training. Guided by the
sensor's physical characteristics, PanBase incorporates spatial detail
from PAN while ensuring that the resulting base reproduces the observed
LRMS after sensor-specific blur and downsampling. Constructing the base
requires no neural network training: closed-form ridge regression~\cite{hoerl1970ridge}
calibrates just nine coefficients per spectral band, giving 36 and 72 coefficients for
four- and eight-band images, respectively. Fig.~\ref{fig:teaser}
shows that PanBase provides a sharper starting point even before neural
refinement. We integrate PanBase with only two changes: (i) replacing interpolated
LRMS with PanBase in the global skip and (ii) concatenating PanBase with
the original inputs.
This design is compatible with diverse backbones that use a global image-space residual connection, including convolution-, attention-, state-space-, and diffusion-based models.

On PanCollection datasets, adding PanBase improves most reduced-resolution (RR)
and full-resolution (FR) metrics on WorldView-3, QuickBird, and GaoFen-2
under matched training budgets. Across all four sensors, including
WorldView-2, it improves RR PSNR in 39 of 40 matched comparisons, with
gains of up to 6.93\,dB. PanBase alone outperforms all compared classical methods in all metrics across the four sensors. On unseen WorldView-2, PanBase reuses the 72 coefficients calibrated on
WorldView-3 and achieves 30.12\,dB PSNR, surpassing all original backbones
in the main comparison without target-sensor fitting or neural refinement.

PanBase also accelerates training convergence by a factor of
4.4$\times$ on average, while keeping the cost of base construction
modest. For four-band data, the one-time calibration takes 88\,s
to load 6,000 training patches and accumulate calibration statistics,
followed by 16\,ms for the coefficient solves. Once the coefficients
and sensor operators are in place, constructing a four-band
256$\times$256 base takes 0.19\,s on CPU.

Our contributions are threefold:
\begin{itemize}
    \item  We rethink residual pan-sharpening by changing its starting point:
    \textbf{PanBase} incorporates the sharp spatial detail already available in PAN
    into the multispectral base before residual learning, while preserving
    consistency with the observed LRMS.
    
    \item We calibrate just nine coefficients per spectral band in closed
    form and integrate PanBase into existing residual backbones by replacing
    the global skip base and concatenating PanBase with the original inputs.
    
    \item Across ten backbones and four sensors, matched comparisons show
    RR PSNR gains of up to 6.93\,dB and faster training, while standalone
    PanBase transfers to an unseen sensor without recalibration.
\end{itemize}
\begin{figure*}[!t]
\centering
\includegraphics[width=\textwidth]{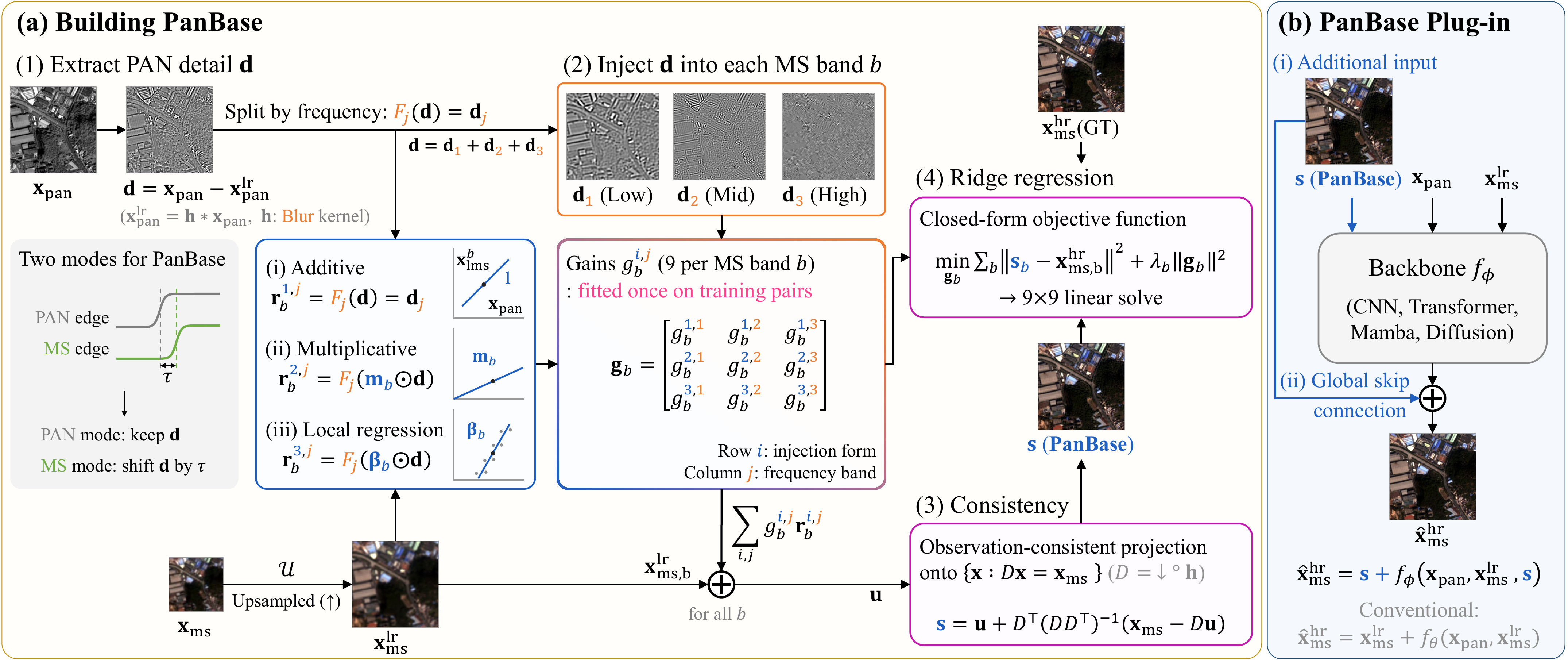}
\vspace{-0.6cm}
\caption{\textbf{PanBase: construct the base, then learn its correction.}
(a) Three PAN-detail forms and three frequency bands yield nine
coefficients per MS band, calibrated by ridge regression and held fixed.
Projection enforces LRMS consistency under the sensor model.
(b) The base replaces the global skip and is concatenated with
the backbone inputs; added input-channel weights are zero-initialized.}
\label{fig:framework}
\end{figure*}

\section{Related Work}
\label{sec:related}

\subsection{Pan-sharpening (PS)}
\label{sec:related_ps}

\noindent\textbf{Traditional methods.}
Traditional PS methods include component substitution (CS), multiresolution analysis (MRA), and variational optimization (VO)~\cite{vivone2015critical,vivone2021newbenchmark}.
CS transfers spatial structure by substituting an MS-derived intensity component with PAN~\cite{lolli2017haze,vivone2019bdsdpc,aiazzi2007gsa,choi2011pracs}.
MRA extracts PAN detail through multiscale decomposition and injects it into upsampled MS using band-dependent gains~\cite{otazu2005awlp,vivone2018fullscale,aiazzi2006mtf,alparone2017intersensor,restaino2016morphological}.
VO reconstructs HRMS by optimizing observation fidelity together with spatial or spectral priors~\cite{palsson2014tv,vivone2015semiblind,palsson2015pwmbf,wu2023lrtcfpan}. While these methods directly reconstruct HRMS, PanBase uses the sensor's physical imaging model to combine PAN spatial detail with LRMS spectral information into an observation-consistent starting point for residual learning.

\noindent\textbf{Learning-based methods.}
Early CNN-based methods, such as PanNet, introduced residual learning for spatial--spectral reconstruction~\cite{yang2017pannet}.
Subsequent designs improve feature aggregation through adaptive convolutions in LAGConv~\cite{jin2022lagconv} and encourage complementary PAN--MS representations through mutual information learning in MutInf~\cite{zhou2022mutinf}.
HFIN extends fusion to hierarchical spatial--frequency representations~\cite{tan2024hfin}, while PAN-Crafter addresses PAN--MS misalignment through modality-aware attention~\cite{do2025pancrafter}.
Recent methods further explore two-step spatial--spectral enhancement in TS-Pan~\cite{li2026tspan} and efficient binary reconstruction in BNNPan~\cite{hou2025bnnpan}.
State-space models also support long-range feature interaction~\cite{peng2024fusionmamba}.
Beyond architectural advances, MDCUN embeds degradation models into deep unfolding~\cite{yang2022mdcun}, while PINO incorporates sensor imaging physics into learned fusion~\cite{liu2025pino}.
PanFlowNet further explores probabilistic reconstruction through conditional normalizing flows~\cite{yang2023panflownet}.
Diffusion-based methods extend generative reconstruction through text conditioning~\cite{xing2024tmdiff}, uncertainty-aware distillation~\cite{kim2025uknowdiffpan}, and few-step sampling~\cite{fang2025pandit}.
PanBase complements these approaches with a sharper, observation-consistent base constructed in closed form and can be integrated into any backbone with a global image-space residual connection.

\subsection{Residual Learning for Image Restoration}
\label{sec:related_residual}

Residual learning preserves an initial estimate and predicts its correction, as in super-resolution~\cite{kim2016vdsr}, denoising~\cite{zhang2017dncnn}, and deblurring~\cite{cho2021mimounet}.
The initial estimate defines the residual target and determines which structures remain for the network to recover.
In PS, residual networks commonly use upsampled LRMS as the base, introducing PAN spatial detail through the learned correction.
PanNet learns spatial corrections in the high-pass domain while preserving spectral information through an upsampled-MS skip~\cite{yang2017pannet}.
DiCNN directly models the injected MS detail~\cite{he2019dicnn}, while CS-Net, MRA-Net, and FusionNet connect classical injection formulations with learned nonlinear detail prediction~\cite{deng2021fusionnet}.
Related efforts also improve initial estimates through PAN-guided learned upsampling in PGCU~\cite{zhu2023pgcu} or use a frozen network-generated spectral base for residual diffusion~\cite{wang2026complex}.
PanBase instead constructs the residual reference through closed-form calibration with explicit LRMS consistency, without requiring a learned upsampler or pre-enhancement network.

\section{PanBase}
\label{sec:method}

PanBase aims to construct a sharper, observation-consistent multispectral
base from PAN and LRMS that serves as the starting point for residual
learning.
Let $\mathbf{x}_{\mathrm{pan}}\in\mathbb{R}^{H\times W}$,
$\mathbf{x}_{\mathrm{ms}}\in\mathbb{R}^{C\times h\times w}$, and
$\mathbf{x}_{\mathrm{ms}}^{\mathrm{hr}}\in\mathbb{R}^{C\times H\times W}$
denote the PAN image, LRMS image, and HRMS target, respectively,
with scale ratio $r=H/h=W/w$.
We write
$\mathbf{x}_{\mathrm{ms}}^{\mathrm{lr}}
=\mathcal{U}(\mathbf{x}_{\mathrm{ms}})$
for upsampled LRMS and
$\mathbf{s}\in\mathbb{R}^{C\times H\times W}$ for PanBase.

Fig.~\ref{fig:framework} summarizes two stages: constructing the base
and learning a residual correction around it.
Starting from $\mathbf{x}_{\mathrm{ms}}^{\mathrm{lr}}$, we inject a
weighted combination of PAN detail components adapted to each spectral
band and frequency range (Sec.~\ref{sec:detail}).
The initial base is then projected to reproduce the observed LRMS under
sensor-specific blur and downsampling (Sec.~\ref{sec:calibration}).
The resulting PanBase serves as both the global residual skip and an
additional backbone input (Sec.~\ref{sec:integration}).
Its coefficients are calibrated once on training pairs and
held fixed during backbone training and inference.

\subsection{PAN Detail Injection}
\label{sec:detail}

We first extract high-frequency detail from the PAN image.
Let $\mathbf{h}_b$ denote the sensor blur kernel for MS band $b$, and
let $\bar{\mathbf{h}}$ denote the mean blur kernel across MS bands.
The blurred PAN image and its detail map are given by
\begin{equation}
\mathbf{x}_{\mathrm{pan}}^{\mathrm{lr}}
=\bar{\mathbf{h}}*\mathbf{x}_{\mathrm{pan}},
\qquad
\mathbf{d}
=\mathbf{x}_{\mathrm{pan}}
-\mathbf{x}_{\mathrm{pan}}^{\mathrm{lr}},
\label{eq:detail}
\end{equation}
where $*$ denotes spatial convolution.
Both quantities remain on the $H\times W$ PAN grid.
We consider two variants, $\mathbf{s}_{\mathrm{pan}}$ and
$\mathbf{s}_{\mathrm{ms}}$, that differ in the spatial alignment of
the PAN detail.
The PAN variant retains $\mathbf{d}$ in its native geometry, whereas
the MS variant shifts it by an input-estimated translation
$\boldsymbol{\tau}$ to align with the MS geometry.
Since all subsequent steps are identical, we use $\mathbf{d}$ for
the detail map of either variant throughout the following formulation.

\noindent\textbf{PAN--MS relation modeling.}
To capture band-dependent PAN--MS relationships, we define two relation
maps: an intensity ratio map $\mathbf{m}_b$ and a local regression map
$\boldsymbol{\beta}_b$~\cite{aiazzi2006mtf}:
\begin{equation}
\mathbf{m}_b
=\frac{\mathbf{x}_{\mathrm{ms},b}^{\mathrm{lr}}}
{\mathbf{x}_{\mathrm{pan}}^{\mathrm{lr}}},
\qquad
\boldsymbol{\beta}_b
=\mathcal{U}\!\left[
\frac{\operatorname{cov}_{w}(\mathbf{x}_{\mathrm{ms},b},\mathbf{p})}
{\operatorname{var}_{w}(\mathbf{p})}
\right],
\label{eq:modulation}
\end{equation}
where
$\mathbf{p}=\downarrow_r\!(\mathbf{x}_{\mathrm{pan}}^{\mathrm{lr}})$
is PAN degraded to the LRMS grid, $w$ is a $7\times7$ local window,
and $\mathcal{U}$ upsamples the regression
map to the HR grid.
The regression map measures the local change in each MS band with
respect to degraded PAN, allowing both positive and negative responses.
Both maps are computed from each PAN--LRMS input pair.

Using these relation maps, we construct three complementary PAN detail
injection forms $\mathbf{r}_b^i$, $i\in\{1,2,3\}$, for each MS band $b$:
\begin{equation}
\mathbf{r}_b^{1}=\mathbf{d}, \qquad
\mathbf{r}_b^{2}=\mathbf{m}_b\odot\mathbf{d}, \qquad
\mathbf{r}_b^{3}=\boldsymbol{\beta}_b\odot\mathbf{d},
\label{eq:families}
\end{equation}
where the three forms correspond to additive, multiplicative, and
local regression injection, respectively, and multiplication $\odot$
and division are performed pointwise.

\noindent\textbf{Frequency-dependent representation.}
Applying a single scalar weight to an injection form scales all its
frequencies equally, limiting independent control over detail at
different spatial scales and noise.
We therefore decompose each $\mathbf{r}_b^i$ into three smooth
frequency bands:
\begin{equation}
\mathbf{r}_b^{i,j}
=F_j(\mathbf{r}_b^i)
=\mathcal{F}^{-1}
\!\left[\alpha_j\odot\mathcal{F}(\mathbf{r}_b^i)\right],
\quad j\in\{1,2,3\},
\label{eq:fft}
\end{equation}
where $F_j$ extracts the $j$-th frequency band,
$\mathcal{F}$ and $\mathcal{F}^{-1}$ denote the Fourier transform and
its inverse, and $\alpha_j$ are fixed, smooth frequency weights for
the low-, middle-, and high-frequency bands, respectively.
The weights $\alpha_j$ sum to one at every Fourier frequency, ensuring
$\sum_{j=1}^{3}\mathbf{r}_b^{i,j}=\mathbf{r}_b^i$.
The three injection forms and three frequency bands yield nine PAN
detail components $\{\mathbf{r}_b^{i,j}\}$ per MS band $b$.

\noindent\textbf{Weighted PAN detail injection.}
We assign a scalar coefficient $g_b^{i,j}$ to each component
$\mathbf{r}_b^{i,j}$, indexed by injection form $i$ and frequency band $j$,
and collect the nine coefficients in the ordered vector
$\mathbf{g}_b=[g_b^{1,1},g_b^{1,2},\ldots,g_b^{3,3}]^\top
\in\mathbb{R}^{9}$.
Adding their weighted combination to upsampled LRMS gives an initial
base parameterized by $\mathbf{g}_b$:
\begin{equation}
\mathbf{u}_b(\mathbf{g}_b)
=\mathbf{x}_{\mathrm{ms},b}^{\mathrm{lr}}
+\sum_{i=1}^{3}\sum_{j=1}^{3}
g_b^{i,j}\mathbf{r}_b^{i,j}.
\label{eq:proposal}
\end{equation}
Stacking the bandwise estimates yields the initial multispectral
base $\mathbf{u}$.

\providecolor{hgood}{HTML}{1B7A3A}
\providecolor{hbad}{HTML}{B3261E}
\providecolor{hgrey}{HTML}{6B6B6B}
\providecommand{\hpos}[1]{\textcolor{hgood}{#1}}
\providecommand{\hneg}[1]{\textcolor{hbad}{#1}}
\providecommand{\hnul}[1]{\textcolor{hgrey}{#1}}

\begin{table*}[!t]
\small
\centering
\sethlcolor{BlueGreen!5}
\caption{\textbf{PanBase as a plug-in for 10 backbones.}
\hl{Highlighted rows} use PanBase $\mathbf{s}_{\mathrm{pan}}$ as the global residual base
and an additional input under matched training.
\textcolor{hgood}{Green}/\textcolor{hbad}{red} deltas indicate gains/drops over each baseline; \textbf{bold}/\underline{underline} denote
the best/second-best PSNR and HQNR.
WV2$^{\dagger}$ uses WV3-trained models without fine-tuning.
Speed-up compares iterations to reach the original best score:
per-backbone medians over reached WV3/QB/GF2 targets,
summarized by their geometric mean.}
\vspace{-0.2cm}
\label{tab:h_top}
\setlength{\tabcolsep}{6pt}
\renewcommand{\arraystretch}{1.1}
\resizebox{\textwidth}{!}{%
\begin{tabular}{ll cccc cccc c}
\toprule
\multirow{2}{*}{\textbf{Method}}
& \multirow{2}{*}{\textbf{Venue}}
& \multicolumn{4}{c}{\textbf{RR PSNR$\uparrow$ (dB)}}
& \multicolumn{4}{c}{\textbf{FR HQNR$\uparrow$}}
& \multirow{2}{*}{\shortstack{\textbf{Training}\\\textbf{speed-up}$\uparrow$}} \\
\cmidrule(lr){3-6}\cmidrule(lr){7-10}
& & WV3 & QB & GF2 & WV2$^{\dagger}$
& WV3 & QB & GF2 & WV2$^{\dagger}$ & \\
\midrule

PanNet~\cite{yang2017pannet} & ICCV'17
& 34.29 & 33.83 & 36.87 & 28.51
& 0.934 & 0.843 & 0.915 & \textbf{0.968}
& -- \\
\rowcolor{BlueGreen!5}
\; + PanBase & --
& 34.96\,{\scriptsize \hpos{+0.67}}
& 34.20\,{\scriptsize \hpos{+0.37}}
& 37.98\,{\scriptsize \hpos{+1.10}}
& 29.10\,{\scriptsize \hpos{+0.59}}
& \underline{0.967}\,{\scriptsize \hpos{+0.034}}
& 0.857\,{\scriptsize \hpos{+0.015}}
& 0.927\,{\scriptsize \hpos{+0.012}}
& 0.958\,{\scriptsize \hneg{$-$0.010}}
& $60.0\times$ \\

LAGConv~\cite{jin2022lagconv} & AAAI'22
& 37.22 & 37.96 & 42.47 & 28.71
& 0.944 & 0.923 & 0.915 & 0.949
& -- \\
\rowcolor{BlueGreen!5}
\; + PanBase & --
& 37.34\,{\scriptsize \hpos{+0.13}}
& 38.11\,{\scriptsize \hpos{+0.15}}
& 42.82\,{\scriptsize \hpos{+0.35}}
& 29.17\,{\scriptsize \hpos{+0.46}}
& 0.958\,{\scriptsize \hpos{+0.014}}
& 0.911\,{\scriptsize \hneg{$-$0.012}}
& 0.949\,{\scriptsize \hpos{+0.034}}
& 0.903\,{\scriptsize \hneg{$-$0.046}}
& $2.1\times$ \\

MutInf~\cite{zhou2022mutinf} & CVPR'22
& 35.77 & 36.64 & 36.65 & 28.87
& 0.913 & 0.876 & 0.943 & 0.902
& -- \\
\rowcolor{BlueGreen!5}
\; + PanBase & --
& 37.05\,{\scriptsize \hpos{+1.28}}
& 38.33\,{\scriptsize \hpos{+1.69}}
& 43.59\,{\scriptsize \hpos{+6.93}}
& 29.01\,{\scriptsize \hpos{+0.14}}
& \underline{0.967}\,{\scriptsize \hpos{+0.055}}
& 0.905\,{\scriptsize \hpos{+0.029}}
& 0.973\,{\scriptsize \hpos{+0.031}}
& 0.940\,{\scriptsize \hpos{+0.038}}
& $9.2\times$ \\

MDCUN~\cite{yang2022mdcun} & CVPR'22
& 36.95 & 35.99 & 42.04 & 28.96
& 0.961 & 0.914 & 0.953 & 0.938
& -- \\
\rowcolor{BlueGreen!5}
\; + PanBase & --
& 37.12\,{\scriptsize \hpos{+0.17}}
& 37.66\,{\scriptsize \hpos{+1.67}}
& 42.57\,{\scriptsize \hpos{+0.53}}
& 30.05\,{\scriptsize \hpos{+1.09}}
& \underline{0.967}\,{\scriptsize \hpos{+0.006}}
& \textbf{0.943}\,{\scriptsize \hpos{+0.029}}
& 0.964\,{\scriptsize \hpos{+0.011}}
& \underline{0.961}\,{\scriptsize \hpos{+0.023}}
& $4.3\times$ \\

PanFlowNet~\cite{yang2023panflownet} & ICCV'23
& 35.76 & 36.23 & 38.56 & 27.78
& 0.918 & 0.853 & 0.904 & 0.869
& -- \\
\rowcolor{BlueGreen!5}
\; + PanBase & --
& 36.53\,{\scriptsize \hpos{+0.76}}
& 37.87\,{\scriptsize \hpos{+1.65}}
& 40.12\,{\scriptsize \hpos{+1.55}}
& 28.19\,{\scriptsize \hpos{+0.41}}
& 0.954\,{\scriptsize \hpos{+0.036}}
& 0.831\,{\scriptsize \hneg{$-$0.022}}
& 0.928\,{\scriptsize \hpos{+0.024}}
& 0.936\,{\scriptsize \hpos{+0.067}}
& $5.0\times$ \\

HFIN~\cite{tan2024hfin} & CVPR'24
& 36.53 & 35.17 & 38.08 & 28.39
& 0.917 & 0.796 & 0.909 & 0.903
& -- \\
\rowcolor{BlueGreen!5}
\quad + PanBase & --
& 37.26\,{\scriptsize \hpos{+0.74}}
& 37.42\,{\scriptsize \hpos{+2.25}}
& 40.95\,{\scriptsize \hpos{+2.87}}
& 29.47\,{\scriptsize \hpos{+1.08}}
& 0.965\,{\scriptsize \hpos{+0.049}}
& 0.924\,{\scriptsize \hpos{+0.128}}
& 0.920\,{\scriptsize \hpos{+0.011}}
& 0.952\,{\scriptsize \hpos{+0.048}}
& $6.1\times$ \\

BNNPan~\cite{hou2025bnnpan} & CVPR'25
& 36.73 & 37.38 & 42.12 & 27.35
& 0.936 & 0.856 & 0.927 & 0.755
& -- \\
\rowcolor{BlueGreen!5}
\; + PanBase & --
& 37.05\,{\scriptsize \hpos{+0.32}}
& 37.71\,{\scriptsize \hpos{+0.33}}
& 42.69\,{\scriptsize \hpos{+0.57}}
& 29.69\,{\scriptsize \hpos{+2.34}}
& \textbf{0.969}\,{\scriptsize \hpos{+0.034}}
& 0.880\,{\scriptsize \hpos{+0.024}}
& 0.942\,{\scriptsize \hpos{+0.015}}
& 0.949\,{\scriptsize \hpos{+0.194}}
& $1.2\times$ \\

PAN-Crafter~\cite{do2025pancrafter} & ICCV'25
& 37.48 & 35.75 & 44.33 & 29.04
& 0.964 & 0.826 & 0.971 & 0.949
& -- \\
\rowcolor{BlueGreen!5}
\; + PanBase & --
& \underline{37.54}\,{\scriptsize \hpos{+0.06}}
& 37.74\,{\scriptsize \hpos{+1.99}}
& \textbf{44.80}\,{\scriptsize \hpos{+0.48}}
& 29.90\,{\scriptsize \hpos{+0.86}}
& 0.962\,{\scriptsize \hneg{$-$0.002}}
& 0.846\,{\scriptsize \hpos{+0.020}}
& \underline{0.979}\,{\scriptsize \hpos{+0.008}}
& 0.937\,{\scriptsize \hneg{$-$0.012}}
& $1.7\times$ \\

PINO~\cite{liu2025pino} & NeurIPS'25
& 36.42 & 37.37 & 42.48 & 29.76
& 0.959 & 0.925 & \underline{0.979} & 0.946
& -- \\
\rowcolor{BlueGreen!5}
\; + PanBase & --
& 37.26\,{\scriptsize \hpos{+0.84}}
& 38.12\,{\scriptsize \hpos{+0.75}}
& 44.20\,{\scriptsize \hpos{+1.72}}
& \textbf{30.14}\,{\scriptsize \hpos{+0.38}}
& 0.961\,{\scriptsize \hpos{+0.002}}
& 0.935\,{\scriptsize \hpos{+0.009}}
& \textbf{0.986}\,{\scriptsize \hpos{+0.007}}
& 0.953\,{\scriptsize \hpos{+0.007}}
& $3.8\times$ \\

TS-Pan~\cite{li2026tspan} & TGRS'26
& \textbf{37.90} & \underline{38.53} & 41.94 & 29.59
& 0.946 & 0.743 & 0.944 & 0.920
& -- \\
\rowcolor{BlueGreen!5}
\; + PanBase & --
& 37.51\,{\scriptsize \hneg{$-$0.40}}
& \textbf{38.54}\,{\scriptsize \hpos{+0.01}}
& \underline{44.34}\,{\scriptsize \hpos{+2.40}}
& \underline{30.10}\,{\scriptsize \hpos{+0.52}}
& 0.955\,{\scriptsize \hpos{+0.010}}
& \underline{0.942}\,{\scriptsize \hpos{+0.198}}
& 0.931\,{\scriptsize \hneg{$-$0.013}}
& 0.930\,{\scriptsize \hpos{+0.010}}
& $2.5\times$\\

\midrule
\multicolumn{2}{l}{Mean change ($\mathbf{s}_{\mathrm{pan}}$)}
& \hpos{+0.46} & \hpos{+1.09} & \hpos{+1.85} & \hpos{+0.79}
& \hpos{+0.024} & \hpos{+0.042} & \hpos{+0.014} & \hpos{+0.032}
& $4.4\times$ \\
\bottomrule
\end{tabular}}
\end{table*}

\subsection{Observation-Consistent Calibration}
\label{sec:calibration}

Throughout this subsection, we vectorize each image band and reuse
the same symbols for the resulting column vectors.
The LRMS observation is $\mathbf{x}_{\mathrm{ms},b}\in\mathbb{R}^{hw}$.
All HR image quantities, including upsampled LRMS
$\mathbf{x}_{\mathrm{ms},b}^{\mathrm{lr}}$, the HRMS target
$\mathbf{x}_{\mathrm{ms},b}^{\mathrm{hr}}$, detail components
$\mathbf{r}_b^{i,j}$, and bases $\mathbf{u}_b$ and $\mathbf{s}_b$,
are vectors in $\mathbb{R}^{HW}$.

\noindent\textbf{Observation-consistent projection.}
Detail injection alone does not guarantee that the initial base
reproduces the observed LRMS after sensor-specific blur and downsampling.
Let $\mathbf{D}_b\in\mathbb{R}^{hw\times HW}$ denote the matrix
representation of the sensor degradation operator
$\downarrow_r\!\circ\,(\mathbf{h}_b*\,\cdot\,)$.
We define the LRMS-consistent set as
\begin{equation}
\mathcal{C}_b
=\left\{
\mathbf{x}\in\mathbb{R}^{HW}
\,\middle|\,
\mathbf{D}_b\mathbf{x}=\mathbf{x}_{\mathrm{ms},b}
\right\}.
\label{eq:consistency_set}
\end{equation}
For full-row-rank $\mathbf{D}_b$, the Euclidean projection onto
$\mathcal{C}_b$ is
\begin{equation}
\begin{aligned}
\Pi_b(\mathbf{u})
&=\arg\min_{\mathbf{x}\in\mathcal{C}_b}
\tfrac{1}{2}\|\mathbf{x}-\mathbf{u}\|_2^2\\
&=\mathbf{u}+\mathbf{D}_b^\dagger
(\mathbf{x}_{\mathrm{ms},b}-\mathbf{D}_b\mathbf{u}),
\end{aligned}
\label{eq:projection}
\end{equation}
where $\mathbf{D}_b^\dagger=\mathbf{D}_b^\top
(\mathbf{D}_b\mathbf{D}_b^\top)^{-1}$.
Applying this projection to the initial base yields PanBase:
\begin{equation}
\mathbf{s}_b(\mathbf{g}_b)
=\Pi_b\!\left(\mathbf{u}_b(\mathbf{g}_b)\right).
\label{eq:panbase}
\end{equation}
By construction, $\mathbf{s}_b(\mathbf{g}_b)\in\mathcal{C}_b$, so
$\mathbf{D}_b\mathbf{s}_b(\mathbf{g}_b)=\mathbf{x}_{\mathrm{ms},b}$
for any coefficient vector $\mathbf{g}_b$, up to numerical solver accuracy.
Thus, calibration adjusts PAN detail while preserving agreement with
the LRMS observation under the prescribed sensor model.

\noindent\textbf{Closed-form coefficient calibration.}
For fixed PAN and LRMS inputs, substituting Eq.~\eqref{eq:proposal}
into Eq.~\eqref{eq:panbase} gives the affine form
\begin{equation}
\mathbf{s}_b(\mathbf{g}_b)
=\mathbf{s}_{0,b}+\mathbf{A}_b\mathbf{g}_b,
\label{eq:nullbase}
\end{equation}
where
$\mathbf{s}_{0,b}=\Pi_b(\mathbf{x}_{\mathrm{ms},b}^{\mathrm{lr}})$
and $\mathbf{A}_b\in\mathbb{R}^{HW\times9}$ contains the nine columns
$(\mathbf{I}_{HW}-\mathbf{D}_b^\dagger\mathbf{D}_b)\mathbf{r}_b^{i,j}$,
ordered consistently with $\mathbf{g}_b$.
For training pair $k\in\{1,\dots,K\}$, define the target correction as
$\mathbf{e}_b^k
=\mathbf{x}_{\mathrm{ms},b}^{\mathrm{hr},k}-\mathbf{s}_{0,b}^k$.
We calibrate the nine coefficients jointly across training pairs
by ridge regression:
\begin{equation}
\mathbf{g}_b^\star
=\arg\min_{\mathbf{g}_b}
\sum_k
\frac{\|\mathbf{A}_b^k\mathbf{g}_b-\mathbf{e}_b^k\|_2^2}{HW}
+\lambda_b\|\mathbf{g}_b\|_2^2.
\label{eq:pb_objective}
\end{equation}
The closed-form solution is $\mathbf{g}_b^\star=(\mathbf{H}_b+\lambda_b\mathbf{I}_9)^{-1}\mathbf{c}_b$, where
\begin{equation}
\mathbf{H}_b
=\sum_k
\frac{(\mathbf{A}_b^k)^\top\mathbf{A}_b^k}{HW},
\quad
\mathbf{c}_b
=\sum_k
\frac{(\mathbf{A}_b^k)^\top\mathbf{e}_b^k}{HW}.
\label{eq:ridge}
\end{equation}
We use $\lambda_b=10^{-2}\operatorname{tr}(\mathbf{H}_b)/9$
and set $\mathbf{g}_b^\star=\mathbf{0}$ when
$\mathbf{H}_b=\mathbf{0}$.
After accumulating the training statistics, coefficient calibration
requires only one 9$\times$9 linear solve per MS band.
This supervised calibration uses HRMS targets without neural network
training and produces $9C$ coefficients that remain fixed during backbone training
and inference.
At inference, PanBase is generated as
$\mathbf{s}_b=\Pi_b(\mathbf{u}_b(\mathbf{g}_b^\star))$
using only PAN, LRMS, the prescribed sensor operators, and
the calibrated coefficients.

\subsection{Residual Backbone Integration}
\label{sec:integration}

We replace the global residual base with PanBase $\mathbf{s}$ and concatenate it
with the original backbone inputs:
\begin{equation}
\widehat{\mathbf{x}}_{\mathrm{ms}}^{\mathrm{hr}}
=\mathbf{s}+f_{\phi}
\left(\mathbf{x}_{\mathrm{pan}},
\mathbf{x}_{\mathrm{ms}}^{\mathrm{lr}},\mathbf{s}\right).
\label{eq:panbase_reconstruction}
\end{equation}
(i) Replacing the interpolated-LRMS skip changes the residual target to
$\mathbf{x}_{\mathrm{ms}}^{\mathrm{hr}}-\mathbf{s}$.
(ii) Supplying $\mathbf{s}$ to the residual branch also exposes the estimate
that the backbone must correct.
We expand only the input layer to accommodate PanBase, keeping
the remaining architecture unchanged.
This integration applies to any backbone with a global image-space
residual connection.
\section{Experiments}
\label{sec:experiment}

\providecolor{BlueGreen}{HTML}{00A88F}
\providecolor{hgood}{HTML}{1B7A3A}
\providecolor{hbad}{HTML}{B3261E}
\providecommand{\hpos}[1]{\textcolor{hgood}{#1}}
\providecommand{\hneg}[1]{\textcolor{hbad}{#1}}

\begin{table*}[!t]
\small
\centering
\caption{\textbf{Classical pan-sharpening versus standalone PanBase.}
All methods operate without neural networks;
PanBase coefficients are calibrated once using HRMS training pairs.
CS/MRA/VO denote component substitution, multiresolution analysis,
and variational optimization.
On WV2, PanBase reuses WV3 coefficients with the WV2 sensor operator.}
\vspace{-0.2cm}
\label{tab:h_classical}
\setlength{\tabcolsep}{6pt}
\renewcommand{\arraystretch}{1.1}
\resizebox{\textwidth}{!}{%
\begin{tabular}{cl ccc ccc ccc ccc}
\toprule
\multirow{2}{*}{\textbf{Type}}
& \multirow{2}{*}{\textbf{Method}}
& \multicolumn{3}{c}{\textbf{WV3}}
& \multicolumn{3}{c}{\textbf{QB}}
& \multicolumn{3}{c}{\textbf{GF2}}
& \multicolumn{3}{c}{\textbf{WV2} (Unseen)} \\
\cmidrule(lr){3-5}
\cmidrule(lr){6-8}
\cmidrule(lr){9-11}
\cmidrule(l){12-14}
& & PSNR$\uparrow$ & SAM$\downarrow$ & HQNR$\uparrow$
  & PSNR$\uparrow$ & SAM$\downarrow$ & HQNR$\uparrow$
  & PSNR$\uparrow$ & SAM$\downarrow$ & HQNR$\uparrow$
  & PSNR$\uparrow$ & SAM$\downarrow$ & HQNR$\uparrow$ \\
\midrule

-- & LRMS$\uparrow$ (interpolated)~\cite{aiazzi2002context}
& 27.12 & 5.80 & 0.909
& 28.06 & 8.44 & 0.831
& 32.08 & 1.82 & 0.905
& 25.38 & 6.53 & 0.922 \\
\midrule

\multirow{4}{*}{CS}
& BT-H~\cite{lolli2017haze}
& 31.22 & 4.92 & 0.891
& 32.09 & 7.23 & 0.685
& 35.80 & 1.56 & 0.832
& 28.90 & 5.90 & 0.879 \\
& BDSD-PC~\cite{vivone2019bdsdpc}
& 30.96 & 5.43 & 0.895
& 31.89 & 8.09 & 0.750
& 35.10 & 1.68 & 0.829
& 29.04 & 6.14 & 0.888 \\
& GSA~\cite{aiazzi2007gsa}
& 30.78 & 5.37 & 0.896
& 31.75 & 8.10 & 0.687
& 34.76 & 1.72 & 0.792
& 28.51 & 6.25 & 0.877 \\
& PRACS~\cite{choi2011pracs}
& 29.95 & 5.58 & 0.935
& 31.00 & 8.19 & 0.781
& 35.32 & 1.69 & 0.865
& 27.24 & 6.32 & 0.941 \\
\midrule

\multirow{4}{*}{MRA}
& AWLP~\cite{otazu2005awlp}
& 31.21 & 5.23 & 0.938
& 31.89 & 8.19 & 0.837
& 34.45 & 1.94 & 0.857
& 28.87 & 6.07 & 0.917 \\
& MTF-GLP-FS~\cite{vivone2018fullscale}
& 30.91 & 5.32 & 0.929
& 32.02 & 7.79 & 0.830
& 35.43 & 1.66 & 0.853
& 28.59 & 6.19 & 0.911 \\
& MTF-GLP-HPM~\cite{aiazzi2006mtf,alparone2017intersensor}
& 31.38 & 5.13 & 0.920
& 32.24 & 7.79 & 0.826
& 34.71 & 1.65 & 0.841
& 29.10 & 6.06 & 0.904 \\
& MF~\cite{restaino2016morphological}
& 30.56 & 5.26 & 0.919
& 31.47 & 7.93 & 0.831
& 34.71 & 1.64 & 0.830
& 28.41 & 6.13 & 0.898 \\
\midrule

\multirow{4}{*}{VO}
& TV~\cite{palsson2014tv}
& 30.48 & 5.69 & \underline{0.952}
& 31.59 & 7.51 & 0.875
& 35.04 & 1.91 & 0.881
& 28.25 & 6.03 & \underline{0.954} \\
& FE-HPM~\cite{vivone2015semiblind}
& 30.87 & 5.23 & 0.923
& 31.71 & 7.95 & 0.826
& 35.39 & 1.61 & 0.822
& 28.71 & 6.10 & 0.902 \\
& PWMBF~\cite{palsson2015pwmbf}
& 30.07 & 5.89 & 0.900
& 31.00 & 8.87 & 0.782
& 33.87 & 1.96 & 0.881
& 27.60 & 7.08 & 0.839 \\
& LRTCFPan~\cite{wu2023lrtcfpan}
& 31.74 & 4.74 & 0.942
& 32.57 & 7.19 & \underline{0.926}
& 37.53 & 1.30 & 0.907
& 29.38 & 5.62 & 0.935 \\
\midrule

\rowcolor{BlueGreen!5}
& \textbf{PanBase} ($\mathbf{s}_{\mathrm{pan}}$)
& \underline{32.28} & \underline{4.65} & \textbf{0.963}
& \underline{33.10} & \underline{6.56} & \textbf{0.944}
& \underline{38.31} & \underline{1.21} & \underline{0.965}
& \underline{30.12} & \underline{5.49} & \textbf{0.959} \\
\rowcolor{BlueGreen!5}\multirow{-2}{*}{\textbf{Ours}} & \textbf{PanBase} ($\mathbf{s}_{\mathrm{ms}}$)
& \textbf{33.88} & \textbf{4.62} & 0.915 & \textbf{36.05} & \textbf{6.11} & 0.703 & \textbf{39.86} & \textbf{1.09} & \textbf{0.971} & \textbf{30.34} & \textbf{5.48} & 0.948 \\
\bottomrule
\end{tabular}}
\end{table*}

\begin{figure*}[!t]
\centering
\includegraphics[width=\textwidth]{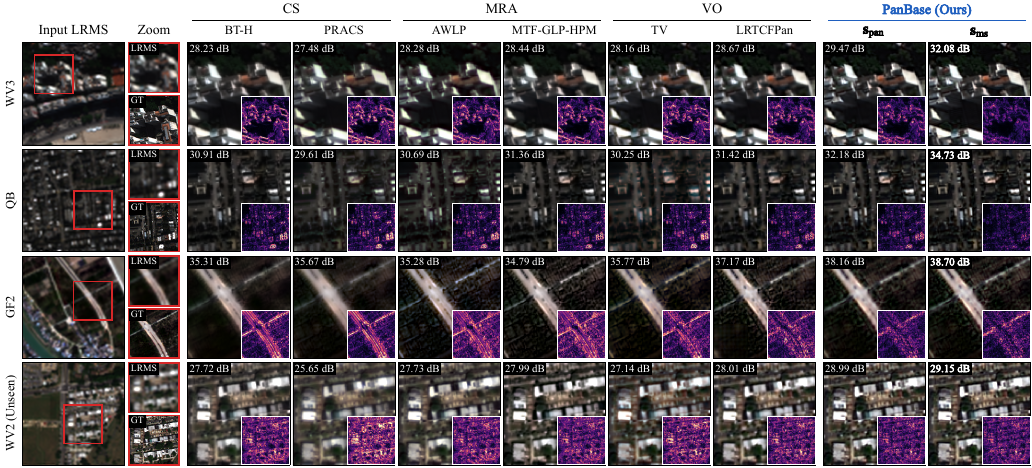}
\vspace{-0.6cm}
\caption{\textbf{Standalone PanBase versus classical fusion at RR.}
PAN- and MS-aligned PanBase are compared with CS/MRA/VO methods
without neural refinement.
\textcolor{red}{Red} boxes mark enlarged regions; scores are PSNR (dB),
and insets show reconstruction errors.
Unseen WV2 reuses WV3-calibrated coefficients.}
\label{fig:classical_rr}
\end{figure*}

\begin{table*}[!t]
\small
\centering
\caption{\textbf{Ablation of alternative residual bases.}
Alternative bases share the same skip replacement,
input concatenation, and training.}
\vspace{-0.2cm}
\label{tab:h_classical_base}
\setlength{\tabcolsep}{6pt}
\renewcommand{\arraystretch}{1.1}
\resizebox{\textwidth}{!}{%
\begin{tabular}{l cc cc cc cc}
\toprule
\multirow{2}{*}{\textbf{Base}}& \multicolumn{2}{c}{\textbf{PanNet}~\cite{yang2017pannet}, \textbf{WV3}}
& \multicolumn{2}{c}{\textbf{PanNet}~\cite{yang2017pannet}, \textbf{GF2}}
& \multicolumn{2}{c}{\textbf{LAGConv}~\cite{jin2022lagconv}, \textbf{WV3}}
& \multicolumn{2}{c}{\textbf{LAGConv}~\cite{jin2022lagconv}, \textbf{GF2}} \\
\cmidrule(lr){2-3}
\cmidrule(lr){4-5}
\cmidrule(lr){6-7}
\cmidrule(l){8-9}
& PSNR$\uparrow$ & HQNR$\uparrow$
& PSNR$\uparrow$ & HQNR$\uparrow$
& PSNR$\uparrow$ & HQNR$\uparrow$
& PSNR$\uparrow$ & HQNR$\uparrow$ \\
\midrule

LRMS$\uparrow$ (Original)
& 34.29 & 0.934
& 36.87 & 0.915
& 37.22 & 0.944
& 42.47 & 0.915 \\
\midrule

$\Pi$(LRMS$\uparrow$)
& 34.75\,{\scriptsize \hpos{+0.46}}
& 0.956\,{\scriptsize \hpos{+0.022}}
& 37.62\,{\scriptsize \hpos{+0.74}}
& 0.915\,{\scriptsize \hnul{-0.000}}
& \textbf{37.35}\,{\scriptsize \hpos{+0.13}}
& 0.956\,{\scriptsize \hpos{+0.012}}
& \textbf{42.90}\,{\scriptsize \hpos{+0.43}}
& 0.918\,{\scriptsize \hpos{+0.004}} \\

GSA~\cite{aiazzi2007gsa} (CS)
& 34.30\,{\scriptsize \hpos{+0.01}}
& 0.915\,{\scriptsize \hneg{$-$0.019}}
& 35.69\,{\scriptsize \hneg{$-$1.19}}
& 0.839\,{\scriptsize \hneg{$-$0.076}}
& 37.14\,{\scriptsize \hneg{$-$0.07}}
& 0.955\,{\scriptsize \hpos{+0.011}}
& 42.01\,{\scriptsize \hneg{$-$0.46}}
& 0.908\,{\scriptsize \hneg{$-$0.006}} \\

MTF-GLP-HPM~\cite{aiazzi2006mtf,alparone2017intersensor} (MRA)
& 34.81\,{\scriptsize \hpos{+0.52}}
& 0.931\,{\scriptsize \hneg{$-$0.003}}
& 36.92\,{\scriptsize \hpos{+0.05}}
& 0.885\,{\scriptsize \hneg{$-$0.031}}
& 37.17\,{\scriptsize \hneg{$-$0.05}}
& 0.944\,{\scriptsize \hnul{-0.000}}
& 42.36\,{\scriptsize \hneg{$-$0.10}}
& 0.921\,{\scriptsize \hpos{+0.007}} \\

LRTCFPan~\cite{wu2023lrtcfpan} (VO)
& 34.81\,{\scriptsize \hpos{+0.52}}
& 0.942\,{\scriptsize \hpos{+0.009}}
& 37.80\,{\scriptsize \hpos{+0.93}}
& 0.904\,{\scriptsize \hneg{$-$0.011}}
& 37.24\,{\scriptsize \hpos{+0.02}}
& 0.956\,{\scriptsize \hpos{+0.013}}
& 42.24\,{\scriptsize \hneg{$-$0.23}}
& 0.948\,{\scriptsize \hpos{+0.034}} \\

\rowcolor{BlueGreen!5}
\textbf{PanBase} ($\mathbf{s}_{\mathrm{pan}}$)
& \textbf{34.96}\,{\scriptsize \hpos{+0.67}}
& \textbf{0.967}\,{\scriptsize \hpos{+0.034}}
& \textbf{37.98}\,{\scriptsize \hpos{+1.10}}
& \textbf{0.927}\,{\scriptsize \hpos{+0.012}}
& 37.34\,{\scriptsize \hpos{+0.13}}
& \textbf{0.958}\,{\scriptsize \hpos{+0.014}}
& 42.82\,{\scriptsize \hpos{+0.35}}
& \textbf{0.949}\,{\scriptsize \hpos{+0.034}} \\

\bottomrule
\end{tabular}}
\end{table*}
\begin{table*}[!t]
\small
\centering
\sethlcolor{BlueGreen!5}
\caption{\textbf{Ablation of PanBase construction.}
Network-free bases add components cumulatively, with coefficients
recalibrated per row and projection applied after detail injection.
Coeff.\ counts calibrated scalars for $C$ MS bands;
the \hl{highlighted row} is the complete $\mathbf{s}_{\mathrm{pan}}$.
PSNR/SAM use RR and HQNR uses FR.
WV2 reuses WV3 coefficients with its own sensor operator.}
\vspace{-0.2cm}
\label{tab:h_construction}
\setlength{\tabcolsep}{3pt}
\renewcommand{\arraystretch}{1.1}
\resizebox{\textwidth}{!}{%
\begin{tabular}{l c ccc ccc ccc ccc}
\toprule
\multirow{2}{*}{\textbf{Configuration}}
& \multirow{2}{*}{\textbf{Coeff.}}
& \multicolumn{3}{c}{\textbf{WV3}}
& \multicolumn{3}{c}{\textbf{QB}}
& \multicolumn{3}{c}{\textbf{GF2}}
& \multicolumn{3}{c}{\textbf{WV2} (Unseen)} \\
\cmidrule(lr){3-5}
\cmidrule(lr){6-8}
\cmidrule(lr){9-11}
\cmidrule(l){12-14}
& & PSNR$\uparrow$ & SAM$\downarrow$ & HQNR$\uparrow$
  & PSNR$\uparrow$ & SAM$\downarrow$ & HQNR$\uparrow$
  & PSNR$\uparrow$ & SAM$\downarrow$ & HQNR$\uparrow$
  & PSNR$\uparrow$ & SAM$\downarrow$ & HQNR$\uparrow$ \\
\midrule

Interpolated LRMS ($\mathbf{x}_{\mathrm{ms},b}^{\mathrm{lr}}$)
& 0
& 27.12 & 5.80 & 0.909
& 28.06 & 8.44 & 0.831
& 32.08 & 1.82 & 0.905
& 25.38 & 6.53 & 0.922 \\
\midrule

+ Projection ($\Pi_b(\cdot)$)
& 0
& 27.99 & 5.44 & 0.927
& 29.09 & 7.86 & 0.822
& 34.21 & 1.47 & 0.951
& 26.20 & 6.05 & 0.901 \\

+ Additive detail ($\mathbf{r}_b^1=\mathbf{d}$)
& $C$
& 31.60 & 5.26 & 0.964
& 32.77 & 7.23 & 0.928
& 37.72 & 1.29 & 0.957
& 29.38 & 5.66 & 0.954 \\

+ Multiplicative detail
($\mathbf{r}_b^2=\mathbf{m}_b\odot\mathbf{d}$)
& $2C$
& 32.07 & 4.87 & 0.963
& 33.07 & 6.59 & 0.931
& 37.84 & 1.24 & 0.960
& 30.01 & 5.53 & 0.956 \\

+ Local regression
($\mathbf{r}_b^3=\boldsymbol{\beta}_b\odot\mathbf{d}$)
& $3C$
& 32.27 & 4.65 & 0.963
& 33.11 & 6.53 & 0.948
& 38.04 & 1.23 & 0.967
& 30.15 & 5.50 & 0.960 \\

\rowcolor{BlueGreen!5}
+ Frequency bands
($\mathbf{r}_b^{i,j}=F_j(\mathbf{r}_b^i)$)
& $9C$
& 32.28 & 4.65 & 0.963
& 33.10 & 6.56 & 0.944
& 38.31 & 1.21 & 0.965
& 30.12 & 5.49 & 0.959 \\

\bottomrule
\end{tabular}}
\end{table*}
\begin{table*}[!t]
\small
\centering
\caption{\textbf{Ablation of PanBase integration.}
With $\mathbf{s}_{\mathrm{pan}}$, we enable skip replacement,
input concatenation, or both under matched training.
Original retains the interpolated-LRMS skip and original inputs.}
\vspace{-0.2cm}
\label{tab:h_skip_input}
\setlength{\tabcolsep}{7pt}
\renewcommand{\arraystretch}{1.1}
\resizebox{\textwidth}{!}{%
\begin{tabular}{ll cccc cccc}
\toprule
\multirow{2}{*}{\textbf{Backbone}}
& \multirow{2}{*}{\textbf{Dataset}}
& \multicolumn{4}{c}{\textbf{RR PSNR$\uparrow$ (dB)}}
& \multicolumn{4}{c}{\textbf{FR HQNR$\uparrow$}} \\
\cmidrule(lr){3-6}\cmidrule(l){7-10}
& & Original & Skip only & Input only & \cellcolor{BlueGreen!5}\textbf{Both (PanBase)}
  & Original & Skip only & Input only & \cellcolor{BlueGreen!5}\textbf{Both (PanBase)} \\
\midrule

\multirow{3}{*}{PanNet~\cite{yang2017pannet}}
& WV3
& 34.29
& 32.56\,{\scriptsize \hneg{$-$1.73}}
& 34.91\,{\scriptsize \hpos{+0.62}}
& 34.96\,{\scriptsize \hpos{+0.67}}
& 0.934
& 0.963\,{\scriptsize \hpos{+0.029}}
& 0.962\,{\scriptsize \hpos{+0.028}}
& 0.967\,{\scriptsize \hpos{+0.034}} \\

& QB
& 33.83
& 33.88\,{\scriptsize \hpos{+0.05}}
& 33.87\,{\scriptsize \hpos{+0.04}}
& 34.20\,{\scriptsize \hpos{+0.37}}
& 0.843
& 0.915\,{\scriptsize \hpos{+0.072}}
& 0.861\,{\scriptsize \hpos{+0.018}}
& 0.857\,{\scriptsize \hpos{+0.015}} \\

& GF2
& 36.87
& 38.11\,{\scriptsize \hpos{+1.24}}
& 37.78\,{\scriptsize \hpos{+0.91}}
& 37.98\,{\scriptsize \hpos{+1.10}}
& 0.915
& 0.976\,{\scriptsize \hpos{+0.061}}
& 0.929\,{\scriptsize \hpos{+0.014}}
& 0.927\,{\scriptsize \hpos{+0.012}} \\
\midrule

\multirow{3}{*}{LAGConv~\cite{jin2022lagconv}}
& WV3
& 37.22
& 37.30\,{\scriptsize \hpos{+0.08}}
& 37.37\,{\scriptsize \hpos{+0.15}}
& 37.34\,{\scriptsize \hpos{+0.13}}
& 0.944
& 0.963\,{\scriptsize \hpos{+0.020}}
& 0.963\,{\scriptsize \hpos{+0.019}}
& 0.958\,{\scriptsize \hpos{+0.014}} \\

& QB
& 37.96
& 38.04\,{\scriptsize \hpos{+0.08}}
& 37.97\,{\scriptsize \hpos{+0.01}}
& 38.11\,{\scriptsize \hpos{+0.15}}
& 0.923
& 0.928\,{\scriptsize \hpos{+0.005}}
& 0.846\,{\scriptsize \hneg{$-$0.077}}
& 0.911\,{\scriptsize \hneg{$-$0.012}} \\

& GF2
& 42.47
& 42.87\,{\scriptsize \hpos{+0.41}}
& 42.81\,{\scriptsize \hpos{+0.35}}
& 42.82\,{\scriptsize \hpos{+0.35}}
& 0.915
& 0.942\,{\scriptsize \hpos{+0.028}}
& 0.939\,{\scriptsize \hpos{+0.025}}
& 0.949\,{\scriptsize \hpos{+0.034}} \\
\midrule

\multirow{3}{*}{PINO~\cite{liu2025pino}}
& WV3
& 36.42
& 37.30\,{\scriptsize \hpos{+0.88}}
& 36.29\,{\scriptsize \hneg{$-$0.13}}
& 37.26\,{\scriptsize \hpos{+0.84}}
& 0.959
& 0.962\,{\scriptsize \hpos{+0.003}}
& 0.957\,{\scriptsize \hneg{$-$0.001}}
& 0.961\,{\scriptsize \hpos{+0.002}} \\

& QB
& 37.37
& 37.21\,{\scriptsize \hneg{$-$0.15}}
& 37.48\,{\scriptsize \hpos{+0.11}}
& 38.12\,{\scriptsize \hpos{+0.75}}
& 0.925
& 0.936\,{\scriptsize \hpos{+0.011}}
& 0.930\,{\scriptsize \hpos{+0.004}}
& 0.935\,{\scriptsize \hpos{+0.009}} \\

& GF2
& 42.48
& 43.71\,{\scriptsize \hpos{+1.23}}
& 42.87\,{\scriptsize \hpos{+0.40}}
& 44.20\,{\scriptsize \hpos{+1.72}}
& 0.979
& 0.981\,{\scriptsize \hpos{+0.002}}
& 0.979\,{\scriptsize 0.000}
& 0.986\,{\scriptsize \hpos{+0.007}} \\
\midrule

\multicolumn{2}{l}{Mean change}
& --
& \hpos{+0.23}
& \hpos{+0.27}
& \hpos{+0.68}
& --
& \hpos{+0.026}
& \hpos{+0.003}
& \hpos{+0.013} \\
\bottomrule
\end{tabular}}
\end{table*}

\begin{figure*}[!t]
\centering
\includegraphics[width=1.0\textwidth]{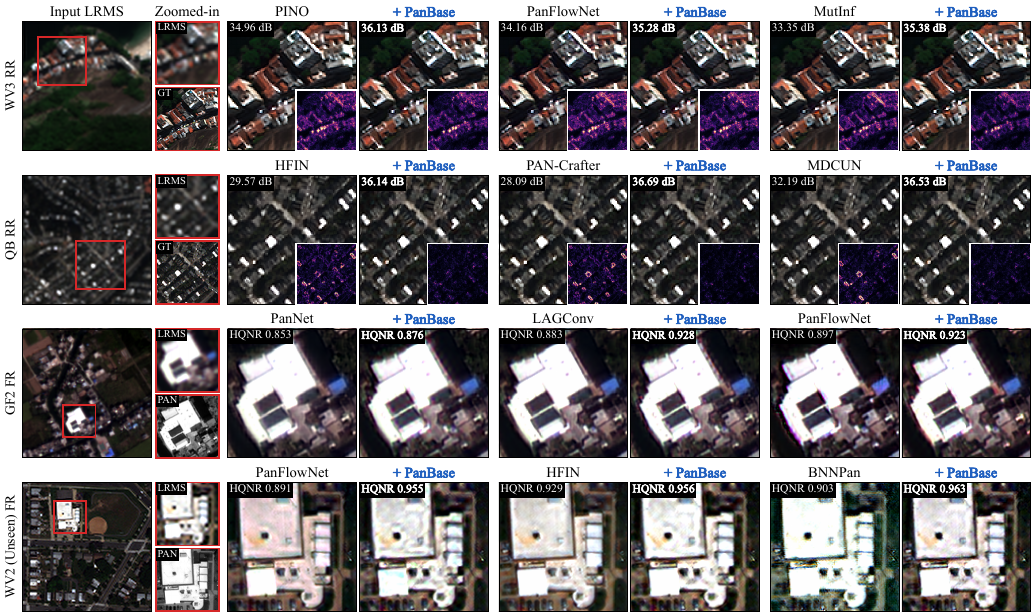}
\vspace{-0.6cm}
\caption{\textbf{Qualitative results before and after PanBase integration.}
Each pair compares an original backbone with its PanBase-augmented
version. WV3/QB rows show RR results with ground-truth references,
PSNR, and error insets; GF2/WV2 rows show FR results with PAN
references and HQNR. Red boxes mark enlarged regions.
Unseen WV2 uses WV3-trained models and coefficients.}
\label{fig:visual_main}
\end{figure*}

\subsection{Datasets}
\label{sec:datasets}

We evaluate PanBase on four pan-sharpening datasets from
PanCollection~\cite{deng2022benchmark}: WorldView-3 (WV3),
QuickBird (QB), GaoFen-2 (GF2), and WorldView-2 (WV2).
For training on WV3, QB, and GF2, we use PAN patches with
$H=W=64$ and LRMS patches with $h=w=16$, giving $r=4$.
The number of MS bands is $C=4$ for QB and GF2 and $C=8$
for WV3 and WV2.
Test PAN images have spatial sizes of 256$\times$256 at
reduced resolution (RR) and 512$\times$512 at full resolution (FR).
We reserve WV2 exclusively for unseen-sensor evaluation,
reusing the WV3-calibrated injection coefficients without
target-sensor recalibration or backbone fine-tuning.

\subsection{Implementation Details}
\label{sec:implementation}

We use each backbone's official implementation and training recipe,
retaining the optimizer, learning rate, batch size, loss, and schedule.
The only changes are replacing the global residual base with PanBase,
concatenating it with the original inputs, and expanding the input layer.

Experiments use up to four NVIDIA RTX A6000 GPUs and an
Intel Core i9-13900K CPU.
For each training dataset, we randomly sample $K=6{,}000$
patches to calibrate the injection coefficients on the GPU.
For four-band (eight-band) data, patch loading and statistics
accumulation take 88\,s (167\,s), followed by
16\,ms (31\,ms) for the ridge solves.
After calibration, generating PanBase on the CPU takes
0.19\,s (0.80\,s) per image at batch size one.

\noindent\textbf{RR evaluation.}
Against HRMS references $\mathbf{x}_{\mathrm{ms}}^{\mathrm{hr}}$,
PSNR measures reconstruction accuracy and SAM measures spectral
angular error; higher PSNR and lower SAM are better.

\noindent\textbf{FR evaluation.}
Without HRMS references, HQNR~\cite{vivone2021newbenchmark} measures
spectral and spatial consistency with observed MS and PAN images;
higher is better.
Additional RR metrics (SSIM~\cite{wang2004ssim}, ERGAS, SCC,
$Q_4/Q_8$~\cite{garzelli2009q2n}) and FR measures ($D_\lambda$, $D_s$)
are reported in the \textit{Appendix}.

\subsection{Comparison with Existing Backbones}
\label{sec:comparison}

The ten backbones in Table~\ref{tab:h_top} cover CNNs (PanNet~\cite{yang2017pannet},
LAGConv~\cite{jin2022lagconv}), mutual-information fusion
(MutInf~\cite{zhou2022mutinf}), spatial--frequency fusion
(HFIN~\cite{tan2024hfin}), attention (PAN-Crafter~\cite{do2025pancrafter}),
binary reconstruction with a diffusion prior (BNNPan~\cite{hou2025bnnpan}),
deep unfolding (MDCUN~\cite{yang2022mdcun}), conditional flows
(PanFlowNet~\cite{yang2023panflownet}), neural operators
(PINO~\cite{liu2025pino}), and two-step mixture-of-experts
(TS-Pan~\cite{li2026tspan}).
Results for other backbones, including state-space models and
additional diffusion networks, are reported in the \textit{Appendix}.

\noindent\textbf{Reduced- and full-resolution reconstruction.}
PanBase improves RR PSNR in 39 of 40 backbone--sensor pairs,
by up to 6.93\,dB, and HQNR in most settings (Table~\ref{tab:h_top}).
These gains support supplying PAN structure before residual learning.
The geometric mean convergence speed-up is $4.4\times$,
measured by iterations to match each original backbone's best score.
Fig.~\ref{fig:visual_main} shows clearer roof boundaries and
building outlines at both RR and FR.

\noindent\textbf{Cross-sensor zero-shot transfer.}
As shown in Table~\ref{tab:h_top}, RR PSNR improves for every
reported backbone on WV2 using WV3-trained models and coefficients
without target-sensor recalibration or fine-tuning.
Standalone MS-aligned PanBase reaches 30.34\,dB, the best PSNR in
Table~\ref{tab:h_classical}, with competitive quality without a
neural network.
FR gains are less uniform, while Fig.~\ref{fig:visual_main}
shows clearer boundaries in the unseen-WV2 examples.

\noindent\textbf{Standalone PanBase versus classical fusion.}
Table~\ref{tab:h_classical} compares standalone PanBase with
CS, MRA, and VO methods without neural networks.
The PAN-aligned $\mathbf{s}_{\mathrm{pan}}$ improves PSNR, SAM,
and HQNR over these methods on all four sensors.
Fig.~\ref{fig:classical_rr} shows clearer boundaries and road structures.
The MS-aligned $\mathbf{s}_{\mathrm{ms}}$ further improves RR
metrics but lowers HQNR on most sensors; its plug-in results
are in the \textit{Appendix}.

\subsection{Base Swap Analysis}
\label{sec:base_swap}

To isolate the effect of the base, we swap in projected LRMS or
classical fused bases under identical integration and training
(Table~\ref{tab:h_classical_base}).
PanBase gives the highest HQNR throughout and the highest PSNR
for PanNet.
Projected LRMS remains competitive for LAGConv at RR,
whereas classical bases can degrade performance.

\subsection{Ablation Studies}
\label{sec:ablations}

\noindent\textbf{PanBase construction.}
Table~\ref{tab:h_construction} isolates cumulative contributions.
Projection improves RR reconstruction; additive PAN detail gives
the largest incremental gain.
Multiplicative and local-regression forms further reduce SAM.
Frequency decomposition mainly benefits GF2 at RR, with small
FR trade-offs. Ablations of the MS-aligned variant $\mathbf{s}_{\mathrm{ms}}$ are reported
in the \textit{Appendix}.

\noindent\textbf{PanBase integration.}
Table~\ref{tab:h_skip_input} separates skip replacement and input
concatenation.
Combining both gives the largest mean RR gain (0.68\,dB) and
improves PSNR throughout, supporting exposure of the base to its
correction branch.
Skip-only integration gives the largest mean FR HQNR gain,
while all three integration strategies improve HQNR on average.
\section{Conclusion}
\label{sec:conclusion}

Why start with a blurry base when sharp PAN detail is
already available?
Our \textbf{PanBase} incorporates this detail into an LRMS-consistent
starting point using nine calibrated gains per band,
then supplies it through the global skip and backbone input.
Across ten backbones and four sensors, it improves RR PSNR
in 39 of 40 matched comparisons and supports faster convergence.
Reusing WV3 gains on unseen WV2 also yields competitive
standalone reconstruction without target-sensor fitting.
These findings highlight base construction as an effective choice for residual PS.

\clearpage

\clearpage
\appendix

\begin{table}[h]
\scriptsize
\centering
\caption{\textbf{Overview of the \textit{Appendix}.}}
\label{tab:supple_overview}
\setlength{\tabcolsep}{5pt}
\renewcommand{\arraystretch}{1.1}
\resizebox{1.0\linewidth}{!}{%
\begin{tabular}{cl}
\toprule
\rowcolor{BlueGreen!5}
\textbf{Section} & \textbf{Contents} \\
\midrule
Sec.~\ref{supp:backbones} & Backbone results and training convergence \\
Sec.~\ref{supp:standalone} & Standalone fusion and transfer to WV2 \\
Sec.~\ref{supp:ablation_ms} & MS-aligned base and integration ablations \\
Sec.~\ref{sec:supp_panbase_construction} & Sensor model, alignment, and projection \\
Sec.~\ref{supp:methods} & Classical methods and backbone families \\
Sec.~\ref{supp:metrics} & Reduced- and full-resolution metrics \\
Sec.~\ref{supp:limitations} & Limitations of the plug-in and standalone base \\
\bottomrule
\end{tabular}%
}
\end{table}

\noindent We evaluate how the choice of residual base affects reconstruction quality and training convergence across 15 pan-sharpening backbones. We also compare the PAN- and MS-aligned bases before neural refinement, including transfer from WV3 to unseen WV2. Table~\ref{tab:supple_overview} lists the experiments, construction details, evaluation protocol, and limitations.

\section{PanBase Across Backbones}
\label{supp:backbones}

\noindent\textbf{Extension to 15 backbones.}
Table~\ref{tab:h_top_pan_additional} extends the ten-backbone comparison in Table~\ref{tab:h_top} to five additional architectures. With the same training recipe as each original backbone, the PAN-aligned base $\mathbf{s}_{\mathrm{pan}}$ improves reduced-resolution (RR) PSNR in 17 of the 20 additional backbone--sensor pairs and in 56 of 60 pairs overall.

\noindent\textbf{PAN versus MS geometry.}
Table~\ref{tab:h_top_ms} compares the two geometries on the 53 pairs with results for both. The MS-aligned base $\mathbf{s}_{\mathrm{ms}}$ gives larger mean PSNR gains on each training sensor. The largest difference occurs on GF2, where the mean gain rises from 1.82 to 2.26\,dB. On unseen WV2 the PAN- and MS-aligned bases gain 0.54 and 0.36\,dB, respectively. For WV2 we reuse the WV3-trained backbone and the WV3 calibration coefficients and construct the base with the WV2 sensor operator.

\begin{figure*}[!t]
\centering
\includegraphics[width=0.95\textwidth]{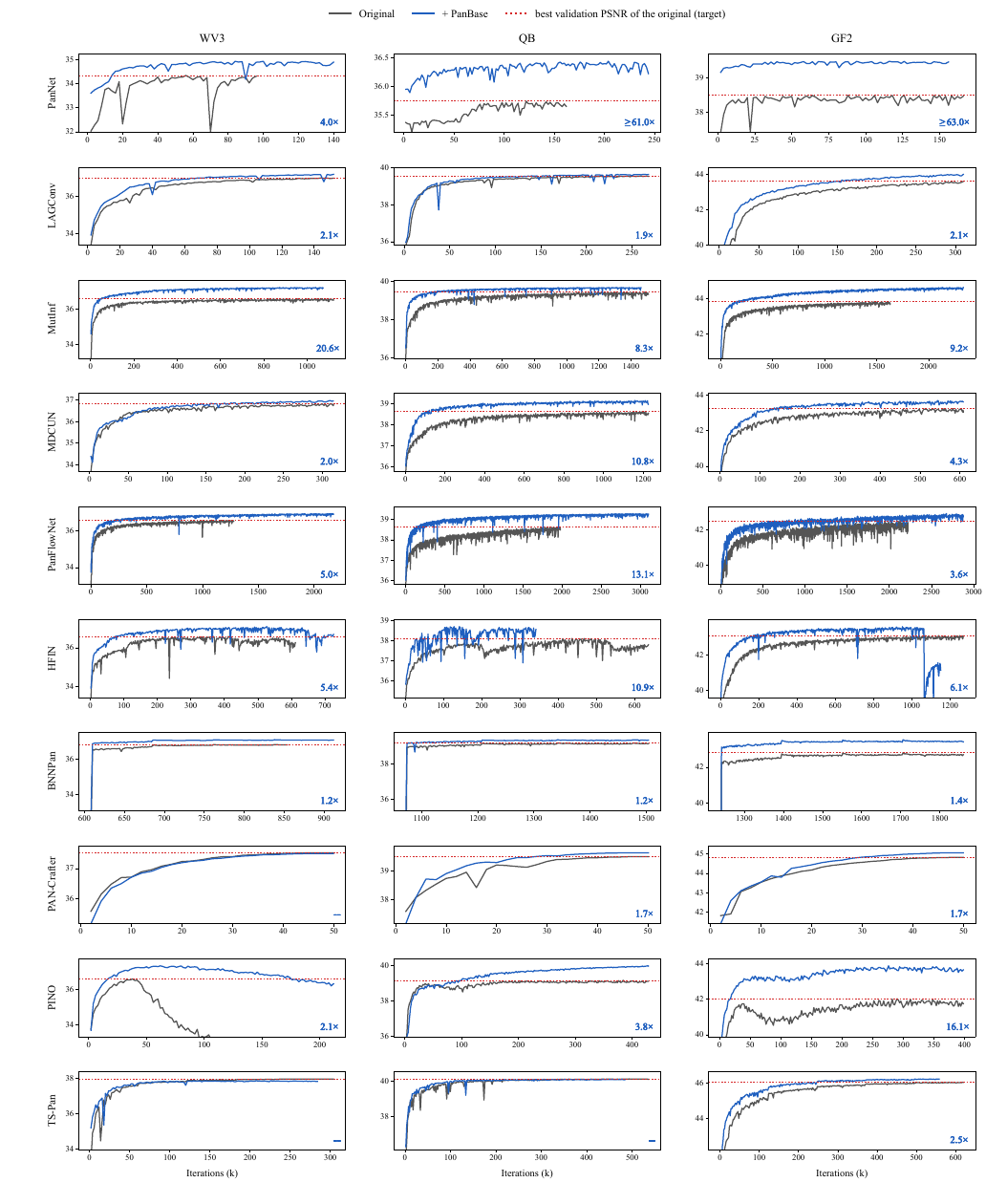}
\caption{\textbf{Validation PSNR over training iterations.} Each panel shows one backbone trained with its original base and with PanBase. The dotted horizontal line marks the best validation PSNR of the original run. Annotations give the iteration ratio at that target. A dash marks an unreached target. For PanNet, the QB and GF2 panels show lower bounds. We use $60\times$ for PanNet when computing the tabulated geometric means.}
\label{fig:supp_convergence}
\end{figure*}

\noindent\textbf{Training convergence.}
Fig.~\ref{fig:supp_convergence} plots validation PSNR for the ten backbones. For each training sensor, the target is the best validation PSNR of the original run. We divide the iteration at which the original run first reaches the target by the corresponding iteration for the PanBase run, take the median over reached sensor targets for each backbone, and report the geometric mean across backbones. Across the 14 backbones with convergence results for both geometries, the geometric means are $4.0\times$ for $\mathbf{s}_{\mathrm{pan}}$ and $4.5\times$ for $\mathbf{s}_{\mathrm{ms}}$.

\subsection{Results by Sensor}
\label{supp:sensor_results}

\begin{figure*}[!t]
\centering
\includegraphics[width=\textwidth]{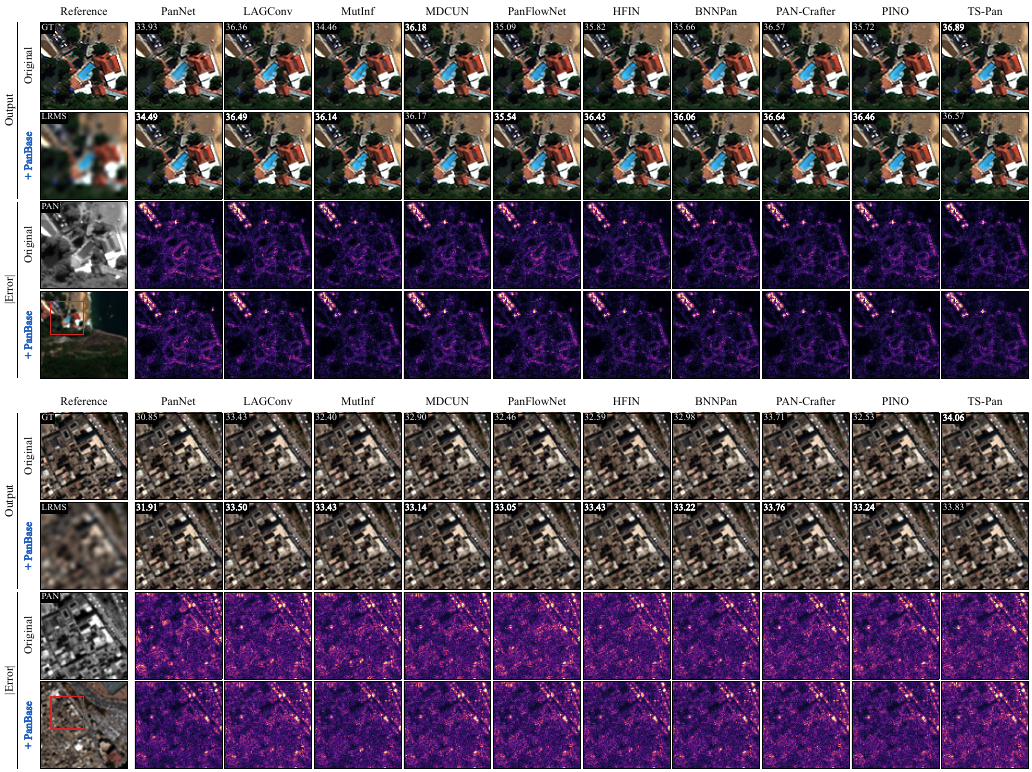}
\caption{\textbf{Backbone reconstructions on WV3 at reduced resolution.} Each column pairs the original and PanBase outputs with their absolute error maps for two scenes. Scores are PSNR in dB. The reference column identifies the crop and shows the HRMS, interpolated LRMS, and PAN images.}
\label{fig:supp_rr_wv3}
\end{figure*}

\begin{figure*}[!t]
\centering
\includegraphics[width=\textwidth]{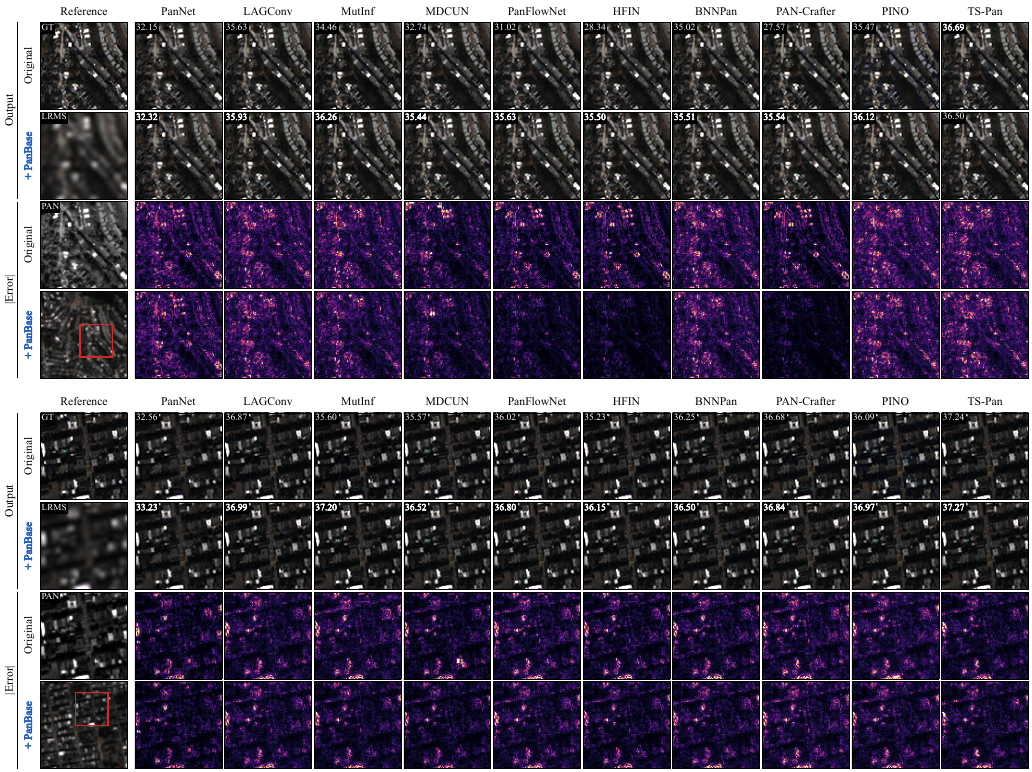}
\caption{\textbf{Backbone reconstructions on QB at reduced resolution.} Each column pairs the original and PanBase outputs with their absolute error maps for two scenes. Scores are PSNR in dB. The HRMS column marks the crop used for comparison.}
\label{fig:supp_rr_qb}
\end{figure*}

\begin{figure*}[!t]
\centering
\includegraphics[width=\textwidth]{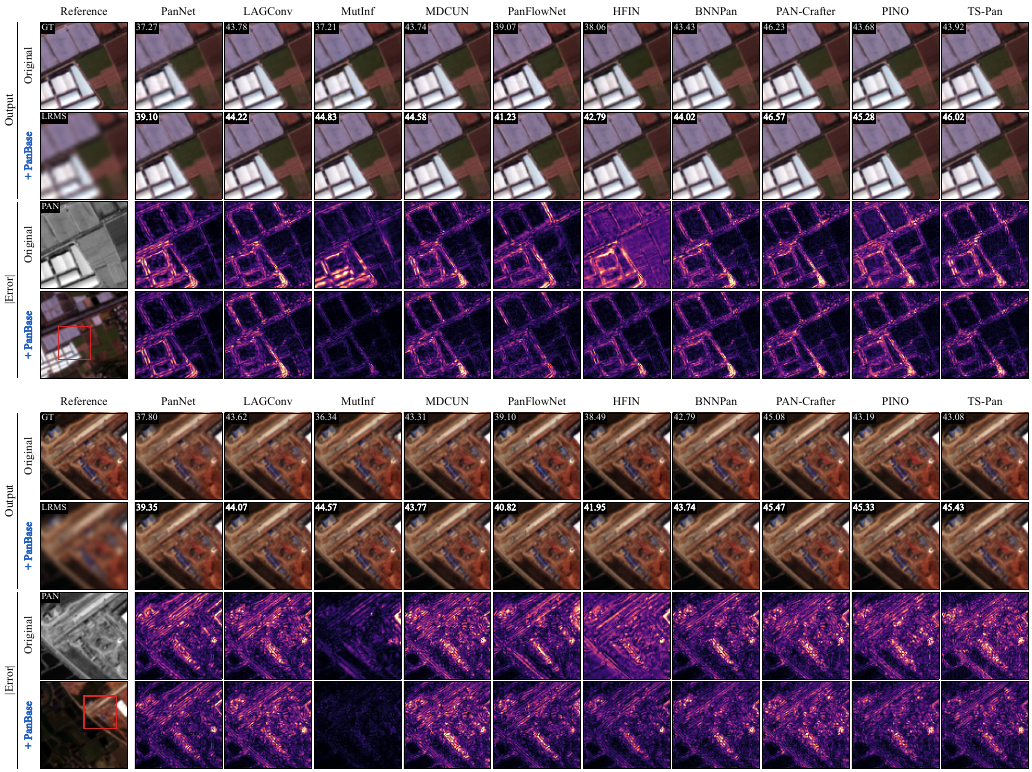}
\caption{\textbf{Backbone reconstructions on GF2 at reduced resolution.} Each column pairs the original and PanBase outputs with their absolute error maps for two scenes. Scores are PSNR in dB. In the upper scene, MutInf improves from 37.21 to 44.83\,dB with smaller errors around the bright roofs.}
\label{fig:supp_rr_gf2}
\end{figure*}

\begin{figure*}[!t]
\centering
\includegraphics[width=\textwidth]{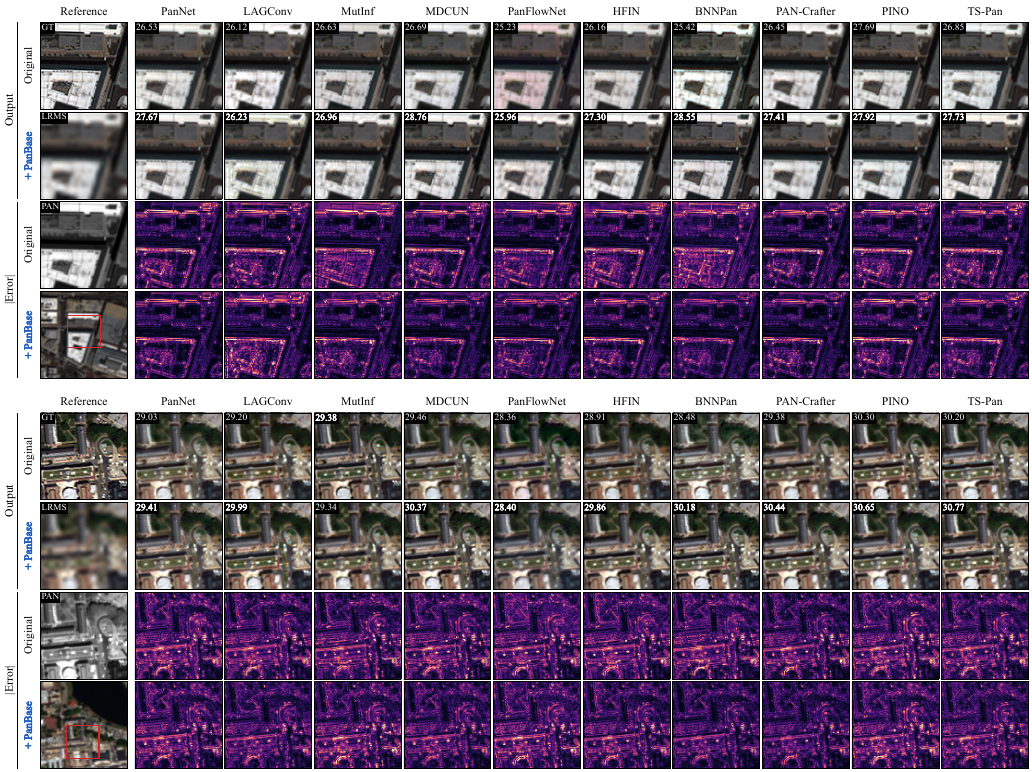}
\caption{\textbf{Backbone reconstructions on unseen WV2 at reduced resolution.} Each column pairs the original and PanBase outputs with their absolute error maps for two scenes. Scores are PSNR in dB. The models and calibration coefficients come from WV3. The upper scene shows BNNPan improving from 25.42 to 28.55\,dB.}
\label{fig:supp_rr_wv2}
\end{figure*}

Tables~\ref{tab:panbase_full_wv3}, \ref{tab:panbase_full_qb}, \ref{tab:panbase_full_gf2}, and~\ref{tab:panbase_full_wv2} report all eight RR and three FR metrics. On GF2, all 15 backbones gain PSNR with $\mathbf{s}_{\mathrm{pan}}$, as do all 12 with $\mathbf{s}_{\mathrm{ms}}$ results. On WV2, 14 of 15 improve with $\mathbf{s}_{\mathrm{pan}}$. We also include the two standalone bases in the WV2 table. Without a learned residual, $\mathbf{s}_{\mathrm{ms}}$ reaches 30.34\,dB and obtains the highest SSIM (0.8747) among the entries shown.
Figs.~\ref{fig:supp_rr_wv3}, \ref{fig:supp_rr_qb}, \ref{fig:supp_rr_gf2}, and~\ref{fig:supp_rr_wv2} pair each reconstruction with its absolute error against HRMS.

\section{Standalone PanBase and Cross-Sensor Transfer}
\label{supp:standalone}

\begin{figure*}[!t]
\centering
\includegraphics[width=\textwidth]{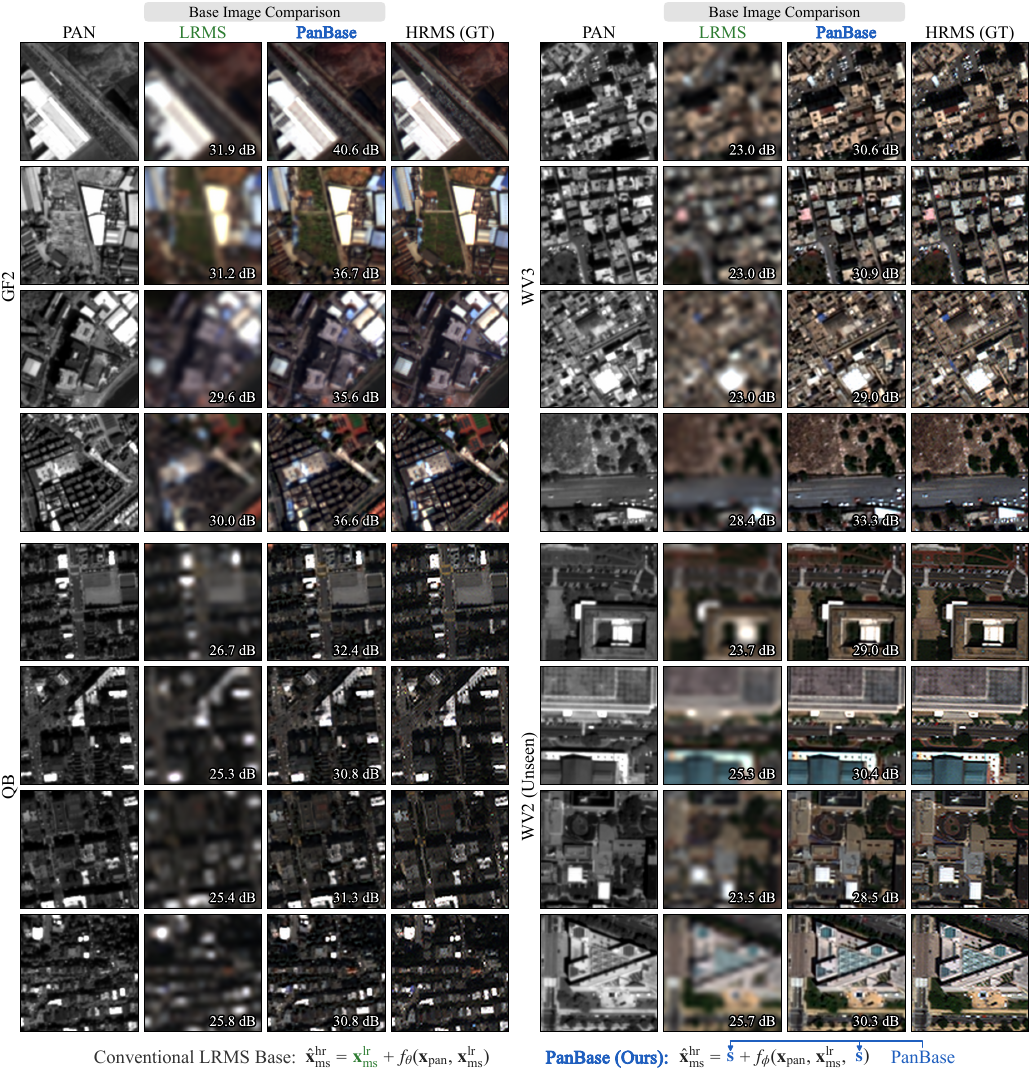}
\caption{\textbf{Residual bases before neural refinement.} Each crop shows the observed PAN, interpolated LRMS, PanBase, and reference HRMS. The scores compare the two bases against HRMS in dB. The examples cover GF2, QB, WV3, and unseen WV2.}
\label{fig:teaser_4}
\end{figure*}

Fig.~\ref{fig:teaser_4} isolates the choice of residual base before any backbone is applied. Interpolated LRMS preserves coarse color but smooths the bright roof boundaries in the first GF2 scene. Injecting PAN detail raises the PSNR of that base from 31.9 to 40.6\,dB.

\begin{figure*}[!t]
\centering
\includegraphics[width=\textwidth]{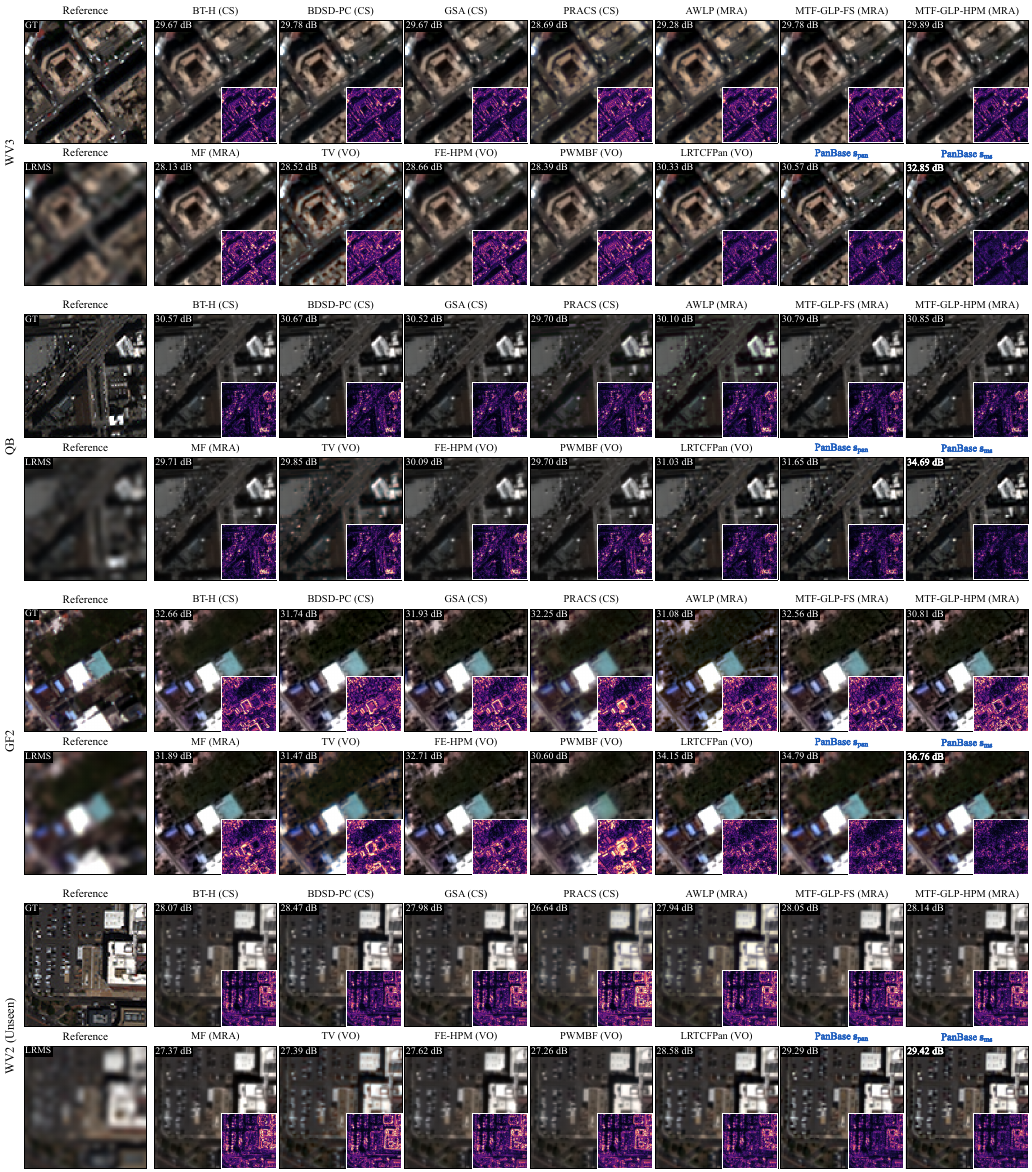}
\caption{\textbf{Standalone reconstruction on four reduced-resolution scenes.} Each sensor panel compares the two PanBase geometries with 12 classical methods. The HRMS and interpolated LRMS provide references. Insets show reconstruction errors, and scores give per-scene PSNR in dB.}
\label{fig:classic_2}
\end{figure*}

\begin{figure*}[!t]
\centering
\includegraphics[width=\textwidth]{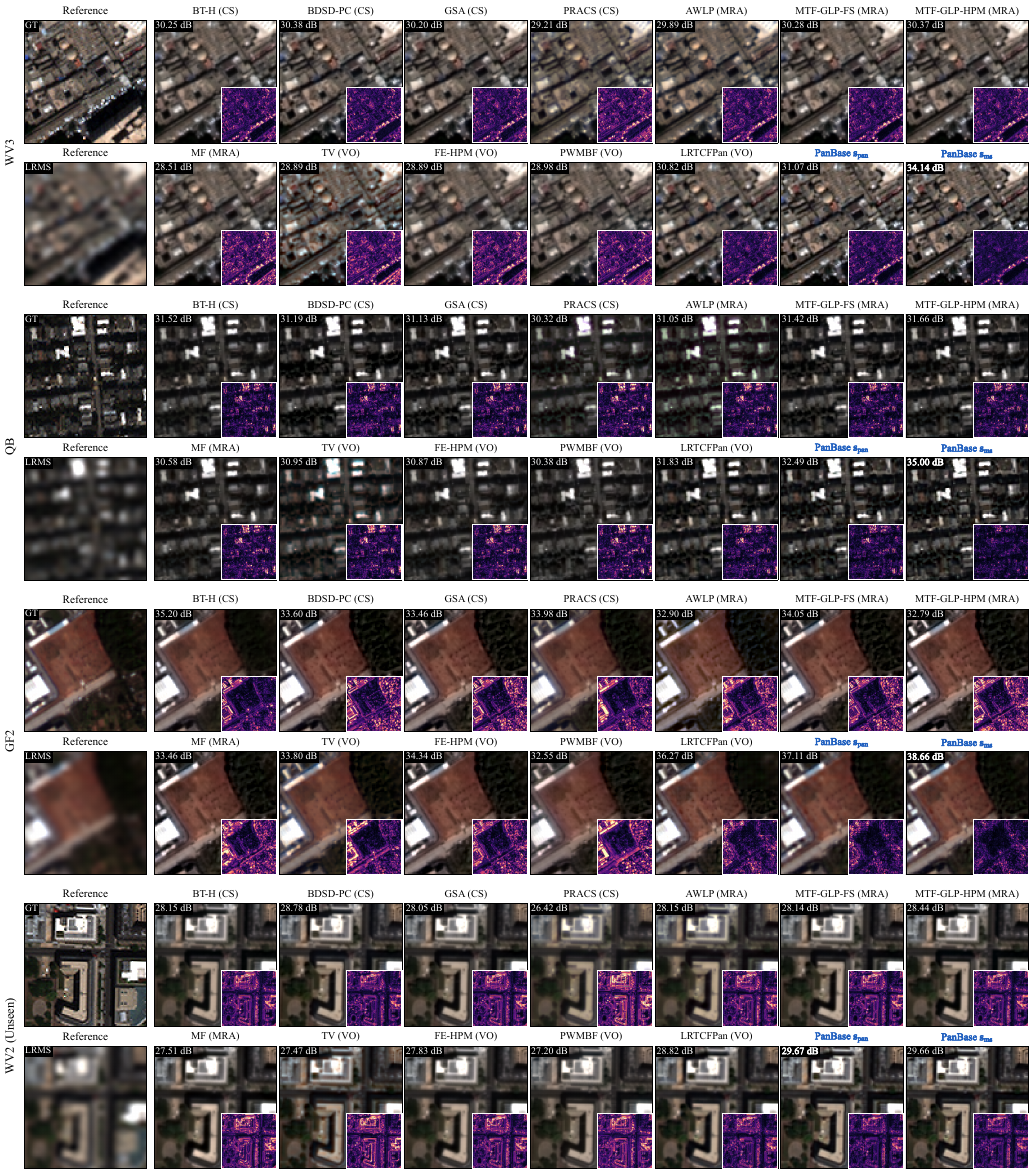}
\caption{\textbf{Additional scenes for standalone fusion.} In the GF2 panel, PAN- and MS-aligned PanBase reach 37.11 and 38.66\,dB. In the unseen WV2 panel, they reach 29.67 and 29.66\,dB using coefficients calibrated on WV3. Insets show the corresponding errors against HRMS.}
\label{fig:classic_3}
\end{figure*}

Tables~\ref{tab:h_classical_wv3}, \ref{tab:h_classical_qb}, \ref{tab:h_classical_gf2}, and~\ref{tab:h_classical_wv2} compare the standalone bases with 12 classical methods. The MS-aligned base has the highest RR PSNR and lowest spectral angle on all four sensors. The PAN-aligned base has higher FR HQNR than every classical method on all four. On unseen WV2, PAN- and MS-aligned PanBase reach 30.12 and 30.34\,dB, respectively. Figs.~\ref{fig:classic_2} and~\ref{fig:classic_3} show the variation across scenes.
\section{Ablations in MS Geometry}
\label{supp:ablation_ms}

\noindent\textbf{Construction.}
Table~\ref{tab:h_construction_ms} adds the detail components cumulatively and refits the coefficients at each step. Adding the first, additive PAN detail term to projected LRMS gives the largest PSNR increase on every sensor. On QB this step raises PSNR from 29.09 to 35.30\,dB. The multiplicative and regression terms, followed by frequency-dependent weighting, bring it to 36.05\,dB. The last step has its largest effect on GF2, increasing PSNR from 39.59 to 39.86\,dB.

\noindent\textbf{Backbone integration.}
Table~\ref{tab:h_skip_input_ms} separates the two ways the backbone uses the base. Across PanNet and PINO on three training sensors, changing only the global skip gains 1.43\,dB on average. Supplying the base only as an input gains 1.25\,dB, and using it in both places gains 1.78\,dB. Input concatenation alone gives the largest mean HQNR gain (0.029), followed by using both (0.023). Table~\ref{tab:h_classical_base_ms} holds the integration fixed and varies the base. For PanNet on GF2, projected LRMS yields 37.62\,dB and 0.915 HQNR, while the complete MS-aligned base yields 40.06\,dB and 0.963.

\section{PanBase Construction Details}
\label{sec:supp_panbase_construction}

\noindent\textbf{Sensor model.}
For each MS band $b$, we construct a $41\times41$ MTF-matched kernel $\mathbf{h}_b$ from a Gaussian frequency response with gain $G_b$ at the MS Nyquist frequency, using a Kaiser window ($\beta=0.5$)~\cite{aiazzi2006mtf}.
We use the PanCollection gains~\cite{deng2022benchmark} for WV3 $(0.325,0.355,0.360,0.350,0.365,0.360,0.335,0.315)$, QB $(0.34,0.32,0.30,0.22)$, and WV2 ($0.35$ for the first seven bands and $0.27$ for NIR2). For GF2, we set $G_b=0.30$ for every band.
We use the same filter design for the PAN smoothing kernel $\bar{\mathbf{h}}$, with the mean gain $\bar G=C^{-1}\sum_b G_b$.
The degradation operator $\mathbf{D}_b$ applies $\mathbf{h}_b$ with replicate padding and samples every fourth pixel with a two-pixel offset along each axis.

\begin{figure}[H]
\centering
\includegraphics[width=\linewidth]{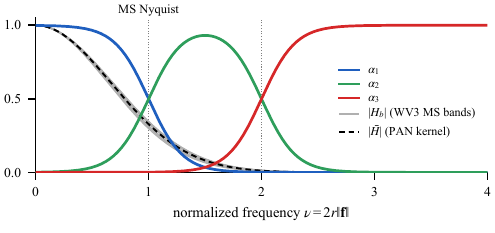}
\caption{\textbf{Frequency weights for PAN detail injection.} The three weights overlap near $\nu=1$ and $2$ and sum to one. Gray curves show the WV3 MS filter responses and the dashed curve shows the PAN smoothing filter. Along the plotted axis, $\nu=1$ and $4$ are the MS and PAN Nyquist frequencies.}
\label{fig:panbase_frequency_bands}
\end{figure}

\noindent\textbf{Frequency decomposition.}
Let $\nu=2r\|\mathbf f\|_2$, where $r=4$ and $\mathbf f$ is measured in cycles per PAN pixel.
With $a_k(\nu)=\sigma((\nu-k)/0.15)$ and logistic function $\sigma$, the low-, middle-, and high-frequency weights are
\begin{equation}
(\alpha_1,\alpha_2,\alpha_3)=(1-a_1,\ a_1-a_2,\ a_2).
\end{equation}
We apply these weights to each injection component after forming the pointwise products, using reflection padding of $\max(4,\lfloor\min(H,W)/4\rfloor)$ pixels before the FFT and cropping after the inverse FFT (Fig.~\ref{fig:panbase_frequency_bands}).

\noindent\textbf{Injection maps and MS alignment.}
In Eq.~\eqref{eq:modulation}, we floor the denominator of the intensity ratio at $10^{-3}\operatorname{mean}(|\mathbf{x}_{\mathrm{pan}}^{\mathrm{lr}}|)$. For local regression, we use a replicate-padded $7\times7$ window on the LRMS grid, floor the local PAN variance at $10^{-6}\operatorname{var}(\mathbf p)$, clip the slope to $[-2,4]$, and upsample it bilinearly.
For $\mathbf{s}_{\mathrm{ms}}$, we fit $\sum_b w_b\mathbf{x}^{\mathrm{lr}}_{\mathrm{ms},b}+c$ to the low-pass PAN image by least squares, using the 23-tap LRMS interpolation supplied by PanCollection.
We estimate one translation per image by Hann-windowed phase correlation over $\nu<1.5$, with translation magnitude bounded by $\min(32,\min(H,W)/4)$ pixels. We refine the shift by parabolic interpolation and up to ten Gauss--Newton steps, stopping when the update is shorter than $10^{-3}$ pixels. We apply a Fourier phase ramp on a reflect-extended grid to the PAN detail and the PAN terms in $\mathbf{m}_b$ and $\boldsymbol{\beta}_b$.

\noindent\textbf{Projection.}
We solve each bandwise system involving $\mathbf{D}_b\mathbf{D}_b^\top$ by conjugate gradients with the exact adjoint $\mathbf{D}_b^\top$. The preconditioner uses a separable approximation to $\mathbf h_b$ formed from its marginals. We stop when the squared residual falls below $10^{-12}$ of its initial value or after 200 iterations.
Before clipping, the relative LRMS residual $\|\mathbf{D}_b \mathbf{s}_b-\mathbf{x}_{\mathrm{ms},b}\|_2/\|\mathbf{x}_{\mathrm{ms},b}\|_2$ is approximately $4\times10^{-7}$ in float32.
We clip the base to $[0,2047]$ for WV3, QB, and WV2 and to $[0,1023]$ for GF2. We fit the $9C$ coefficients once per training sensor on 6,000 training patches. Both geometries use their respective fixed coefficients throughout backbone training and testing.

\section{Comparison Methods}
\label{supp:methods}

We compare standalone PanBase with three families of classical fusion methods~\cite{vivone2015critical,vivone2021newbenchmark}. Component substitution transfers PAN detail through an MS-derived intensity component. Multiresolution analysis extracts PAN detail across scales and injects it into the MS bands. Variational optimization reconstructs the HRMS image from the observations and a regularization term. Interpolated LRMS provides a common reference for these comparisons.

Table~\ref{tab:supp_backbone_families} groups the 15 learned backbones by the network used for reconstruction. BNNPan uses a binary convolutional backbone with an auxiliary diffusion prior. For the four backbones that sample at inference (PanFlowNet, BNNPan, U-Know-DiffPAN, and PanDiT), we report one sample per test image with a fixed random seed. U-Know-DiffPAN and PanDiT use 25 DDIM steps in our evaluation.

\begin{table*}[!t]
\centering
\caption{\textbf{Backbones evaluated with PanBase.} Methods are grouped by their reconstruction architecture.}
\label{tab:supp_backbone_families}
\small
\setlength{\tabcolsep}{7pt}
\renewcommand{\arraystretch}{1.1}
\resizebox{0.9\textwidth}{!}{%
\begin{tabular}{lll}
\toprule
\textbf{Family} & \textbf{Backbone} & \textbf{Main components} \\
\midrule
\multirow{10}{*}{CNN} & PanNet~\cite{yang2017pannet} & High-pass residual network with an upsampled-MS skip \\
& LAGConv~\cite{jin2022lagconv} & Residual network with locally adaptive convolutions \\
& MutInf~\cite{zhou2022mutinf} & Edge-aware PAN and MS features fused by invertible coupling blocks \\
& MDCUN~\cite{yang2022mdcun} & Six-stage deep unfolding with memory and convolutional priors \\
& PanFlowNet~\cite{yang2023panflownet} & Conditional normalizing flow with convolutional coupling blocks \\
& MSDDN~\cite{he2023msddn} & Invertible blocks for multiscale spatial and Fourier-domain fusion \\
& HFIN~\cite{tan2024hfin} & Hierarchical fusion of spatial, Fourier, and short-time Fourier features \\
& UTeRM~\cite{mai2024uterm} & Ten-stage tensor-rank minimization with convolutional proximal operators \\
& BNNPan~\cite{hou2025bnnpan} & Binary wavelet CNN with a latent diffusion prior \\
& PAN-Crafter~\cite{do2025pancrafter} & Convolutional U-Net with local attention for PAN--MS alignment \\
\midrule
\multirow{2}{*}{Transformer} & PINO~\cite{liu2025pino} & Window and channel attention encoder with a Galerkin-attention decoder \\
& TS-Pan~\cite{li2026tspan} & Convolutional experts followed by attention and RWKV refinement \\
\midrule
State-space & MMMamba~\cite{wang2026mmmamba} & Selective scans over interleaved MS and PAN windows \\
\midrule
\multirow{2}{*}{Diffusion} & U-Know-DiffPAN~\cite{kim2025uknowdiffpan} & U-Net student distilled from a diffusion teacher with uncertainty weighting \\
& PanDiT~\cite{fang2025pandit} & Diffusion Transformer with DDIM sampling \\
\bottomrule
\end{tabular}}
\end{table*}

\section{Evaluation Metrics}
\label{supp:metrics}

We compute each metric per image and average over the 20 test images of each sensor and resolution, following PanCollection~\cite{deng2022benchmark}.

\noindent\textbf{Reduced resolution.} We compare the fused image with its HRMS reference on the same $197\times197$ interior crop of each $256\times256$ image. We divide digital numbers by 2,047 on WV3, QB, and WV2 and by 1,023 on GF2. PSNR uses a peak value of one and the mean squared error over all pixels and bands. SSIM~\cite{wang2004ssim} is averaged across bands using an $11\times11$ Gaussian window ($\sigma=1.5$). SAM is the mean pixelwise spectral angle in degrees. ERGAS is $(100/r)\sqrt{C^{-1}\sum_{b=1}^C(\mathrm{RMSE}_b/\mu_b)^2}$, where $\mu_b$ is the mean of reference band $b$ and $r=4$. SCC averages the bandwise correlations between the fused and reference images after a $3\times3$ Laplacian filter. We compute $Q2^n$~\cite{garzelli2009q2n} on the original digital numbers in non-overlapping $32\times32$ blocks, using $Q_4$ for four bands and $Q_8$ for eight.

For LPIPS and DISTS, we form RGB composites from bands 5, 3, and 2 on WV3 and WV2, and bands 3, 2, and 1 on QB and GF2 (one-based indexing). The 1st and 99th percentiles of each reference band define a linear stretch shared by the reference and all reconstructions of that image. We use the \texttt{pyiqa} implementation for both distances.

\noindent\textbf{Full resolution.} We evaluate the entire $512\times512$ fused image with the benchmark toolbox~\cite{vivone2021protocol,vivone2021newbenchmark}. Spectral distortion $D_\lambda$ is one minus $Q2^n$ between the observed LRMS and the fused image after MS MTF filtering and $4\times$ decimation. For this evaluation we use the QB and WV3 MTF presets on their respective sensors, the WV3 preset on WV2, and a Nyquist gain of 0.3 on GF2. Spatial distortion $D_s$ is the mean across bands of the absolute difference between the local MS--PAN quality indices at the fused and degraded scales~\cite{alparone2008qnr}. We retain the toolbox's $33\times33$ sliding window and SSIM data-range convention for digital numbers. The per-image score is $\mathrm{HQNR}=(1-D_\lambda)(1-D_s)$.

\section{Limitations}
\label{supp:limitations}

Some recent backbones gain little from replacing the residual base. On WV3, $\mathbf{s}_{\mathrm{pan}}$ improves PAN-Crafter by only 0.06\,dB and reduces TS-Pan PSNR by 0.40\,dB. On GF2, both still improve, by 0.48 and 2.40\,dB, respectively.

Standalone $\mathbf{s}_{\mathrm{ms}}$ has lower FR HQNR than $\mathbf{s}_{\mathrm{pan}}$ on WV3, QB, and WV2. The gap is largest on QB. MS alignment raises RR PSNR from 33.10 to 36.05\,dB, but spatial distortion $D_s$ rises from 0.0525 to 0.2953 and HQNR falls from 0.944 to 0.703.

\begin{table*}[!t]
\small
\centering
\sethlcolor{BlueGreen!5}
\caption{\textbf{PanBase in PAN geometry on five additional backbones.} \hl{Highlighted rows} use $\mathbf{s}_{\mathrm{pan}}$ as the global residual base and an additional input under matched training.
Deltas are relative to each original backbone. \textbf{Bold} and \underline{underline} mark the best and second-best scores per column. WV2$^{\dagger}$ uses WV3-trained models without fine-tuning.
For each backbone, speed-up is the median ratio of iterations to reach the original best score across reached WV3, QB, and GF2 targets. We report the geometric mean across backbones.}
\vspace{-0.2cm}
\label{tab:h_top_pan_additional}
\setlength{\tabcolsep}{6pt}
\renewcommand{\arraystretch}{1.1}
\resizebox{\textwidth}{!}{%
\begin{tabular}{ll cccc cccc c}
\toprule
\multirow{2}{*}{\textbf{Method}} & \multirow{2}{*}{\textbf{Venue}}
& \multicolumn{4}{c}{\textbf{RR PSNR$\uparrow$ (dB)}} & \multicolumn{4}{c}{\textbf{FR HQNR$\uparrow$}}
& \multirow{2}{*}{\shortstack{\textbf{Training}\\\textbf{speed-up}$\uparrow$}} \\
\cmidrule(lr){3-6}\cmidrule(lr){7-10}
& & WV3 & QB & GF2 & WV2$^{\dagger}$ & WV3 & QB & GF2 & WV2$^{\dagger}$ & \\
\midrule

MSDDN~\cite{he2023msddn} & TGRS'23
& 35.62 & 36.47 & 38.72 & 28.65 & 0.956 & 0.910 & 0.945 & 0.931 & -- \\
\rowcolor{BlueGreen!5}
\; + PanBase ($\mathbf{s}_{\mathrm{pan}}$) & --
& 36.66\,{\scriptsize \hpos{+1.04}} & \textbf{38.11}\,{\scriptsize \hpos{+1.63}} & 41.80\,{\scriptsize \hpos{+3.08}} & 28.82\,{\scriptsize \hpos{+0.17}}
& 0.928\,{\scriptsize \hneg{$-$0.028}} & 0.941\,{\scriptsize \hpos{+0.031}} & 0.956\,{\scriptsize \hpos{+0.011}} & 0.920\,{\scriptsize \hneg{$-$0.011}} & $6.7\times$ \\

UTeRM~\cite{mai2024uterm} & TGRS'24
& 35.96 & 37.26 & 41.89 & 28.85 & 0.942 & \underline{0.949} & 0.940 & 0.954 & -- \\
\rowcolor{BlueGreen!5}
\; + PanBase ($\mathbf{s}_{\mathrm{pan}}$) & --
& 36.45\,{\scriptsize \hpos{+0.50}} & 37.40\,{\scriptsize \hpos{+0.14}} & 42.56\,{\scriptsize \hpos{+0.67}} & 29.13\,{\scriptsize \hpos{+0.28}}
& 0.956\,{\scriptsize \hpos{+0.014}} & 0.941\,{\scriptsize \hneg{$-$0.007}} & 0.966\,{\scriptsize \hpos{+0.026}} & 0.922\,{\scriptsize \hneg{$-$0.032}} & $1.5\times$ \\

U-Know-DiffPAN~\cite{kim2025uknowdiffpan} & CVPR'25
& \underline{37.44} & 37.00 & \underline{43.24} & 29.20 & \underline{0.964} & 0.939 & 0.966 & \underline{0.966} & -- \\
\rowcolor{BlueGreen!5}
\; + PanBase ($\mathbf{s}_{\mathrm{pan}}$) & --
& \textbf{37.83}\,{\scriptsize \hpos{+0.39}} & 37.45\,{\scriptsize \hpos{+0.45}} & \textbf{44.05}\,{\scriptsize \hpos{+0.81}} & \underline{29.80}\,{\scriptsize \hpos{+0.60}}
& 0.959\,{\scriptsize \hneg{$-$0.004}} & 0.818\,{\scriptsize \hneg{$-$0.121}} & \underline{0.976}\,{\scriptsize \hpos{+0.010}} & 0.947\,{\scriptsize \hneg{$-$0.019}} & $3.0\times$ \\

PanDiT~\cite{fang2025pandit} & TGRS'25
& 36.26 & 37.10 & 42.00 & 29.15 & \underline{0.964} & 0.944 & 0.954 & \textbf{0.978} & -- \\
\rowcolor{BlueGreen!5}
\; + PanBase ($\mathbf{s}_{\mathrm{pan}}$) & --
& 36.13\,{\scriptsize \hneg{$-$0.13}} & 36.67\,{\scriptsize \hneg{$-$0.43}} & 42.16\,{\scriptsize \hpos{+0.16}} & \textbf{30.17}\,{\scriptsize \hpos{+1.03}}
& \textbf{0.965}\,{\scriptsize \hpos{+0.001}} & \textbf{0.953}\,{\scriptsize \hpos{+0.009}} & \textbf{0.982}\,{\scriptsize \hpos{+0.027}} & 0.962\,{\scriptsize \hneg{$-$0.016}} & $1.4\times$ \\

MMMamba~\cite{wang2026mmmamba} & AAAI'26
& 36.23 & 36.97 & 39.68 & 28.89 & 0.952 & 0.788 & 0.928 & 0.932 & -- \\
\rowcolor{BlueGreen!5}
\; + PanBase ($\mathbf{s}_{\mathrm{pan}}$) & --
& 37.05\,{\scriptsize \hpos{+0.81}} & \underline{37.75}\,{\scriptsize \hpos{+0.78}} & 42.72\,{\scriptsize \hpos{+3.04}} & 28.88\,{\scriptsize \hneg{$-$0.02}}
& 0.952\,{\scriptsize 0.000} & 0.916\,{\scriptsize \hpos{+0.127}} & 0.961\,{\scriptsize \hpos{+0.034}} & 0.899\,{\scriptsize \hneg{$-$0.033}} & $2.4\times$ \\

\midrule
\multicolumn{2}{l}{Mean change ($\mathbf{s}_{\mathrm{pan}}$)} & \hpos{+0.52} & \hpos{+0.51} & \hpos{+1.55} & \hpos{+0.41} & \hneg{$-$0.003} & \hpos{+0.008} & \hpos{+0.022} & \hneg{$-$0.022} & $2.5\times$ \\
\bottomrule
\end{tabular}}
\end{table*}
\begin{table*}[!t]
\small
\centering
\sethlcolor{BlueGreen!5}
\caption{\textbf{PanBase in MS geometry across 15 backbones.} \hl{Highlighted rows} use $\mathbf{s}_{\mathrm{ms}}$ as the global residual base and an additional input under matched training.
Deltas are relative to each original backbone. \textbf{Bold} and \underline{underline} mark the best and second-best scores per column. WV2$^{\dagger}$ uses WV3-trained models without fine-tuning.
For each backbone, speed-up is the median ratio of iterations to reach the original best score across reached WV3, QB, and GF2 targets. We report the geometric mean across backbones. Dashes indicate unavailable results. The final row uses the same available backbone--dataset pairs for $\mathbf{s}_{\mathrm{pan}}$.}
\vspace{-0.2cm}
\label{tab:h_top_ms}
\setlength{\tabcolsep}{6pt}
\renewcommand{\arraystretch}{1.1}
\resizebox{\textwidth}{!}{%
\begin{tabular}{ll cccc cccc c}
\toprule
\multirow{2}{*}{\textbf{Method}} & \multirow{2}{*}{\textbf{Venue}}
& \multicolumn{4}{c}{\textbf{RR PSNR$\uparrow$ (dB)}} & \multicolumn{4}{c}{\textbf{FR HQNR$\uparrow$}}
& \multirow{2}{*}{\shortstack{\textbf{Training}\\\textbf{speed-up}$\uparrow$}} \\
\cmidrule(lr){3-6}\cmidrule(lr){7-10}
& & WV3 & QB & GF2 & WV2$^{\dagger}$ & WV3 & QB & GF2 & WV2$^{\dagger}$ & \\
\midrule
PanNet~\cite{yang2017pannet} & ICCV'17
& 34.29 & 33.83 & 36.87 & 28.51 & 0.934 & 0.843 & 0.915 & \underline{0.968} & -- \\
\rowcolor{BlueGreen!5}
\; + PanBase ($\mathbf{s}_{\mathrm{ms}}$) & --
& 35.22\,{\scriptsize \hpos{+0.93}} & 36.41\,{\scriptsize \hpos{+2.57}} & 40.06\,{\scriptsize \hpos{+3.18}} & 29.40\,{\scriptsize \hpos{+0.89}}
& \textbf{0.967}\,{\scriptsize \hpos{+0.034}} & 0.881\,{\scriptsize \hpos{+0.038}} & 0.963\,{\scriptsize \hpos{+0.047}} & 0.963\,{\scriptsize \hneg{$-$0.005}} & $60.0\times$ \\

LAGConv~\cite{jin2022lagconv} & AAAI'22
& 37.22 & 37.96 & 42.47 & 28.71 & 0.944 & 0.923 & 0.915 & 0.949 & -- \\
\rowcolor{BlueGreen!5}
\; + PanBase ($\mathbf{s}_{\mathrm{ms}}$) & --
& 37.29\,{\scriptsize \hpos{+0.07}} & 38.12\,{\scriptsize \hpos{+0.16}} & 43.02\,{\scriptsize \hpos{+0.55}} & 29.07\,{\scriptsize \hpos{+0.36}}
& 0.962\,{\scriptsize \hpos{+0.018}} & 0.900\,{\scriptsize \hneg{$-$0.022}} & 0.969\,{\scriptsize \hpos{+0.055}} & 0.936\,{\scriptsize \hneg{$-$0.013}} & $1.7\times$ \\

MutInf~\cite{zhou2022mutinf} & CVPR'22
& 35.77 & 36.64 & 36.65 & 28.87 & 0.913 & 0.876 & 0.943 & 0.902 & -- \\
\rowcolor{BlueGreen!5}
\; + PanBase ($\mathbf{s}_{\mathrm{ms}}$) & --
& 36.97\,{\scriptsize \hpos{+1.20}} & 38.29\,{\scriptsize \hpos{+1.65}} & 43.03\,{\scriptsize \hpos{+6.38}} & 29.23\,{\scriptsize \hpos{+0.36}}
& 0.964\,{\scriptsize \hpos{+0.052}} & 0.927\,{\scriptsize \hpos{+0.051}} & \underline{0.984}\,{\scriptsize \hpos{+0.041}} & 0.935\,{\scriptsize \hpos{+0.033}} & $12.3\times$ \\

MDCUN~\cite{yang2022mdcun} & CVPR'22
& 36.95 & 35.99 & 42.04 & 28.96 & 0.961 & 0.914 & 0.953 & 0.938 & -- \\
\rowcolor{BlueGreen!5}
\; + PanBase ($\mathbf{s}_{\mathrm{ms}}$) & --
& 36.92\,{\scriptsize \hneg{$-$0.03}} & 38.08\,{\scriptsize \hpos{+2.09}} & 43.15\,{\scriptsize \hpos{+1.11}} & 28.79\,{\scriptsize \hneg{$-$0.18}}
& 0.961\,{\scriptsize 0.000} & 0.937\,{\scriptsize \hpos{+0.023}} & \textbf{0.985}\,{\scriptsize \hpos{+0.032}} & 0.941\,{\scriptsize \hpos{+0.003}} & $9.4\times$ \\

PanFlowNet~\cite{yang2023panflownet} & ICCV'23
& 35.76 & 36.23 & 38.56 & 27.78 & 0.918 & 0.853 & 0.904 & 0.869 & -- \\
\rowcolor{BlueGreen!5}
\; + PanBase ($\mathbf{s}_{\mathrm{ms}}$) & --
& 36.56\,{\scriptsize \hpos{+0.80}} & 37.93\,{\scriptsize \hpos{+1.70}} & 40.19\,{\scriptsize \hpos{+1.62}} & 28.22\,{\scriptsize \hpos{+0.44}}
& 0.865\,{\scriptsize \hneg{$-$0.053}} & 0.909\,{\scriptsize \hpos{+0.055}} & 0.934\,{\scriptsize \hpos{+0.030}} & 0.912\,{\scriptsize \hpos{+0.043}} & $7.5\times$ \\

MSDDN~\cite{he2023msddn} & TGRS'23
& 35.62 & 36.47 & 38.72 & 28.65 & 0.956 & 0.910 & 0.945 & 0.931 & -- \\
\rowcolor{BlueGreen!5}
\; + PanBase ($\mathbf{s}_{\mathrm{ms}}$) & --
& 36.85\,{\scriptsize \hpos{+1.24}} & 38.08\,{\scriptsize \hpos{+1.61}} & 41.86\,{\scriptsize \hpos{+3.14}} & 28.68\,{\scriptsize \hpos{+0.03}}
& 0.942\,{\scriptsize \hneg{$-$0.014}} & \underline{0.946}\,{\scriptsize \hpos{+0.035}} & 0.971\,{\scriptsize \hpos{+0.027}} & 0.899\,{\scriptsize \hneg{$-$0.031}} & $7.1\times$ \\

HFIN~\cite{tan2024hfin} & CVPR'24
& 36.53 & 35.17 & 38.08 & 28.39 & 0.917 & 0.796 & 0.909 & 0.903 & -- \\
\rowcolor{BlueGreen!5}
\; + PanBase ($\mathbf{s}_{\mathrm{ms}}$) & --
& 37.40\,{\scriptsize \hpos{+0.87}} & 38.03\,{\scriptsize \hpos{+2.86}} & 42.56\,{\scriptsize \hpos{+4.48}} & 28.68\,{\scriptsize \hpos{+0.29}}
& \underline{0.966}\,{\scriptsize \hpos{+0.050}} & 0.896\,{\scriptsize \hpos{+0.101}} & 0.966\,{\scriptsize \hpos{+0.057}} & 0.922\,{\scriptsize \hpos{+0.019}} & $12.7\times$ \\

UTeRM~\cite{mai2024uterm} & TGRS'24
& 35.96 & 37.26 & 41.89 & 28.85 & 0.942 & \textbf{0.949} & 0.940 & 0.954 & -- \\
\rowcolor{BlueGreen!5}
\; + PanBase ($\mathbf{s}_{\mathrm{ms}}$) & --
& 36.46\,{\scriptsize \hpos{+0.50}} & 37.54\,{\scriptsize \hpos{+0.28}} & 42.69\,{\scriptsize \hpos{+0.80}} & 29.45\,{\scriptsize \hpos{+0.60}}
& 0.963\,{\scriptsize \hpos{+0.021}} & 0.894\,{\scriptsize \hneg{$-$0.055}} & 0.974\,{\scriptsize \hpos{+0.034}} & 0.946\,{\scriptsize \hneg{$-$0.009}} & $1.4\times$ \\

BNNPan~\cite{hou2025bnnpan} & CVPR'25
& 36.73 & 37.38 & 42.12 & 27.35 & 0.936 & 0.856 & 0.927 & 0.755 & -- \\
\rowcolor{BlueGreen!5}
\; + PanBase ($\mathbf{s}_{\mathrm{ms}}$) & --
& -- & -- & -- & --
& -- & -- & -- & -- & -- \\

U-Know-DiffPAN~\cite{kim2025uknowdiffpan} & CVPR'25
& 37.44 & 37.00 & 43.24 & 29.20 & 0.964 & 0.939 & 0.966 & 0.966 & -- \\
\rowcolor{BlueGreen!5}
\; + PanBase ($\mathbf{s}_{\mathrm{ms}}$) & --
& \underline{37.89}\,{\scriptsize \hpos{+0.45}} & 37.90\,{\scriptsize \hpos{+0.90}} & -- & 30.04\,{\scriptsize \hpos{+0.84}}
& 0.963\,{\scriptsize 0.000} & 0.871\,{\scriptsize \hneg{$-$0.068}} & -- & 0.960\,{\scriptsize \hneg{$-$0.006}} & $2.2\times$ \\

PAN-Crafter~\cite{do2025pancrafter} & ICCV'25
& 37.48 & 35.75 & 44.33 & 29.04 & 0.964 & 0.826 & 0.971 & 0.949 & -- \\
\rowcolor{BlueGreen!5}
\; + PanBase ($\mathbf{s}_{\mathrm{ms}}$) & --
& 37.60\,{\scriptsize \hpos{+0.12}} & 37.93\,{\scriptsize \hpos{+2.18}} & \textbf{45.00}\,{\scriptsize \hpos{+0.67}} & 28.88\,{\scriptsize \hneg{$-$0.16}}
& 0.960\,{\scriptsize \hneg{$-$0.004}} & 0.935\,{\scriptsize \hpos{+0.109}} & 0.981\,{\scriptsize \hpos{+0.010}} & 0.933\,{\scriptsize \hneg{$-$0.015}} & $1.7\times$ \\

PINO~\cite{liu2025pino} & NeurIPS'25
& 36.42 & 37.37 & 42.48 & 29.76 & 0.959 & 0.925 & 0.979 & 0.946 & -- \\
\rowcolor{BlueGreen!5}
\; + PanBase ($\mathbf{s}_{\mathrm{ms}}$) & --
& 37.55\,{\scriptsize \hpos{+1.13}} & 38.03\,{\scriptsize \hpos{+0.66}} & \underline{44.66}\,{\scriptsize \hpos{+2.19}} & \textbf{30.43}\,{\scriptsize \hpos{+0.67}}
& 0.954\,{\scriptsize \hneg{$-$0.005}} & 0.945\,{\scriptsize \hpos{+0.020}} & \underline{0.984}\,{\scriptsize \hpos{+0.004}} & 0.941\,{\scriptsize \hneg{$-$0.005}} & $12.0\times$ \\

PanDiT~\cite{fang2025pandit} & TGRS'25
& 36.26 & 37.10 & 42.00 & 29.15 & 0.964 & 0.944 & 0.954 & \textbf{0.978} & -- \\
\rowcolor{BlueGreen!5}
\; + PanBase ($\mathbf{s}_{\mathrm{ms}}$) & --
& 36.47\,{\scriptsize \hpos{+0.21}} & 37.40\,{\scriptsize \hpos{+0.30}} & 42.34\,{\scriptsize \hpos{+0.35}} & \underline{30.37}\,{\scriptsize \hpos{+1.22}}
& 0.962\,{\scriptsize \hneg{$-$0.002}} & 0.837\,{\scriptsize \hneg{$-$0.107}} & 0.978\,{\scriptsize \hpos{+0.024}} & 0.962\,{\scriptsize \hneg{$-$0.016}} & $1.3\times$ \\

MMMamba~\cite{wang2026mmmamba} & AAAI'26
& 36.23 & 36.97 & 39.68 & 28.89 & 0.952 & 0.788 & 0.928 & 0.932 & -- \\
\rowcolor{BlueGreen!5}
\; + PanBase ($\mathbf{s}_{\mathrm{ms}}$) & --
& 36.70\,{\scriptsize \hpos{+0.46}} & -- & -- & 27.96\,{\scriptsize \hneg{$-$0.93}}
& 0.954\,{\scriptsize \hpos{+0.002}} & -- & -- & 0.877\,{\scriptsize \hneg{$-$0.056}} & $2.0\times$ \\

TS-Pan~\cite{li2026tspan} & TGRS'26
& \textbf{37.90} & \textbf{38.53} & 41.94 & 29.59 & 0.946 & 0.743 & 0.944 & 0.920 & -- \\
\rowcolor{BlueGreen!5}
\; + PanBase ($\mathbf{s}_{\mathrm{ms}}$) & --
& 37.83\,{\scriptsize \hneg{$-$0.08}} & \underline{38.49}\,{\scriptsize \hneg{$-$0.04}} & 44.56\,{\scriptsize \hpos{+2.62}} & 30.15\,{\scriptsize \hpos{+0.56}}
& 0.934\,{\scriptsize \hneg{$-$0.012}} & 0.868\,{\scriptsize \hpos{+0.124}} & 0.979\,{\scriptsize \hpos{+0.034}} & 0.935\,{\scriptsize \hpos{+0.015}} & $1.2\times$ \\

\midrule
\multicolumn{2}{l}{Mean change ($\mathbf{s}_{\mathrm{ms}}$)} & \hpos{+0.56} & \hpos{+1.30} & \hpos{+2.26} & \hpos{+0.36} & \hpos{+0.006} & \hpos{+0.023} & \hpos{+0.033} & \hneg{$-$0.003} & $4.5\times$ \\
\multicolumn{2}{l}{Mean change ($\mathbf{s}_{\mathrm{pan}}$), same cells} & \hpos{+0.49} & \hpos{+0.95} & \hpos{+1.82} & \hpos{+0.54} & \hpos{+0.013} & \hpos{+0.024} & \hpos{+0.016} & \hpos{+0.001} & $4.0\times$ \\
\bottomrule
\end{tabular}}
\end{table*}

\begin{table*}[!t]
\small
\centering
\sethlcolor{BlueGreen!5}
\caption{\textbf{PanBase results on WV3.} 
\hl{Highlighted rows} use PanBase as the residual base and an additional input. \hpos{Green} and \hneg{red} indicate gains and drops relative to each original backbone.
\textbf{Bold} and \underline{underline} mark the best and second-best scores per metric. Means exclude missing entries (--).}
\vspace{-0.2cm}
\label{tab:panbase_full_wv3}
\setlength{\tabcolsep}{6pt}
\renewcommand{\arraystretch}{1.1}
\resizebox{\textwidth}{!}{%
\begin{tabular}{l cccccccc ccc}
\toprule
\multirow{2}{*}{\textbf{Method}} & \multicolumn{8}{c}{\textbf{Reduced resolution (HRMS reference)}} & \multicolumn{3}{c}{\textbf{Full resolution (no reference)}} \\
\cmidrule(lr){2-9}\cmidrule(l){10-12}
& PSNR$\uparrow$ & SSIM$\uparrow$ & SAM$\downarrow$ & ERGAS$\downarrow$ & SCC$\uparrow$ & $Q2^{n}\!\uparrow$ & LPIPS$\downarrow$ & DISTS$\downarrow$ & $D_{\lambda}\!\downarrow$ & $D_s\!\downarrow$ & HQNR$\uparrow$ \\
\midrule
PanNet~\cite{yang2017pannet} & 34.29 & 0.9467 & 4.24 & 3.14 & 0.8046 & 0.8719 & 0.0808 & 0.1187 & 0.0107 & 0.0564 & 0.934 \\
\rowcolor{BlueGreen!5}
\quad + PanBase ($\mathbf{s}_{\mathrm{pan}}$) & \hpos{34.96} & \hpos{0.9531} & \hpos{3.89} & \hpos{2.92} & \hpos{0.8132} & \hpos{0.8885} & \hpos{0.0510} & \hpos{0.0895} & \hpos{0.0032} & \hpos{0.0296} & \hpos{\underline{0.967}} \\
\rowcolor{BlueGreen!5}
\quad + PanBase ($\mathbf{s}_{\mathrm{ms}}$) & \hpos{35.22} & \hpos{0.9546} & \hpos{3.91} & \hpos{2.84} & \hpos{0.8252} & \hpos{0.8812} & \hpos{0.0574} & \hpos{0.0900} & \hpos{0.0047} & \hpos{0.0281} & \hpos{\underline{0.967}} \\

LAGConv~\cite{jin2022lagconv} & 37.22 & 0.9713 & 3.09 & 2.25 & 0.8739 & 0.9099 & 0.0317 & 0.0622 & 0.0132 & 0.0438 & 0.944 \\
\rowcolor{BlueGreen!5}
\quad + PanBase ($\mathbf{s}_{\mathrm{pan}}$) & \hpos{37.34} & \hpos{0.9719} & \hpos{3.05} & \hpos{2.21} & \hpos{0.8753} & \hpos{0.9141} & \hpos{0.0275} & \hpos{0.0599} & \hpos{0.0056} & \hpos{0.0368} & \hpos{0.958} \\
\rowcolor{BlueGreen!5}
\quad + PanBase ($\mathbf{s}_{\mathrm{ms}}$) & \hpos{37.29} & \hpos{0.9716} & \hpos{3.08} & \hpos{2.23} & \hneg{0.8732} & \hpos{0.9120} & \hpos{0.0289} & \hpos{0.0602} & \hpos{0.0061} & \hpos{0.0323} & \hpos{0.962} \\

MutInf~\cite{zhou2022mutinf} & 35.77 & 0.9641 & 3.55 & 2.72 & 0.8408 & 0.9030 & 0.0386 & 0.0679 & 0.0320 & 0.0577 & 0.913 \\
\rowcolor{BlueGreen!5}
\quad + PanBase ($\mathbf{s}_{\mathrm{pan}}$) & \hpos{37.05} & \hpos{0.9712} & \hpos{3.10} & \hpos{2.29} & \hpos{0.8693} & \hpos{0.9143} & \hpos{0.0286} & \hpos{0.0578} & \hpos{0.0040} & \hpos{0.0290} & \hpos{\underline{0.967}} \\
\rowcolor{BlueGreen!5}
\quad + PanBase ($\mathbf{s}_{\mathrm{ms}}$) & \hpos{36.97} & \hpos{0.9709} & \hpos{3.15} & \hpos{2.32} & \hpos{0.8670} & \hpos{0.9153} & \hpos{0.0291} & \hpos{0.0563} & \hpos{0.0063} & \hpos{0.0297} & \hpos{0.964} \\

MDCUN~\cite{yang2022mdcun} & 36.95 & 0.9709 & 3.08 & 2.31 & 0.8680 & 0.9156 & 0.0301 & 0.0622 & 0.0080 & 0.0314 & 0.961 \\
\rowcolor{BlueGreen!5}
\quad + PanBase ($\mathbf{s}_{\mathrm{pan}}$) & \hpos{37.12} & \hpos{0.9710} & \hpos{3.06} & \hpos{2.27} & \hpos{0.8701} & \hpos{0.9171} & \hneg{0.0302} & \hpos{0.0601} & \hpos{0.0038} & \hpos{0.0290} & \hpos{\underline{0.967}} \\
\rowcolor{BlueGreen!5}
\quad + PanBase ($\mathbf{s}_{\mathrm{ms}}$) & \hneg{36.92} & \hneg{0.9702} & \hneg{3.12} & \hneg{2.33} & \hneg{0.8648} & \hpos{0.9161} & \hpos{0.0286} & \hpos{0.0590} & \hpos{0.0034} & \hneg{0.0356} & 0.961 \\

PanFlowNet~\cite{yang2023panflownet} & 35.76 & 0.9657 & 3.61 & 2.63 & 0.8512 & 0.9034 & 0.0376 & 0.0736 & 0.0276 & 0.0565 & 0.918 \\
\rowcolor{BlueGreen!5}
\quad + PanBase ($\mathbf{s}_{\mathrm{pan}}$) & \hpos{36.53} & \hpos{0.9679} & \hpos{3.33} & \hpos{2.46} & \hpos{0.8547} & \hpos{0.9039} & \hpos{0.0335} & \hpos{0.0626} & \hpos{0.0116} & \hpos{0.0346} & \hpos{0.954} \\
\rowcolor{BlueGreen!5}
\quad + PanBase ($\mathbf{s}_{\mathrm{ms}}$) & \hpos{36.56} & \hpos{0.9686} & \hpos{3.27} & \hpos{2.44} & \hpos{0.8602} & \hpos{0.9045} & \hpos{0.0328} & \hpos{0.0630} & \hneg{0.0501} & \hneg{0.0997} & \hneg{0.865} \\

MSDDN~\cite{he2023msddn} & 35.62 & 0.9655 & 3.49 & 2.68 & 0.8582 & 0.8968 & 0.0360 & 0.0705 & 0.0215 & \textbf{0.0234} & 0.956 \\
\rowcolor{BlueGreen!5}
\quad + PanBase ($\mathbf{s}_{\mathrm{pan}}$) & \hpos{36.66} & \hpos{0.9692} & \hpos{3.26} & \hpos{2.39} & \hpos{0.8643} & \hpos{0.9091} & \hpos{0.0314} & \hpos{0.0598} & \hpos{0.0069} & \hneg{0.0658} & \hneg{0.928} \\
\rowcolor{BlueGreen!5}
\quad + PanBase ($\mathbf{s}_{\mathrm{ms}}$) & \hpos{36.85} & \hpos{0.9696} & \hpos{3.22} & \hpos{2.37} & \hpos{0.8705} & \hpos{0.9100} & \hpos{0.0290} & \hpos{0.0583} & \hpos{0.0080} & \hneg{0.0507} & \hneg{0.942} \\

HFIN~\cite{tan2024hfin} & 36.53 & 0.9696 & 3.31 & 2.46 & 0.8687 & 0.8987 & 0.0361 & 0.0682 & 0.0300 & 0.0556 & 0.917 \\
\rowcolor{BlueGreen!5}
\quad + PanBase ($\mathbf{s}_{\mathrm{pan}}$) & \hpos{37.26} & \hpos{0.9720} & \hpos{3.03} & \hpos{2.23} & \hpos{0.8740} & \hpos{0.9166} & \hpos{0.0297} & \hpos{0.0592} & \hpos{0.0050} & \hpos{0.0301} & \hpos{0.965} \\
\rowcolor{BlueGreen!5}
\quad + PanBase ($\mathbf{s}_{\mathrm{ms}}$) & \hpos{37.40} & \hpos{0.9726} & \hpos{3.01} & \hpos{2.19} & \hpos{0.8766} & \hpos{0.9169} & \hpos{0.0304} & \hpos{0.0590} & \hpos{0.0052} & \hpos{0.0289} & \hpos{0.966} \\

UTeRM~\cite{mai2024uterm} & 35.96 & 0.9630 & 3.39 & 2.59 & 0.8436 & 0.9025 & 0.0417 & 0.0762 & 0.0165 & 0.0422 & 0.942 \\
\rowcolor{BlueGreen!5}
\quad + PanBase ($\mathbf{s}_{\mathrm{pan}}$) & \hpos{36.45} & \hpos{0.9667} & \hpos{3.30} & \hpos{2.45} & \hpos{0.8564} & \hpos{0.9077} & \hpos{0.0340} & \hpos{0.0653} & \hpos{0.0140} & \hpos{0.0303} & \hpos{0.956} \\
\rowcolor{BlueGreen!5}
\quad + PanBase ($\mathbf{s}_{\mathrm{ms}}$) & \hpos{36.46} & \hpos{0.9664} & \hpos{3.29} & \hpos{2.45} & \hpos{0.8548} & \hpos{0.9079} & \hpos{0.0362} & \hpos{0.0692} & \hpos{0.0085} & \hpos{0.0287} & \hpos{0.963} \\

BNNPan~\cite{hou2025bnnpan} & 36.73 & 0.9691 & 3.20 & 2.37 & 0.8641 & 0.9110 & 0.0318 & 0.0657 & 0.0170 & 0.0484 & 0.936 \\
\rowcolor{BlueGreen!5}
\quad + PanBase ($\mathbf{s}_{\mathrm{pan}}$) & \hpos{37.05} & \hpos{0.9710} & \hpos{3.10} & \hpos{2.28} & \hpos{0.8686} & \hpos{0.9144} & \hpos{0.0289} & \hpos{0.0589} & \hpos{0.0067} & \hpos{\underline{0.0242}} & \hpos{\textbf{0.969}} \\
\rowcolor{BlueGreen!5}
\quad + PanBase ($\mathbf{s}_{\mathrm{ms}}$) & -- & -- & -- & -- & -- & -- & -- & -- & -- & -- & -- \\

U-Know-DiffPAN~\cite{kim2025uknowdiffpan} & 37.44 & 0.9728 & 2.91 & 2.18 & 0.8783 & 0.9191 & 0.0297 & 0.0618 & 0.0070 & 0.0298 & 0.964 \\
\rowcolor{BlueGreen!5}
\quad + PanBase ($\mathbf{s}_{\mathrm{pan}}$) & \hpos{37.83} & \hpos{0.9747} & \hpos{\underline{2.83}} & \hpos{2.09} & \hpos{0.8843} & \hpos{\underline{0.9224}} & \hpos{0.0250} & \hpos{0.0566} & \hpos{0.0044} & \hneg{0.0363} & \hneg{0.959} \\
\rowcolor{BlueGreen!5}
\quad + PanBase ($\mathbf{s}_{\mathrm{ms}}$) & \hpos{\underline{37.89}} & \hpos{\underline{0.9750}} & \hpos{\textbf{2.82}} & \hpos{\textbf{2.07}} & \hpos{\textbf{0.8859}} & \hpos{\textbf{0.9226}} & \hpos{0.0244} & \hpos{0.0561} & \hpos{0.0047} & \hneg{0.0321} & \hneg{0.963} \\

PAN-Crafter~\cite{do2025pancrafter} & 37.48 & 0.9733 & 2.93 & 2.18 & 0.8772 & 0.9183 & 0.0284 & 0.0591 & 0.0090 & 0.0269 & 0.964 \\
\rowcolor{BlueGreen!5}
\quad + PanBase ($\mathbf{s}_{\mathrm{pan}}$) & \hpos{37.54} & \hpos{0.9738} & 2.93 & \hpos{2.16} & \hneg{0.8764} & \hpos{0.9206} & \hpos{0.0249} & \hpos{0.0559} & \hpos{0.0039} & \hneg{0.0341} & \hneg{0.962} \\
\rowcolor{BlueGreen!5}
\quad + PanBase ($\mathbf{s}_{\mathrm{ms}}$) & \hpos{37.60} & \hpos{0.9739} & \hpos{2.91} & \hpos{2.14} & \hpos{0.8784} & \hpos{0.9199} & \hpos{0.0258} & \hpos{0.0575} & \hpos{0.0055} & \hneg{0.0344} & \hneg{0.960} \\

PINO~\cite{liu2025pino} & 36.42 & 0.9675 & 3.36 & 2.48 & 0.8554 & 0.9095 & 0.0440 & 0.0737 & 0.0101 & 0.0312 & 0.959 \\
\rowcolor{BlueGreen!5}
\quad + PanBase ($\mathbf{s}_{\mathrm{pan}}$) & \hpos{37.26} & \hpos{0.9717} & \hpos{3.07} & \hpos{2.24} & \hpos{0.8686} & \hpos{0.9136} & \hpos{0.0271} & \hpos{0.0570} & \hpos{0.0035} & \hneg{0.0353} & \hpos{0.961} \\
\rowcolor{BlueGreen!5}
\quad + PanBase ($\mathbf{s}_{\mathrm{ms}}$) & \hpos{37.55} & \hpos{0.9735} & \hpos{2.99} & \hpos{2.17} & \hpos{0.8774} & \hpos{0.9188} & \hpos{0.0239} & \hpos{0.0543} & \hpos{0.0042} & \hneg{0.0423} & \hneg{0.954} \\

PanDiT~\cite{fang2025pandit} & 36.26 & 0.9655 & 3.31 & 2.51 & 0.8440 & 0.8998 & 0.0343 & 0.0670 & 0.0050 & 0.0308 & 0.964 \\
\rowcolor{BlueGreen!5}
\quad + PanBase ($\mathbf{s}_{\mathrm{pan}}$) & \hneg{36.13} & \hneg{0.9641} & \hneg{3.37} & \hneg{2.56} & \hneg{0.8370} & \hpos{0.9018} & \hpos{0.0313} & \hpos{0.0638} & \hpos{\textbf{0.0012}} & \hneg{0.0338} & \hpos{0.965} \\
\rowcolor{BlueGreen!5}
\quad + PanBase ($\mathbf{s}_{\mathrm{ms}}$) & \hpos{36.47} & \hpos{0.9664} & \hpos{3.30} & \hpos{2.45} & \hpos{0.8463} & \hpos{0.9042} & \hpos{0.0294} & \hpos{0.0613} & \hpos{\underline{0.0015}} & \hneg{0.0366} & \hneg{0.962} \\

MMMamba~\cite{wang2026mmmamba} & 36.23 & 0.9690 & 3.29 & 2.55 & 0.8581 & 0.9150 & 0.0269 & 0.0565 & 0.0045 & 0.0436 & 0.952 \\
\rowcolor{BlueGreen!5}
\quad + PanBase ($\mathbf{s}_{\mathrm{pan}}$) & \hpos{37.05} & \hpos{0.9720} & \hpos{3.02} & \hpos{2.33} & \hpos{0.8697} & \hpos{0.9188} & \hpos{0.0233} & \hpos{\textbf{0.0511}} & \hneg{0.0069} & \hpos{0.0413} & 0.952 \\
\rowcolor{BlueGreen!5}
\quad + PanBase ($\mathbf{s}_{\mathrm{ms}}$) & \hpos{36.70} & \hpos{0.9711} & \hpos{3.11} & \hpos{2.34} & \hpos{0.8676} & \hpos{0.9181} & \hpos{0.0265} & \hpos{\underline{0.0531}} & \hpos{0.0041} & \hpos{0.0424} & \hpos{0.954} \\

TS-Pan~\cite{li2026tspan} & \textbf{37.90} & \textbf{0.9754} & 2.84 & \textbf{2.07} & \underline{0.8847} & 0.9173 & \underline{0.0230} & 0.0543 & 0.0051 & 0.0495 & 0.946 \\
\rowcolor{BlueGreen!5}
\quad + PanBase ($\mathbf{s}_{\mathrm{pan}}$) & \hneg{37.51} & \hneg{0.9716} & \hneg{3.01} & \hneg{2.19} & \hneg{0.8593} & \hneg{0.9061} & \hneg{0.0326} & \hneg{0.1059} & \hpos{\underline{0.0015}} & \hpos{0.0432} & \hpos{0.955} \\
\rowcolor{BlueGreen!5}
\quad + PanBase ($\mathbf{s}_{\mathrm{ms}}$) & \hneg{37.83} & \hneg{\underline{0.9750}} & \hneg{2.89} & \hneg{\underline{2.08}} & \hneg{0.8814} & \hneg{0.9125} & \hpos{\textbf{0.0226}} & \hneg{0.0550} & \hpos{0.0030} & \hneg{0.0632} & \hneg{0.934} \\

\midrule
Mean change ($\mathbf{s}_{\mathrm{pan}}$) & \hpos{+0.48} & \hpos{+0.0022} & \hpos{$-$0.15} & \hpos{$-$0.14} & \hpos{+0.0047} & \hpos{+0.0051} & \hpos{$-$0.0061} & \hpos{$-$0.0050} & \hpos{$-$0.0090} & \hpos{$-$0.0062} & \hpos{+0.015} \\
Mean change ($\mathbf{s}_{\mathrm{ms}}$) & \hpos{+0.56} & \hpos{+0.0028} & \hpos{$-$0.17} & \hpos{$-$0.17} & \hpos{+0.0087} & \hpos{+0.0057} & \hpos{$-$0.0067} & \hpos{$-$0.0085} & \hpos{$-$0.0060} & \hneg{+0.0004} & \hpos{+0.006} \\
\bottomrule
\end{tabular}}
\end{table*}

\begin{table*}[!t]
\small
\centering
\sethlcolor{BlueGreen!5}
\caption{\textbf{PanBase results on QB.} 
\hl{Highlighted rows} use PanBase as the residual base and an additional input. \hpos{Green} and \hneg{red} indicate gains and drops relative to each original backbone.
\textbf{Bold} and \underline{underline} mark the best and second-best scores per metric. Means exclude missing entries (--).}
\vspace{-0.2cm}
\label{tab:panbase_full_qb}
\setlength{\tabcolsep}{6pt}
\renewcommand{\arraystretch}{1.1}
\resizebox{\textwidth}{!}{%
\begin{tabular}{l cccccccc ccc}
\toprule
\multirow{2}{*}{\textbf{Method}} & \multicolumn{8}{c}{\textbf{Reduced resolution (HRMS reference)}} & \multicolumn{3}{c}{\textbf{Full resolution (no reference)}} \\
\cmidrule(lr){2-9}\cmidrule(l){10-12}
& PSNR$\uparrow$ & SSIM$\uparrow$ & SAM$\downarrow$ & ERGAS$\downarrow$ & SCC$\uparrow$ & $Q2^{n}\!\uparrow$ & LPIPS$\downarrow$ & DISTS$\downarrow$ & $D_{\lambda}\!\downarrow$ & $D_s\!\downarrow$ & HQNR$\uparrow$ \\
\midrule
PanNet~\cite{yang2017pannet} & 33.83 & 0.9084 & 6.35 & 6.01 & 0.7354 & 0.8826 & 0.1197 & 0.1565 & 0.0262 & 0.1350 & 0.843 \\
\rowcolor{BlueGreen!5}
\quad + PanBase ($\mathbf{s}_{\mathrm{pan}}$) & \hpos{34.20} & \hpos{0.9125} & \hpos{6.04} & \hpos{5.75} & \hpos{0.7378} & \hpos{0.8889} & \hpos{0.0947} & \hpos{0.1302} & \hneg{0.0409} & \hpos{0.1064} & \hpos{0.857} \\
\rowcolor{BlueGreen!5}
\quad + PanBase ($\mathbf{s}_{\mathrm{ms}}$) & \hpos{36.41} & \hpos{0.9405} & \hpos{5.53} & \hpos{4.45} & \hpos{0.8315} & \hpos{0.9178} & \hpos{0.0691} & \hpos{0.1128} & \hneg{0.0502} & \hpos{0.0723} & \hpos{0.881} \\

LAGConv~\cite{jin2022lagconv} & 37.96 & 0.9598 & 4.54 & 3.73 & 0.8636 & 0.9344 & 0.0583 & 0.0960 & 0.0245 & 0.0539 & 0.923 \\
\rowcolor{BlueGreen!5}
\quad + PanBase ($\mathbf{s}_{\mathrm{pan}}$) & \hpos{38.11} & \hpos{0.9607} & \hpos{4.50} & \hpos{3.66} & \hpos{0.8658} & \hpos{0.9374} & \hpos{0.0502} & \hpos{0.0916} & \hpos{0.0231} & \hneg{0.0677} & \hneg{0.911} \\
\rowcolor{BlueGreen!5}
\quad + PanBase ($\mathbf{s}_{\mathrm{ms}}$) & \hpos{38.12} & \hpos{0.9609} & \hpos{4.50} & \hpos{3.66} & \hpos{0.8645} & \hpos{0.9369} & \hpos{0.0497} & \hpos{0.0910} & \hpos{0.0230} & \hneg{0.0782} & \hneg{0.900} \\

MutInf~\cite{zhou2022mutinf} & 36.64 & 0.9510 & 5.07 & 4.37 & 0.8416 & 0.9174 & 0.0704 & 0.1095 & 0.0609 & 0.0681 & 0.876 \\
\rowcolor{BlueGreen!5}
\quad + PanBase ($\mathbf{s}_{\mathrm{pan}}$) & \hpos{38.33} & \hpos{0.9628} & \hpos{4.41} & \hpos{3.57} & \hpos{\underline{0.8693}} & \hpos{\underline{0.9393}} & \hpos{0.0501} & \hpos{0.0912} & \hpos{0.0392} & \hpos{0.0583} & \hpos{0.905} \\
\rowcolor{BlueGreen!5}
\quad + PanBase ($\mathbf{s}_{\mathrm{ms}}$) & \hpos{38.29} & \hpos{0.9628} & \hpos{4.41} & \hpos{3.59} & \hpos{0.8664} & \hpos{0.9389} & \hpos{0.0560} & \hpos{0.0902} & \hpos{0.0353} & \hpos{0.0394} & \hpos{0.927} \\

MDCUN~\cite{yang2022mdcun} & 35.99 & 0.9451 & 4.92 & 4.73 & 0.8089 & 0.9153 & 0.0663 & 0.1055 & 0.0321 & 0.0564 & 0.914 \\
\rowcolor{BlueGreen!5}
\quad + PanBase ($\mathbf{s}_{\mathrm{pan}}$) & \hpos{37.66} & \hpos{0.9580} & \hpos{4.56} & \hpos{3.86} & \hpos{0.8553} & \hpos{0.9335} & \hpos{0.0573} & \hpos{0.0961} & \hpos{0.0153} & \hpos{0.0426} & \hpos{0.943} \\
\rowcolor{BlueGreen!5}
\quad + PanBase ($\mathbf{s}_{\mathrm{ms}}$) & \hpos{38.08} & \hpos{0.9606} & \hpos{4.45} & \hpos{3.67} & \hpos{0.8642} & \hpos{0.9364} & \hpos{0.0542} & \hpos{0.0937} & \hpos{0.0213} & \hpos{0.0427} & \hpos{0.937} \\

PanFlowNet~\cite{yang2023panflownet} & 36.23 & 0.9419 & 5.14 & 4.72 & 0.7913 & 0.9123 & 0.0653 & 0.1053 & 0.0712 & 0.0813 & 0.853 \\
\rowcolor{BlueGreen!5}
\quad + PanBase ($\mathbf{s}_{\mathrm{pan}}$) & \hpos{37.87} & \hpos{0.9589} & \hpos{4.62} & \hpos{3.76} & \hpos{0.8626} & \hpos{0.9329} & \hpos{0.0557} & \hpos{0.0937} & \hneg{0.1184} & \hpos{0.0575} & \hneg{0.831} \\
\rowcolor{BlueGreen!5}
\quad + PanBase ($\mathbf{s}_{\mathrm{ms}}$) & \hpos{37.93} & \hpos{0.9590} & \hpos{4.62} & \hpos{3.74} & \hpos{0.8643} & \hpos{0.9343} & \hpos{0.0518} & \hpos{0.0944} & \hpos{0.0426} & \hpos{0.0509} & \hpos{0.909} \\

MSDDN~\cite{he2023msddn} & 36.47 & 0.9459 & 5.29 & 4.43 & 0.8451 & 0.9199 & 0.0572 & 0.1024 & 0.0503 & 0.0415 & 0.910 \\
\rowcolor{BlueGreen!5}
\quad + PanBase ($\mathbf{s}_{\mathrm{pan}}$) & \hpos{38.11} & \hpos{0.9608} & \hpos{4.55} & \hpos{3.66} & \hpos{0.8672} & \hpos{0.9368} & \hpos{0.0489} & \hpos{\textbf{0.0846}} & \hpos{0.0247} & \hpos{0.0353} & \hpos{0.941} \\
\rowcolor{BlueGreen!5}
\quad + PanBase ($\mathbf{s}_{\mathrm{ms}}$) & \hpos{38.08} & \hpos{0.9607} & \hpos{4.52} & \hpos{3.67} & \hpos{0.8651} & \hpos{0.9376} & \hpos{0.0510} & \hpos{0.0899} & \hpos{0.0238} & \hpos{\underline{0.0313}} & \hpos{0.946} \\

HFIN~\cite{tan2024hfin} & 35.17 & 0.9263 & 5.45 & 5.35 & 0.7545 & 0.8920 & 0.0819 & 0.1187 & 0.1024 & 0.1141 & 0.796 \\
\rowcolor{BlueGreen!5}
\quad + PanBase ($\mathbf{s}_{\mathrm{pan}}$) & \hpos{37.42} & \hpos{0.9544} & \hpos{4.77} & \hpos{3.97} & \hpos{0.8524} & \hpos{0.9306} & \hpos{0.0547} & \hpos{0.0939} & \hpos{0.0341} & \hpos{0.0436} & \hpos{0.924} \\
\rowcolor{BlueGreen!5}
\quad + PanBase ($\mathbf{s}_{\mathrm{ms}}$) & \hpos{38.03} & \hpos{0.9599} & \hpos{4.50} & \hpos{3.70} & \hpos{0.8650} & \hpos{0.9355} & \hpos{0.0543} & \hpos{0.0941} & \hpos{0.0402} & \hpos{0.0665} & \hpos{0.896} \\

UTeRM~\cite{mai2024uterm} & 37.26 & 0.9544 & 4.68 & 4.04 & 0.8462 & 0.9278 & 0.0755 & 0.1063 & 0.0197 & 0.0324 & \underline{0.949} \\
\rowcolor{BlueGreen!5}
\quad + PanBase ($\mathbf{s}_{\mathrm{pan}}$) & \hpos{37.40} & \hpos{0.9552} & \hneg{4.69} & \hpos{3.97} & \hpos{0.8512} & \hpos{0.9280} & \hpos{0.0731} & \hpos{0.1061} & \hneg{0.0303} & \hpos{\textbf{0.0295}} & \hneg{0.941} \\
\rowcolor{BlueGreen!5}
\quad + PanBase ($\mathbf{s}_{\mathrm{ms}}$) & \hpos{37.54} & \hpos{0.9564} & \hpos{4.63} & \hpos{3.90} & \hpos{0.8536} & \hpos{0.9308} & \hpos{0.0670} & \hpos{0.1042} & \hneg{0.0621} & \hneg{0.0469} & \hneg{0.894} \\

BNNPan~\cite{hou2025bnnpan} & 37.38 & 0.9541 & 4.82 & 3.99 & 0.8515 & 0.9293 & 0.0571 & 0.1004 & 0.0451 & 0.1032 & 0.856 \\
\rowcolor{BlueGreen!5}
\quad + PanBase ($\mathbf{s}_{\mathrm{pan}}$) & \hpos{37.71} & \hpos{0.9570} & \hpos{4.70} & \hpos{3.83} & \hpos{0.8571} & \hpos{0.9326} & \hpos{0.0534} & \hpos{0.0945} & \hpos{0.0307} & \hpos{0.0923} & \hpos{0.880} \\
\rowcolor{BlueGreen!5}
\quad + PanBase ($\mathbf{s}_{\mathrm{ms}}$) & -- & -- & -- & -- & -- & -- & -- & -- & -- & -- & -- \\

U-Know-DiffPAN~\cite{kim2025uknowdiffpan} & 37.00 & 0.9534 & 4.61 & 4.15 & 0.8188 & 0.9304 & 0.0570 & 0.0955 & 0.0211 & 0.0411 & 0.939 \\
\rowcolor{BlueGreen!5}
\quad + PanBase ($\mathbf{s}_{\mathrm{pan}}$) & \hpos{37.45} & \hpos{0.9572} & \hpos{4.48} & \hpos{3.94} & \hpos{0.8305} & \hpos{0.9330} & \hpos{0.0567} & \hpos{0.0916} & \hneg{0.1018} & \hneg{0.0893} & \hneg{0.818} \\
\rowcolor{BlueGreen!5}
\quad + PanBase ($\mathbf{s}_{\mathrm{ms}}$) & \hpos{37.90} & \hpos{0.9602} & \hpos{4.38} & \hpos{3.75} & \hpos{0.8468} & \hpos{0.9368} & \hpos{0.0526} & \hpos{0.0897} & \hneg{0.0536} & \hneg{0.0798} & \hneg{0.871} \\

PAN-Crafter~\cite{do2025pancrafter} & 35.75 & 0.9271 & 5.34 & 5.23 & 0.7378 & 0.8920 & 0.0853 & 0.1097 & 0.0684 & 0.1133 & 0.826 \\
\rowcolor{BlueGreen!5}
\quad + PanBase ($\mathbf{s}_{\mathrm{pan}}$) & \hpos{37.74} & \hpos{0.9571} & \hpos{4.74} & \hpos{3.83} & \hpos{0.8553} & \hpos{0.9336} & \hpos{0.0563} & \hpos{0.0972} & \hpos{0.0474} & \hpos{0.1116} & \hpos{0.846} \\
\rowcolor{BlueGreen!5}
\quad + PanBase ($\mathbf{s}_{\mathrm{ms}}$) & \hpos{37.93} & \hpos{0.9605} & \hpos{4.53} & \hpos{3.73} & \hpos{0.8580} & \hpos{0.9368} & \hpos{\underline{0.0480}} & \hpos{0.0923} & \hpos{0.0180} & \hpos{0.0479} & \hpos{0.935} \\

PINO~\cite{liu2025pino} & 37.37 & 0.9529 & 5.40 & 4.00 & 0.8509 & 0.9258 & 0.0947 & 0.1279 & 0.0200 & 0.0558 & 0.925 \\
\rowcolor{BlueGreen!5}
\quad + PanBase ($\mathbf{s}_{\mathrm{pan}}$) & \hpos{38.12} & \hpos{0.9604} & \hpos{4.59} & \hpos{3.65} & \hpos{0.8648} & \hpos{0.9365} & \hpos{0.0577} & \hpos{0.0920} & \hneg{0.0231} & \hpos{0.0431} & \hpos{0.935} \\
\rowcolor{BlueGreen!5}
\quad + PanBase ($\mathbf{s}_{\mathrm{ms}}$) & \hpos{38.03} & \hpos{0.9602} & \hpos{4.53} & \hpos{3.69} & \hpos{0.8616} & \hpos{0.9375} & \hpos{0.0532} & \hpos{0.0911} & \hpos{0.0176} & \hpos{0.0376} & \hpos{0.945} \\

PanDiT~\cite{fang2025pandit} & 37.10 & 0.9522 & 4.91 & 4.11 & 0.8406 & 0.9282 & 0.0619 & 0.0993 & 0.0146 & 0.0424 & 0.944 \\
\rowcolor{BlueGreen!5}
\quad + PanBase ($\mathbf{s}_{\mathrm{pan}}$) & \hneg{36.67} & \hneg{0.9481} & \hneg{4.98} & \hneg{4.32} & \hneg{0.8265} & \hneg{0.9244} & \hpos{0.0603} & \hpos{0.0949} & \hpos{\underline{0.0059}} & \hpos{0.0417} & \hpos{\textbf{0.953}} \\
\rowcolor{BlueGreen!5}
\quad + PanBase ($\mathbf{s}_{\mathrm{ms}}$) & \hpos{37.40} & \hpos{0.9541} & \hpos{4.81} & \hpos{3.97} & \hpos{0.8473} & \hpos{0.9296} & \hpos{0.0573} & \hpos{0.0917} & \hpos{\textbf{0.0049}} & \hneg{0.1590} & \hneg{0.837} \\

MMMamba~\cite{wang2026mmmamba} & 36.97 & 0.9561 & 4.74 & 4.19 & 0.8294 & 0.9269 & 0.0505 & 0.0929 & 0.0633 & 0.1586 & 0.788 \\
\rowcolor{BlueGreen!5}
\quad + PanBase ($\mathbf{s}_{\mathrm{pan}}$) & \hpos{37.75} & \hpos{0.9605} & \hpos{4.46} & \hpos{3.81} & \hpos{0.8514} & \hpos{0.9365} & \hneg{0.0508} & \hpos{0.0894} & \hpos{0.0382} & \hpos{0.0478} & \hpos{0.916} \\
\rowcolor{BlueGreen!5}
\quad + PanBase ($\mathbf{s}_{\mathrm{ms}}$) & -- & -- & -- & -- & -- & -- & -- & -- & -- & -- & -- \\

TS-Pan~\cite{li2026tspan} & \underline{38.53} & 0.9634 & \underline{4.33} & \textbf{3.48} & 0.8674 & 0.9384 & 0.0535 & 0.0971 & 0.1402 & 0.1357 & 0.743 \\
\rowcolor{BlueGreen!5}
\quad + PanBase ($\mathbf{s}_{\mathrm{pan}}$) & \hpos{\textbf{38.54}} & \hpos{\textbf{0.9642}} & \hpos{\textbf{4.31}} & \textbf{3.48} & \hpos{\textbf{0.8697}} & \hpos{\textbf{0.9407}} & \hpos{0.0492} & \hpos{\underline{0.0869}} & \hpos{0.0251} & \hpos{0.0341} & \hpos{0.942} \\
\rowcolor{BlueGreen!5}
\quad + PanBase ($\mathbf{s}_{\mathrm{ms}}$) & \hneg{38.49} & \hpos{\underline{0.9636}} & \hneg{4.37} & \hneg{\underline{3.50}} & \hneg{0.8652} & \hpos{0.9391} & \hpos{\textbf{0.0475}} & \hpos{0.0896} & \hpos{0.0095} & \hpos{0.1239} & \hpos{0.868} \\

\midrule
Mean change ($\mathbf{s}_{\mathrm{pan}}$) & \hpos{+0.89} & \hpos{+0.0091} & \hpos{$-$0.35} & \hpos{$-$0.50} & \hpos{+0.0289} & \hpos{+0.0128} & \hpos{$-$0.0124} & \hpos{$-$0.0126} & \hpos{$-$0.0108} & \hpos{$-$0.0221} & \hpos{+0.030} \\
Mean change ($\mathbf{s}_{\mathrm{ms}}$) & \hpos{+1.30} & \hpos{+0.0137} & \hpos{$-$0.48} & \hpos{$-$0.72} & \hpos{+0.0424} & \hpos{+0.0178} & \hpos{$-$0.0181} & \hpos{$-$0.0158} & \hpos{$-$0.0192} & \hpos{$-$0.0073} & \hpos{+0.023} \\
\bottomrule
\end{tabular}}
\end{table*}

\begin{table*}[!t]
\small
\centering
\sethlcolor{BlueGreen!5}
\caption{\textbf{PanBase results on GF2.} 
\hl{Highlighted rows} use PanBase as the residual base and an additional input. \hpos{Green} and \hneg{red} indicate gains and drops relative to each original backbone.
\textbf{Bold} and \underline{underline} mark the best and second-best scores per metric. Means exclude missing entries (--).}
\vspace{-0.2cm}
\label{tab:panbase_full_gf2}
\setlength{\tabcolsep}{6pt}
\renewcommand{\arraystretch}{1.1}
\resizebox{\textwidth}{!}{%
\begin{tabular}{l cccccccc ccc}
\toprule
\multirow{2}{*}{\textbf{Method}} & \multicolumn{8}{c}{\textbf{Reduced resolution (HRMS reference)}} & \multicolumn{3}{c}{\textbf{Full resolution (no reference)}} \\
\cmidrule(lr){2-9}\cmidrule(l){10-12}
& PSNR$\uparrow$ & SSIM$\uparrow$ & SAM$\downarrow$ & ERGAS$\downarrow$ & SCC$\uparrow$ & $Q2^{n}\!\uparrow$ & LPIPS$\downarrow$ & DISTS$\downarrow$ & $D_{\lambda}\!\downarrow$ & $D_s\!\downarrow$ & HQNR$\uparrow$ \\
\midrule
PanNet~\cite{yang2017pannet} & 36.87 & 0.9361 & 1.33 & 1.34 & 0.7032 & 0.9371 & 0.0846 & 0.1205 & 0.0041 & 0.0811 & 0.915 \\
\rowcolor{BlueGreen!5}
\quad + PanBase ($\mathbf{s}_{\mathrm{pan}}$) & \hpos{37.98} & \hpos{0.9460} & \hpos{1.15} & \hpos{1.19} & \hpos{0.7276} & \hpos{0.9499} & \hpos{0.0609} & \hpos{0.0957} & \hpos{0.0018} & \hpos{0.0711} & \hpos{0.927} \\
\rowcolor{BlueGreen!5}
\quad + PanBase ($\mathbf{s}_{\mathrm{ms}}$) & \hpos{40.06} & \hpos{0.9629} & \hpos{1.03} & \hpos{0.92} & \hpos{0.8412} & \hpos{0.9642} & \hpos{0.0537} & \hpos{0.0918} & \hpos{0.0012} & \hpos{0.0363} & \hpos{0.963} \\

LAGConv~\cite{jin2022lagconv} & 42.47 & 0.9795 & 0.78 & 0.70 & 0.8921 & 0.9795 & 0.0252 & 0.0662 & 0.0041 & 0.0817 & 0.915 \\
\rowcolor{BlueGreen!5}
\quad + PanBase ($\mathbf{s}_{\mathrm{pan}}$) & \hpos{42.82} & \hpos{0.9806} & \hpos{0.76} & \hpos{0.67} & \hpos{0.8933} & \hpos{0.9811} & \hpos{0.0230} & \hpos{0.0622} & \hpos{0.0015} & \hpos{0.0496} & \hpos{0.949} \\
\rowcolor{BlueGreen!5}
\quad + PanBase ($\mathbf{s}_{\mathrm{ms}}$) & \hpos{43.02} & \hpos{0.9810} & \hpos{0.76} & \hpos{0.66} & \hpos{0.8964} & \hpos{0.9815} & \hpos{0.0228} & \hpos{0.0627} & \hpos{0.0008} & \hpos{0.0300} & \hpos{0.969} \\

MutInf~\cite{zhou2022mutinf} & 36.65 & 0.9545 & 1.35 & 1.39 & 0.8053 & 0.9413 & 0.0434 & 0.0962 & 0.0228 & 0.0355 & 0.943 \\
\rowcolor{BlueGreen!5}
\quad + PanBase ($\mathbf{s}_{\mathrm{pan}}$) & \hpos{43.59} & \hpos{0.9832} & \hpos{0.72} & \hpos{0.61} & \hpos{0.9100} & \hpos{0.9839} & \hpos{0.0186} & \hpos{0.0551} & \hpos{0.0015} & \hpos{0.0252} & \hpos{0.973} \\
\rowcolor{BlueGreen!5}
\quad + PanBase ($\mathbf{s}_{\mathrm{ms}}$) & \hpos{43.03} & \hpos{0.9814} & \hpos{0.74} & \hpos{0.65} & \hpos{0.9011} & \hpos{0.9817} & \hpos{0.0202} & \hpos{0.0572} & \hpos{0.0007} & \hpos{0.0153} & \hpos{0.984} \\

MDCUN~\cite{yang2022mdcun} & 42.04 & 0.9768 & 0.81 & 0.74 & 0.8721 & 0.9783 & 0.0241 & 0.0625 & 0.0006 & 0.0467 & 0.953 \\
\rowcolor{BlueGreen!5}
\quad + PanBase ($\mathbf{s}_{\mathrm{pan}}$) & \hpos{42.57} & \hpos{0.9788} & \hpos{0.77} & \hpos{0.70} & \hpos{0.8813} & \hpos{0.9801} & \hpos{0.0213} & \hpos{0.0597} & \hpos{0.0004} & \hpos{0.0358} & \hpos{0.964} \\
\rowcolor{BlueGreen!5}
\quad + PanBase ($\mathbf{s}_{\mathrm{ms}}$) & \hpos{43.15} & \hpos{0.9818} & \hpos{0.74} & \hpos{0.65} & \hpos{0.8926} & \hpos{0.9826} & \hpos{0.0201} & \hpos{0.0573} & \hpos{\underline{0.0002}} & \hpos{\underline{0.0152}} & \hpos{\underline{0.985}} \\

PanFlowNet~\cite{yang2023panflownet} & 38.56 & 0.9619 & 1.28 & 1.09 & 0.8226 & 0.9593 & 0.0451 & 0.1001 & 0.0262 & 0.0724 & 0.904 \\
\rowcolor{BlueGreen!5}
\quad + PanBase ($\mathbf{s}_{\mathrm{pan}}$) & \hpos{40.12} & \hpos{0.9669} & \hpos{0.99} & \hpos{0.89} & \hpos{0.8366} & \hpos{0.9656} & \hpos{0.0366} & \hpos{0.0848} & \hpos{0.0094} & \hpos{0.0634} & \hpos{0.928} \\
\rowcolor{BlueGreen!5}
\quad + PanBase ($\mathbf{s}_{\mathrm{ms}}$) & \hpos{40.19} & \hpos{0.9674} & \hpos{0.97} & \hpos{0.87} & \hpos{0.8581} & \hpos{0.9640} & \hpos{0.0394} & \hpos{0.0916} & \hpos{0.0070} & \hpos{0.0598} & \hpos{0.934} \\

MSDDN~\cite{he2023msddn} & 38.72 & 0.9664 & 1.15 & 1.07 & 0.8667 & 0.9486 & 0.0330 & 0.0812 & 0.0142 & 0.0416 & 0.945 \\
\rowcolor{BlueGreen!5}
\quad + PanBase ($\mathbf{s}_{\mathrm{pan}}$) & \hpos{41.80} & \hpos{0.9769} & \hpos{0.79} & \hpos{0.75} & \hpos{0.8685} & \hpos{0.9773} & \hpos{0.0237} & \hpos{0.0635} & \hpos{0.0010} & \hneg{0.0429} & \hpos{0.956} \\
\rowcolor{BlueGreen!5}
\quad + PanBase ($\mathbf{s}_{\mathrm{ms}}$) & \hpos{41.86} & \hpos{0.9767} & \hpos{0.80} & \hpos{0.74} & \hpos{0.8686} & \hpos{0.9777} & \hpos{0.0252} & \hpos{0.0676} & \hpos{0.0005} & \hpos{0.0281} & \hpos{0.971} \\

HFIN~\cite{tan2024hfin} & 38.08 & 0.9716 & 1.07 & 1.13 & 0.8615 & 0.9282 & 0.0364 & 0.0882 & 0.0392 & 0.0538 & 0.909 \\
\rowcolor{BlueGreen!5}
\quad + PanBase ($\mathbf{s}_{\mathrm{pan}}$) & \hpos{40.95} & \hneg{0.9711} & \hpos{0.95} & \hpos{0.87} & \hneg{0.8451} & \hpos{0.9729} & \hpos{0.0260} & \hpos{0.0678} & \hpos{0.0019} & \hneg{0.0786} & \hpos{0.920} \\
\rowcolor{BlueGreen!5}
\quad + PanBase ($\mathbf{s}_{\mathrm{ms}}$) & \hpos{42.56} & \hpos{0.9793} & \hpos{0.80} & \hpos{0.69} & \hpos{0.8879} & \hpos{0.9796} & \hpos{0.0230} & \hpos{0.0675} & \hpos{0.0009} & \hpos{0.0328} & \hpos{0.966} \\

UTeRM~\cite{mai2024uterm} & 41.89 & 0.9759 & 0.84 & 0.75 & 0.8713 & 0.9771 & 0.0286 & 0.0748 & 0.0024 & 0.0574 & 0.940 \\
\rowcolor{BlueGreen!5}
\quad + PanBase ($\mathbf{s}_{\mathrm{pan}}$) & \hpos{42.56} & \hpos{0.9785} & \hpos{0.80} & \hpos{0.70} & \hpos{0.8827} & \hpos{0.9793} & \hpos{0.0247} & \hpos{0.0698} & \hpos{0.0009} & \hpos{0.0327} & \hpos{0.966} \\
\rowcolor{BlueGreen!5}
\quad + PanBase ($\mathbf{s}_{\mathrm{ms}}$) & \hpos{42.69} & \hpos{0.9794} & \hpos{0.79} & \hpos{0.70} & \hpos{0.8832} & \hpos{0.9808} & \hpos{0.0226} & \hpos{0.0634} & \hpos{\underline{0.0002}} & \hpos{0.0259} & \hpos{0.974} \\

BNNPan~\cite{hou2025bnnpan} & 42.12 & 0.9774 & 0.81 & 0.72 & 0.8813 & 0.9778 & 0.0271 & 0.0715 & 0.0036 & 0.0694 & 0.927 \\
\rowcolor{BlueGreen!5}
\quad + PanBase ($\mathbf{s}_{\mathrm{pan}}$) & \hpos{42.69} & \hpos{0.9791} & \hpos{0.77} & \hpos{0.68} & \hpos{0.8870} & \hpos{0.9799} & \hpos{0.0265} & \hpos{0.0697} & \hpos{0.0020} & \hpos{0.0557} & \hpos{0.942} \\
\rowcolor{BlueGreen!5}
\quad + PanBase ($\mathbf{s}_{\mathrm{ms}}$) & -- & -- & -- & -- & -- & -- & -- & -- & -- & -- & -- \\

U-Know-DiffPAN~\cite{kim2025uknowdiffpan} & 43.24 & 0.9824 & 0.68 & 0.64 & 0.8924 & 0.9833 & 0.0233 & 0.0594 & 0.0012 & 0.0325 & 0.966 \\
\rowcolor{BlueGreen!5}
\quad + PanBase ($\mathbf{s}_{\mathrm{pan}}$) & \hpos{44.05} & \hpos{0.9845} & \hpos{0.64} & \hpos{0.59} & \hpos{0.9054} & \hpos{0.9853} & \hpos{0.0213} & \hpos{0.0552} & \hpos{0.0005} & \hpos{0.0235} & \hpos{0.976} \\
\rowcolor{BlueGreen!5}
\quad + PanBase ($\mathbf{s}_{\mathrm{ms}}$) & -- & -- & -- & -- & -- & -- & -- & -- & -- & -- & -- \\

PAN-Crafter~\cite{do2025pancrafter} & 44.33 & 0.9857 & 0.65 & 0.57 & 0.9192 & 0.9860 & 0.0205 & 0.0565 & 0.0041 & 0.0254 & 0.971 \\
\rowcolor{BlueGreen!5}
\quad + PanBase ($\mathbf{s}_{\mathrm{pan}}$) & \hpos{\underline{44.80}} & \hpos{\underline{0.9868}} & \hpos{\underline{0.62}} & \hpos{\underline{0.53}} & \hpos{\underline{0.9224}} & \hpos{\underline{0.9872}} & \hpos{\textbf{0.0167}} & \hpos{\textbf{0.0524}} & \hpos{0.0013} & \hpos{0.0198} & \hpos{0.979} \\
\rowcolor{BlueGreen!5}
\quad + PanBase ($\mathbf{s}_{\mathrm{ms}}$) & \hpos{\textbf{45.00}} & \hpos{\textbf{0.9872}} & \hpos{\textbf{0.61}} & \hpos{\textbf{0.52}} & \hpos{\textbf{0.9235}} & \hpos{\textbf{0.9875}} & \hpos{\underline{0.0174}} & \hpos{\underline{0.0526}} & \hpos{0.0014} & \hpos{0.0179} & \hpos{0.981} \\

PINO~\cite{liu2025pino} & 42.48 & 0.9808 & 0.91 & 0.72 & 0.9023 & 0.9811 & 0.0300 & 0.0717 & 0.0024 & 0.0185 & 0.979 \\
\rowcolor{BlueGreen!5}
\quad + PanBase ($\mathbf{s}_{\mathrm{pan}}$) & \hpos{44.20} & \hpos{0.9852} & \hpos{0.69} & \hpos{0.57} & \hpos{0.9136} & \hpos{0.9855} & \hpos{0.0204} & \hpos{0.0569} & \hpos{0.0011} & \hpos{\textbf{0.0125}} & \hpos{\textbf{0.986}} \\
\rowcolor{BlueGreen!5}
\quad + PanBase ($\mathbf{s}_{\mathrm{ms}}$) & \hpos{44.66} & \hpos{0.9860} & \hpos{0.65} & \hpos{0.54} & \hpos{0.9172} & \hpos{0.9864} & \hpos{0.0205} & \hpos{0.0597} & \hpos{0.0004} & \hpos{0.0161} & \hpos{0.984} \\

PanDiT~\cite{fang2025pandit} & 42.00 & 0.9767 & 0.80 & 0.74 & 0.8690 & 0.9786 & 0.0245 & 0.0698 & 0.0009 & 0.0449 & 0.954 \\
\rowcolor{BlueGreen!5}
\quad + PanBase ($\mathbf{s}_{\mathrm{pan}}$) & \hpos{42.16} & \hpos{0.9769} & 0.80 & \hpos{0.73} & \hpos{0.8692} & \hpos{0.9791} & \hpos{0.0236} & \hpos{0.0691} & \hpos{\textbf{0.0001}} & \hpos{0.0184} & \hpos{0.982} \\
\rowcolor{BlueGreen!5}
\quad + PanBase ($\mathbf{s}_{\mathrm{ms}}$) & \hpos{42.34} & \hpos{0.9777} & 0.80 & \hpos{0.71} & \hpos{0.8727} & \hpos{0.9794} & \hpos{0.0240} & \hpos{0.0674} & \hpos{\textbf{0.0001}} & \hpos{0.0220} & \hpos{0.978} \\

MMMamba~\cite{wang2026mmmamba} & 39.68 & 0.9484 & 1.23 & 1.11 & 0.8040 & 0.9457 & 0.0582 & 0.0882 & 0.0201 & 0.0535 & 0.928 \\
\rowcolor{BlueGreen!5}
\quad + PanBase ($\mathbf{s}_{\mathrm{pan}}$) & \hpos{42.72} & \hpos{0.9791} & \hpos{0.84} & \hpos{0.72} & \hpos{0.8864} & \hpos{0.9806} & \hpos{0.0193} & \hpos{0.0574} & \hpos{0.0063} & \hpos{0.0325} & \hpos{0.961} \\
\rowcolor{BlueGreen!5}
\quad + PanBase ($\mathbf{s}_{\mathrm{ms}}$) & -- & -- & -- & -- & -- & -- & -- & -- & -- & -- & -- \\

TS-Pan~\cite{li2026tspan} & 41.94 & 0.9761 & 0.81 & 0.74 & 0.8567 & 0.9742 & 0.0273 & 0.1073 & 0.0133 & 0.0430 & 0.944 \\
\rowcolor{BlueGreen!5}
\quad + PanBase ($\mathbf{s}_{\mathrm{pan}}$) & \hpos{44.34} & \hpos{0.9854} & \hpos{0.64} & \hpos{0.56} & \hpos{0.9182} & \hpos{0.9857} & \hpos{0.0250} & \hpos{0.0590} & \hpos{0.0095} & \hneg{0.0600} & \hneg{0.931} \\
\rowcolor{BlueGreen!5}
\quad + PanBase ($\mathbf{s}_{\mathrm{ms}}$) & \hpos{44.56} & \hpos{0.9859} & \hpos{0.63} & \hpos{0.55} & \hpos{0.9170} & \hpos{0.9858} & \hpos{0.0204} & \hpos{0.0548} & \hpos{0.0004} & \hpos{0.0210} & \hpos{0.979} \\

\midrule
Mean change ($\mathbf{s}_{\mathrm{pan}}$) & \hpos{+1.75} & \hpos{+0.0072} & \hpos{$-$0.17} & \hpos{$-$0.18} & \hpos{+0.0218} & \hpos{+0.0132} & \hpos{$-$0.0096} & \hpos{$-$0.0157} & \hpos{$-$0.0080} & \hpos{$-$0.0090} & \hpos{+0.017} \\
Mean change ($\mathbf{s}_{\mathrm{ms}}$) & \hpos{+2.26} & \hpos{+0.0087} & \hpos{$-$0.20} & \hpos{$-$0.23} & \hpos{+0.0348} & \hpos{+0.0152} & \hpos{$-$0.0095} & \hpos{$-$0.0168} & \hpos{$-$0.0101} & \hpos{$-$0.0235} & \hpos{+0.033} \\
\bottomrule
\end{tabular}}
\end{table*}

\begin{table*}[!t]
\small
\centering
\sethlcolor{BlueGreen!5}
\caption{\textbf{PanBase results on unseen WV2.} 
\hl{Highlighted rows} use PanBase as the residual base and an additional input. \hpos{Green} and \hneg{red} indicate gains and drops relative to each original backbone.
\textbf{Bold} and \underline{underline} mark the best and second-best scores per metric. Means exclude missing entries (--).
WV3-trained models transfer without fine-tuning. The standalone bases use WV3 coefficients with the WV2 sensor operator.}
\vspace{-0.2cm}
\label{tab:panbase_full_wv2}
\setlength{\tabcolsep}{6pt}
\renewcommand{\arraystretch}{1.08}
\resizebox{\textwidth}{!}{%
\begin{tabular}{l cccccccc ccc}
\toprule
\multirow{2}{*}{\textbf{Method}} & \multicolumn{8}{c}{\textbf{Reduced resolution (HRMS reference)}} & \multicolumn{3}{c}{\textbf{Full resolution (no reference)}} \\
\cmidrule(lr){2-9}\cmidrule(l){10-12}
& PSNR$\uparrow$ & SSIM$\uparrow$ & SAM$\downarrow$ & ERGAS$\downarrow$ & SCC$\uparrow$ & $Q2^{n}\!\uparrow$ & LPIPS$\downarrow$ & DISTS$\downarrow$ & $D_{\lambda}\!\downarrow$ & $D_s\!\downarrow$ & HQNR$\uparrow$ \\
\midrule
PanNet~\cite{yang2017pannet} & 28.51 & 0.8095 & 5.81 & 4.64 & 0.6885 & 0.8230 & 0.2294 & 0.1866 & 0.0169 & 0.0158 & \underline{0.968} \\
\rowcolor{BlueGreen!5}
\quad + PanBase ($\mathbf{s}_{\mathrm{pan}}$) & \hpos{29.10} & \hpos{0.8277} & \hpos{5.52} & \hpos{4.34} & \hneg{0.6765} & \hpos{0.8360} & \hpos{0.1885} & \hpos{0.1686} & \hpos{0.0054} & \hneg{0.0371} & \hneg{0.958} \\
\rowcolor{BlueGreen!5}
\quad + PanBase ($\mathbf{s}_{\mathrm{ms}}$) & \hpos{29.40} & \hpos{0.8384} & \hpos{5.48} & \hpos{4.17} & \hpos{0.7036} & \hpos{0.8434} & \hpos{0.1525} & \hpos{0.1550} & \hpos{0.0061} & \hneg{0.0312} & \hneg{0.963} \\

LAGConv~\cite{jin2022lagconv} & 28.71 & 0.8195 & 5.61 & 4.55 & 0.6603 & 0.8317 & 0.2178 & 0.1769 & 0.0244 & 0.0275 & 0.949 \\
\rowcolor{BlueGreen!5}
\quad + PanBase ($\mathbf{s}_{\mathrm{pan}}$) & \hpos{29.17} & \hpos{0.8341} & \hpos{5.48} & \hpos{4.34} & \hneg{0.6327} & \hpos{0.8410} & \hpos{0.2059} & \hpos{0.1745} & \hpos{0.0120} & \hneg{0.0857} & \hneg{0.903} \\
\rowcolor{BlueGreen!5}
\quad + PanBase ($\mathbf{s}_{\mathrm{ms}}$) & \hpos{29.07} & \hpos{0.8324} & \hpos{5.51} & \hpos{4.36} & \hneg{0.6522} & \hpos{0.8434} & \hpos{0.2038} & \hpos{0.1715} & \hpos{0.0129} & \hneg{0.0512} & \hneg{0.936} \\

MutInf~\cite{zhou2022mutinf} & 28.87 & 0.8399 & 5.47 & 4.49 & 0.6737 & 0.8306 & 0.1602 & 0.1388 & 0.0634 & 0.0371 & 0.902 \\
\rowcolor{BlueGreen!5}
\quad + PanBase ($\mathbf{s}_{\mathrm{pan}}$) & \hpos{29.01} & \hneg{0.8234} & \hpos{5.28} & \hpos{4.42} & \hneg{0.6073} & \hpos{0.8356} & \hneg{0.2412} & \hneg{0.1857} & \hpos{0.0083} & \hneg{0.0521} & \hpos{0.940} \\
\rowcolor{BlueGreen!5}
\quad + PanBase ($\mathbf{s}_{\mathrm{ms}}$) & \hpos{29.23} & \hneg{0.8351} & \hpos{5.18} & \hpos{4.31} & \hneg{0.6364} & \hpos{0.8417} & \hneg{0.2075} & \hneg{0.1618} & \hpos{0.0095} & \hneg{0.0558} & \hpos{0.935} \\

MDCUN~\cite{yang2022mdcun} & 28.96 & 0.8263 & 5.40 & 4.45 & 0.6937 & 0.8368 & 0.2318 & 0.1892 & 0.0156 & 0.0473 & 0.938 \\
\rowcolor{BlueGreen!5}
\quad + PanBase ($\mathbf{s}_{\mathrm{pan}}$) & \hpos{30.05} & \hpos{0.8638} & \hpos{5.28} & \hpos{3.84} & \hpos{0.7206} & \hpos{0.8687} & \hpos{0.1579} & \hpos{0.1377} & \hpos{0.0029} & \hpos{0.0359} & \hpos{0.961} \\
\rowcolor{BlueGreen!5}
\quad + PanBase ($\mathbf{s}_{\mathrm{ms}}$) & \hneg{28.79} & \hneg{0.8139} & \hneg{5.58} & \hneg{4.53} & \hneg{0.6160} & \hneg{0.8247} & \hneg{0.2632} & \hpos{0.1891} & \hpos{0.0096} & \hneg{0.0501} & \hpos{0.941} \\

PanFlowNet~\cite{yang2023panflownet} & 27.78 & 0.8060 & 6.17 & 5.16 & 0.6582 & 0.7883 & 0.2700 & 0.1887 & 0.0745 & 0.0609 & 0.869 \\
\rowcolor{BlueGreen!5}
\quad + PanBase ($\mathbf{s}_{\mathrm{pan}}$) & \hpos{28.19} & \hneg{0.8026} & \hpos{5.96} & \hpos{4.84} & \hneg{0.6355} & \hpos{0.8085} & \hpos{0.2491} & \hneg{0.1940} & \hpos{0.0256} & \hpos{0.0393} & \hpos{0.936} \\
\rowcolor{BlueGreen!5}
\quad + PanBase ($\mathbf{s}_{\mathrm{ms}}$) & \hpos{28.22} & \hneg{0.7970} & \hpos{5.80} & \hpos{4.86} & \hneg{0.6202} & \hpos{0.8067} & \hpos{0.2595} & \hneg{0.1949} & \hpos{0.0370} & \hpos{0.0545} & \hpos{0.912} \\

MSDDN~\cite{he2023msddn} & 28.65 & 0.8304 & 5.73 & 4.59 & 0.6770 & 0.8235 & 0.1853 & 0.1715 & 0.0426 & 0.0279 & 0.931 \\
\rowcolor{BlueGreen!5}
\quad + PanBase ($\mathbf{s}_{\mathrm{pan}}$) & \hpos{28.82} & \hneg{0.8237} & \hpos{5.47} & \hpos{4.51} & \hneg{0.6085} & \hpos{0.8338} & \hneg{0.2094} & \hneg{0.1734} & \hpos{0.0101} & \hneg{0.0710} & \hneg{0.920} \\
\rowcolor{BlueGreen!5}
\quad + PanBase ($\mathbf{s}_{\mathrm{ms}}$) & \hpos{28.68} & \hneg{0.8270} & \hpos{5.57} & \hpos{4.55} & \hneg{0.6229} & \hpos{0.8367} & \hpos{0.1807} & \hneg{0.1721} & \hpos{0.0138} & \hneg{0.0881} & \hneg{0.899} \\

HFIN~\cite{tan2024hfin} & 28.39 & 0.8135 & 5.87 & 4.75 & 0.6967 & 0.8123 & 0.2461 & 0.1907 & 0.0504 & 0.0488 & 0.903 \\
\rowcolor{BlueGreen!5}
\quad + PanBase ($\mathbf{s}_{\mathrm{pan}}$) & \hpos{29.47} & \hpos{0.8413} & \hpos{5.23} & \hpos{4.18} & \hneg{0.6799} & \hpos{0.8490} & \hpos{0.2020} & \hpos{0.1659} & \hpos{0.0076} & \hpos{0.0411} & \hpos{0.952} \\
\rowcolor{BlueGreen!5}
\quad + PanBase ($\mathbf{s}_{\mathrm{ms}}$) & \hpos{28.68} & \hneg{0.8078} & \hpos{5.36} & \hpos{4.63} & \hneg{0.6101} & \hpos{0.8239} & \hneg{0.2691} & \hneg{0.2036} & \hpos{0.0137} & \hneg{0.0650} & \hpos{0.922} \\

UTeRM~\cite{mai2024uterm} & 28.85 & 0.8347 & 5.73 & 4.48 & 0.7300 & 0.8350 & 0.2173 & 0.1730 & 0.0319 & \underline{0.0144} & 0.954 \\
\rowcolor{BlueGreen!5}
\quad + PanBase ($\mathbf{s}_{\mathrm{pan}}$) & \hpos{29.13} & \hpos{0.8366} & \hpos{5.62} & \hpos{4.33} & \hneg{0.6739} & \hpos{0.8447} & \hneg{0.2311} & \hneg{0.1844} & \hpos{0.0300} & \hneg{0.0492} & \hneg{0.922} \\
\rowcolor{BlueGreen!5}
\quad + PanBase ($\mathbf{s}_{\mathrm{ms}}$) & \hpos{29.45} & \hpos{0.8514} & \hpos{5.50} & \hpos{4.18} & \hneg{0.7198} & \hpos{0.8494} & \hpos{0.1994} & \hpos{0.1694} & \hpos{0.0225} & \hneg{0.0326} & \hneg{0.946} \\

BNNPan~\cite{hou2025bnnpan} & 27.35 & 0.7740 & 7.10 & 5.40 & 0.4945 & 0.7837 & 0.2420 & 0.2196 & 0.1584 & 0.1241 & 0.755 \\
\rowcolor{BlueGreen!5}
\quad + PanBase ($\mathbf{s}_{\mathrm{pan}}$) & \hpos{29.69} & \hpos{0.8604} & \hpos{5.46} & \hpos{4.00} & \hpos{0.7059} & \hpos{0.8614} & \hpos{0.1358} & \hpos{0.1308} & \hpos{0.0244} & \hpos{0.0274} & \hpos{0.949} \\
\rowcolor{BlueGreen!5}
\quad + PanBase ($\mathbf{s}_{\mathrm{ms}}$) & -- & -- & -- & -- & -- & -- & -- & -- & -- & -- & -- \\

U-Know-DiffPAN~\cite{kim2025uknowdiffpan} & 29.20 & 0.8350 & 5.22 & 4.28 & 0.6857 & 0.8466 & 0.2267 & 0.1817 & 0.0154 & 0.0192 & 0.966 \\
\rowcolor{BlueGreen!5}
\quad + PanBase ($\mathbf{s}_{\mathrm{pan}}$) & \hpos{29.80} & \hpos{0.8539} & \hpos{5.07} & \hpos{4.03} & \hpos{0.6895} & \hpos{0.8595} & \hpos{0.2165} & \hpos{0.1763} & \hpos{0.0092} & \hneg{0.0440} & \hneg{0.947} \\
\rowcolor{BlueGreen!5}
\quad + PanBase ($\mathbf{s}_{\mathrm{ms}}$) & \hpos{30.04} & \hpos{0.8614} & \hpos{4.98} & \hpos{3.89} & \hpos{0.7314} & \hpos{0.8650} & \hpos{0.2003} & \hpos{0.1664} & \hpos{0.0066} & \hneg{0.0341} & \hneg{0.960} \\

PAN-Crafter~\cite{do2025pancrafter} & 29.04 & 0.8267 & 5.17 & 4.42 & 0.6698 & 0.8378 & 0.2783 & 0.1977 & 0.0236 & 0.0286 & 0.949 \\
\rowcolor{BlueGreen!5}
\quad + PanBase ($\mathbf{s}_{\mathrm{pan}}$) & \hpos{29.90} & \hpos{0.8551} & \hpos{5.08} & \hpos{3.94} & \hpos{0.7389} & \hpos{0.8621} & \hpos{0.2092} & \hpos{0.1735} & \hpos{0.0133} & \hneg{0.0505} & \hneg{0.937} \\
\rowcolor{BlueGreen!5}
\quad + PanBase ($\mathbf{s}_{\mathrm{ms}}$) & \hneg{28.88} & \hneg{0.8179} & \hneg{5.26} & \hneg{4.49} & \hneg{0.6259} & \hneg{0.8326} & \hneg{0.2939} & \hneg{0.2035} & \hpos{0.0105} & \hneg{0.0568} & \hneg{0.933} \\

PINO~\cite{liu2025pino} & 29.76 & 0.8584 & 5.11 & 4.00 & 0.7295 & 0.8595 & 0.2154 & 0.1706 & 0.0270 & 0.0273 & 0.946 \\
\rowcolor{BlueGreen!5}
\quad + PanBase ($\mathbf{s}_{\mathrm{pan}}$) & \hpos{30.14} & \hpos{0.8636} & \hpos{\underline{4.87}} & \hpos{3.84} & \hpos{\underline{0.7515}} & \hpos{0.8675} & \hpos{0.2086} & \hpos{0.1613} & \hpos{0.0070} & \hneg{0.0400} & \hpos{0.953} \\
\rowcolor{BlueGreen!5}
\quad + PanBase ($\mathbf{s}_{\mathrm{ms}}$) & \hpos{\textbf{30.43}} & \hpos{0.8718} & \hpos{\textbf{4.83}} & \hpos{3.73} & \hpos{\textbf{0.7517}} & \hpos{0.8732} & \hpos{0.1993} & \hpos{0.1624} & \hpos{0.0069} & \hneg{0.0522} & \hneg{0.941} \\

PanDiT~\cite{fang2025pandit} & 29.15 & 0.8310 & 5.33 & 4.32 & 0.6904 & 0.8383 & 0.2087 & 0.1748 & 0.0098 & \textbf{0.0123} & \textbf{0.978} \\
\rowcolor{BlueGreen!5}
\quad + PanBase ($\mathbf{s}_{\mathrm{pan}}$) & \hpos{30.17} & \hpos{0.8680} & \hpos{5.15} & \hpos{3.80} & \hpos{0.7155} & \hpos{0.8713} & \hpos{0.1510} & \hpos{0.1322} & \hpos{\underline{0.0009}} & \hneg{0.0368} & \hneg{0.962} \\
\rowcolor{BlueGreen!5}
\quad + PanBase ($\mathbf{s}_{\mathrm{ms}}$) & \hpos{\underline{30.37}} & \hpos{\underline{0.8724}} & \hpos{5.11} & \hpos{\underline{3.69}} & \hpos{0.7325} & \hpos{\underline{0.8743}} & \hpos{0.1374} & \hpos{\underline{0.1253}} & \hpos{0.0010} & \hneg{0.0375} & \hneg{0.962} \\

MMMamba~\cite{wang2026mmmamba} & 28.89 & 0.8224 & 5.46 & 4.45 & 0.6199 & 0.8372 & 0.2007 & 0.1785 & 0.0129 & 0.0554 & 0.932 \\
\rowcolor{BlueGreen!5}
\quad + PanBase ($\mathbf{s}_{\mathrm{pan}}$) & \hneg{28.88} & \hneg{0.8197} & \hpos{5.41} & \hneg{4.47} & \hneg{0.5611} & \hneg{0.8311} & \hneg{0.2298} & \hpos{0.1752} & \hpos{0.0093} & \hneg{0.0920} & \hneg{0.899} \\
\rowcolor{BlueGreen!5}
\quad + PanBase ($\mathbf{s}_{\mathrm{ms}}$) & \hneg{27.96} & \hneg{0.7975} & \hneg{5.84} & \hneg{5.00} & \hneg{0.4591} & \hneg{0.8151} & \hpos{0.1976} & \hneg{0.1960} & \hpos{0.0115} & \hneg{0.1128} & \hneg{0.877} \\

TS-Pan~\cite{li2026tspan} & 29.59 & 0.8473 & 5.12 & 4.10 & 0.6941 & 0.8559 & 0.2025 & 0.1685 & 0.0104 & 0.0705 & 0.920 \\
\rowcolor{BlueGreen!5}
\quad + PanBase ($\mathbf{s}_{\mathrm{pan}}$) & \hpos{30.10} & \hpos{0.8641} & \hneg{5.13} & \hpos{3.84} & \hneg{0.6932} & \hpos{0.8688} & \hpos{0.1760} & \hpos{0.1550} & \hpos{0.0030} & \hpos{0.0672} & \hpos{0.930} \\
\rowcolor{BlueGreen!5}
\quad + PanBase ($\mathbf{s}_{\mathrm{ms}}$) & \hpos{30.15} & \hpos{0.8625} & \hpos{5.01} & \hpos{3.82} & \hneg{0.6805} & \hpos{0.8672} & \hpos{0.1644} & \hneg{0.2004} & \hpos{0.0055} & \hpos{0.0596} & \hpos{0.935} \\

\midrule
Mean change ($\mathbf{s}_{\mathrm{pan}}$) & \hpos{+0.66} & \hpos{+0.0175} & \hpos{$-$0.29} & \hpos{$-$0.36} & \hpos{+0.0019} & \hpos{+0.0199} & \hpos{$-$0.0213} & \hpos{$-$0.0146} & \hpos{$-$0.0272} & \hneg{+0.0101} & \hpos{+0.014} \\
Mean change ($\mathbf{s}_{\mathrm{ms}}$) & \hpos{+0.36} & \hpos{+0.0061} & \hpos{$-$0.16} & \hpos{$-$0.18} & \hneg{$-$0.0289} & \hpos{+0.0101} & \hpos{$-$0.0116} & \hpos{$-$0.0011} & \hpos{$-$0.0180} & \hneg{+0.0206} & \hneg{$-$0.003} \\
\midrule
Standalone PanBase ($\mathbf{s}_{\mathrm{pan}}$) & 30.12 & 0.8691 & 5.49 & 3.77 & 0.7319 & 0.8741 & \underline{0.1323} & 0.1263 & \textbf{0.0005} & 0.0404 & 0.959 \\
Standalone PanBase ($\mathbf{s}_{\mathrm{ms}}$) & 30.34 & \textbf{0.8747} & 5.48 & \textbf{3.65} & 0.7503 & \textbf{0.8784} & \textbf{0.1203} & \textbf{0.1212} & \textbf{0.0005} & 0.0517 & 0.948 \\
\bottomrule
\end{tabular}}
\end{table*}

\begin{table*}[t]
\small
\centering
\caption{\textbf{Classical pan-sharpening versus standalone PanBase on WV3.}
All methods operate without neural networks. PanBase coefficients are calibrated once using HRMS training pairs.
CS/MRA/VO denote component substitution, multiresolution analysis, and variational optimization.
Bold and underline indicate the best and second-best values per metric.}
\vspace{-0.2cm}
\label{tab:h_classical_wv3}
\setlength{\tabcolsep}{6pt}
\renewcommand{\arraystretch}{1.1}
\resizebox{\textwidth}{!}{%
\begin{tabular}{cl cccccccc ccc}
\toprule
\multirow{2}{*}{\textbf{Type}}
& \multirow{2}{*}{\textbf{Method}}
& \multicolumn{8}{c}{\textbf{Reduced resolution (RR)}}
& \multicolumn{3}{c}{\textbf{Full resolution (FR)}} \\
\cmidrule(lr){3-10}\cmidrule(l){11-13}
& & PSNR$\uparrow$ & SSIM$\uparrow$ & SAM$\downarrow$ & ERGAS$\downarrow$
  & SCC$\uparrow$ & $Q2^{n}\uparrow$ & LPIPS$\downarrow$ & DISTS$\downarrow$
  & $D_{\lambda}\downarrow$ & $D_s\downarrow$ & HQNR$\uparrow$ \\
\midrule
-- & LRMS$\uparrow$ (interpolated)~\cite{aiazzi2002context}
  & 27.12 & 0.7130 & 5.80 & 7.16 & 0.1678 & 0.6272 & 0.5881 & 0.3175
  & 0.0155 & 0.0772 & 0.909 \\
\midrule
\multirow{4}{*}{CS} & BT-H~\cite{lolli2017haze}
  & 31.22 & 0.8947 & 4.92 & 4.58 & \underline{0.6614} & 0.8327 & 0.1650 & 0.1575
  & 0.0308 & 0.0811 & 0.891 \\
& BDSD-PC~\cite{vivone2019bdsdpc}
  & 30.96 & 0.8890 & 5.43 & 4.70 & 0.6400 & 0.8296 & 0.1591 & 0.1502
  & 0.0360 & 0.0727 & 0.895 \\
& GSA~\cite{aiazzi2007gsa}
  & 30.78 & 0.8821 & 5.37 & 4.77 & 0.6387 & 0.8273 & 0.1704 & 0.1578
  & 0.0271 & 0.0794 & 0.896 \\
& PRACS~\cite{choi2011pracs}
  & 29.95 & 0.8559 & 5.58 & 5.25 & 0.6285 & 0.7899 & 0.2056 & 0.1704
  & 0.0206 & 0.0453 & 0.935 \\
\midrule
\multirow{4}{*}{MRA} & AWLP~\cite{otazu2005awlp}
  & 31.21 & 0.8902 & 5.23 & 4.66 & 0.6492 & 0.8352 & 0.1753 & 0.1646
  & 0.0105 & 0.0524 & 0.938 \\
& MTF-GLP-FS~\cite{vivone2018fullscale}
  & 30.91 & 0.8837 & 5.32 & 4.70 & 0.6386 & 0.8335 & 0.1735 & 0.1602
  & 0.0089 & 0.0627 & 0.929 \\
& MTF-GLP-HPM~\cite{aiazzi2006mtf,alparone2017intersensor}
  & 31.38 & 0.8960 & 5.13 & 4.50 & 0.6547 & 0.8384 & 0.1593 & 0.1531
  & 0.0104 & 0.0702 & 0.920 \\
& MF~\cite{restaino2016morphological}
  & 30.56 & 0.8786 & 5.26 & 4.86 & 0.6481 & 0.8220 & 0.1935 & 0.1845
  & 0.0171 & 0.0653 & 0.919 \\
\midrule
\multirow{4}{*}{VO} & TV~\cite{palsson2014tv}
  & 30.48 & 0.8801 & 5.69 & 4.86 & 0.6330 & 0.7955 & 0.1510 & 0.1546
  & 0.0095 & \underline{0.0386} & \underline{0.952} \\
& FE-HPM~\cite{vivone2015semiblind}
  & 30.87 & 0.8821 & 5.23 & 4.67 & 0.6549 & 0.8325 & 0.1791 & 0.1731
  & 0.0109 & 0.0667 & 0.923 \\
& PWMBF~\cite{palsson2015pwmbf}
  & 30.07 & 0.8699 & 5.89 & 5.14 & 0.6359 & 0.8056 & 0.1968 & 0.1779
  & 0.0330 & 0.0695 & 0.900 \\
& LRTCFPan~\cite{wu2023lrtcfpan}
  & 31.74 & 0.9024 & 4.74 & 4.32 & 0.6452 & 0.8467 & 0.1720 & 0.1521
  & 0.0054 & 0.0525 & 0.942 \\
\midrule
\rowcolor{BlueGreen!5} & \textbf{PanBase} ($\mathbf{s}_{\mathrm{pan}}$)
  & \underline{32.28} & \underline{0.9147} & \underline{4.65} & \underline{4.07} & 0.6608 & \underline{0.8561} & \underline{0.0806} & \underline{0.1136}
  & \textbf{0.0005} & \textbf{0.0365} & \textbf{0.963} \\
\rowcolor{BlueGreen!5}\multirow{-2}{*}{\textbf{Ours}} & \textbf{PanBase} ($\mathbf{s}_{\mathrm{ms}}$)
  & \textbf{33.88} & \textbf{0.9402} & \textbf{4.62} & \textbf{3.30} & \textbf{0.7917} & \textbf{0.8830} & \textbf{0.0663} & \textbf{0.1014}
  & \underline{0.0006} & 0.0841 & 0.915 \\
\bottomrule
\end{tabular}}
\end{table*}

\begin{table*}[t]
\small
\centering
\caption{\textbf{Classical pan-sharpening versus standalone PanBase on QB.}
All methods operate without neural networks. PanBase coefficients are calibrated once using HRMS training pairs.
CS/MRA/VO denote component substitution, multiresolution analysis, and variational optimization.
Bold and underline indicate the best and second-best values per metric.}
\vspace{-0.2cm}
\label{tab:h_classical_qb}
\setlength{\tabcolsep}{6pt}
\renewcommand{\arraystretch}{1.1}
\resizebox{\textwidth}{!}{%
\begin{tabular}{cl cccccccc ccc}
\toprule
\multirow{2}{*}{\textbf{Type}}
& \multirow{2}{*}{\textbf{Method}}
& \multicolumn{8}{c}{\textbf{Reduced resolution (RR)}}
& \multicolumn{3}{c}{\textbf{Full resolution (FR)}} \\
\cmidrule(lr){3-10}\cmidrule(l){11-13}
& & PSNR$\uparrow$ & SSIM$\uparrow$ & SAM$\downarrow$ & ERGAS$\downarrow$
  & SCC$\uparrow$ & $Q2^{n}\uparrow$ & LPIPS$\downarrow$ & DISTS$\downarrow$
  & $D_{\lambda}\downarrow$ & $D_s\downarrow$ & HQNR$\uparrow$ \\
\midrule
-- & LRMS$\uparrow$ (interpolated)~\cite{aiazzi2002context}
  & 28.06 & 0.6915 & 8.44 & 11.82 & 0.1681 & 0.5839 & 0.6627 & 0.3384
  & 0.0273 & 0.1456 & 0.831 \\
\midrule
\multirow{4}{*}{CS} & BT-H~\cite{lolli2017haze}
  & 32.09 & 0.8695 & 7.23 & 7.34 & 0.6459 & 0.8386 & 0.1587 & 0.1805
  & 0.1143 & 0.2273 & 0.685 \\
& BDSD-PC~\cite{vivone2019bdsdpc}
  & 31.89 & 0.8603 & 8.09 & 7.51 & 0.6298 & 0.8305 & 0.1766 & 0.1843
  & 0.1020 & 0.1659 & 0.750 \\
& GSA~\cite{aiazzi2007gsa}
  & 31.75 & 0.8554 & 8.10 & 7.62 & 0.6293 & 0.8269 & 0.1942 & 0.1895
  & 0.1120 & 0.2270 & 0.687 \\
& PRACS~\cite{choi2011pracs}
  & 31.00 & 0.8347 & 8.19 & 8.32 & 0.6290 & 0.7889 & 0.2244 & 0.1950
  & 0.0651 & 0.1649 & 0.781 \\
\midrule
\multirow{4}{*}{MRA} & AWLP~\cite{otazu2005awlp}
  & 31.89 & 0.8575 & 8.19 & 7.58 & 0.6386 & 0.8292 & 0.1814 & 0.1735
  & 0.0238 & 0.1427 & 0.837 \\
& MTF-GLP-FS~\cite{vivone2018fullscale}
  & 32.02 & 0.8613 & 7.79 & 7.37 & 0.6304 & 0.8349 & 0.1879 & 0.1843
  & 0.0220 & 0.1518 & 0.830 \\
& MTF-GLP-HPM~\cite{aiazzi2006mtf,alparone2017intersensor}
  & 32.24 & 0.8696 & 7.79 & 7.22 & 0.6441 & 0.8454 & 0.1656 & 0.1765
  & 0.0262 & 0.1519 & 0.826 \\
& MF~\cite{restaino2016morphological}
  & 31.47 & 0.8458 & 7.93 & 7.87 & 0.6364 & 0.8185 & 0.2059 & 0.2089
  & 0.0348 & 0.1397 & 0.831 \\
\midrule
\multirow{4}{*}{VO} & TV~\cite{palsson2014tv}
  & 31.59 & 0.8561 & 7.51 & 7.69 & 0.6255 & 0.8214 & 0.2202 & 0.1831
  & 0.0254 & 0.1020 & 0.875 \\
& FE-HPM~\cite{vivone2015semiblind}
  & 31.71 & 0.8495 & 7.95 & 7.65 & 0.6437 & 0.8266 & 0.1943 & 0.2010
  & 0.0266 & 0.1517 & 0.826 \\
& PWMBF~\cite{palsson2015pwmbf}
  & 31.00 & 0.8328 & 8.87 & 8.28 & 0.6266 & 0.7892 & 0.2303 & 0.2155
  & 0.0686 & 0.1604 & 0.782 \\
& LRTCFPan~\cite{wu2023lrtcfpan}
  & 32.57 & 0.8749 & 7.19 & 6.93 & 0.6349 & 0.8553 & 0.1780 & 0.1725
  & 0.0042 & \underline{0.0701} & \underline{0.926} \\
\midrule
\rowcolor{BlueGreen!5} & \textbf{PanBase} ($\mathbf{s}_{\mathrm{pan}}$)
  & \underline{33.10} & \underline{0.8909} & \underline{6.56} & \underline{6.53} & \underline{0.6470} & \underline{0.8707} & \underline{0.1157} & \underline{0.1495}
  & \underline{0.0033} & \textbf{0.0525} & \textbf{0.944} \\
\rowcolor{BlueGreen!5}\multirow{-2}{*}{\textbf{Ours}} & \textbf{PanBase} ($\mathbf{s}_{\mathrm{ms}}$)
  & \textbf{36.05} & \textbf{0.9311} & \textbf{6.11} & \textbf{4.63} & \textbf{0.8098} & \textbf{0.9130} & \textbf{0.0887} & \textbf{0.1399}
  & \textbf{0.0028} & 0.2953 & 0.703 \\
\bottomrule
\end{tabular}}
\end{table*}

\begin{table*}[t]
\small
\centering
\caption{\textbf{Classical pan-sharpening versus standalone PanBase on GF2.}
All methods operate without neural networks. PanBase coefficients are calibrated once using HRMS training pairs.
CS/MRA/VO denote component substitution, multiresolution analysis, and variational optimization.
Bold and underline indicate the best and second-best values per metric.}
\vspace{-0.2cm}
\label{tab:h_classical_gf2}
\setlength{\tabcolsep}{6pt}
\renewcommand{\arraystretch}{1.1}
\resizebox{\textwidth}{!}{%
\begin{tabular}{cl cccccccc ccc}
\toprule
\multirow{2}{*}{\textbf{Type}}
& \multirow{2}{*}{\textbf{Method}}
& \multicolumn{8}{c}{\textbf{Reduced resolution (RR)}}
& \multicolumn{3}{c}{\textbf{Full resolution (FR)}} \\
\cmidrule(lr){3-10}\cmidrule(l){11-13}
& & PSNR$\uparrow$ & SSIM$\uparrow$ & SAM$\downarrow$ & ERGAS$\downarrow$
  & SCC$\uparrow$ & $Q2^{n}\uparrow$ & LPIPS$\downarrow$ & DISTS$\downarrow$
  & $D_{\lambda}\downarrow$ & $D_s\downarrow$ & HQNR$\uparrow$ \\
\midrule
-- & LRMS$\uparrow$ (interpolated)~\cite{aiazzi2002context}
  & 32.08 & 0.8239 & 1.82 & 2.37 & 0.3183 & 0.8125 & 0.4409 & 0.2764
  & 0.0072 & 0.0885 & 0.905 \\
\midrule
\multirow{4}{*}{CS} & BT-H~\cite{lolli2017haze}
  & 35.80 & 0.9239 & 1.56 & 1.57 & 0.7232 & 0.9176 & \underline{0.0868} & 0.1163
  & 0.0232 & 0.1481 & 0.832 \\
& BDSD-PC~\cite{vivone2019bdsdpc}
  & 35.10 & 0.9059 & 1.68 & 1.67 & 0.6918 & 0.8924 & 0.1123 & 0.1243
  & 0.0200 & 0.1540 & 0.829 \\
& GSA~\cite{aiazzi2007gsa}
  & 34.76 & 0.8961 & 1.72 & 1.72 & 0.6902 & 0.8864 & 0.1267 & 0.1320
  & 0.0288 & 0.1846 & 0.792 \\
& PRACS~\cite{choi2011pracs}
  & 35.32 & 0.9132 & 1.69 & 1.63 & 0.6987 & 0.9046 & 0.1257 & 0.1310
  & 0.0151 & 0.1221 & 0.865 \\
\midrule
\multirow{4}{*}{MRA} & AWLP~\cite{otazu2005awlp}
  & 34.45 & 0.8846 & 1.94 & 1.64 & 0.6903 & 0.8731 & 0.1358 & 0.1491
  & 0.0092 & 0.1352 & 0.857 \\
& MTF-GLP-FS~\cite{vivone2018fullscale}
  & 35.43 & 0.9021 & 1.66 & 1.59 & 0.6915 & 0.8968 & 0.1253 & 0.1317
  & 0.0056 & 0.1419 & 0.853 \\
& MTF-GLP-HPM~\cite{aiazzi2006mtf,alparone2017intersensor}
  & 34.71 & 0.8893 & 1.65 & 1.72 & 0.6979 & 0.8846 & 0.1221 & 0.1351
  & 0.0071 & 0.1527 & 0.841 \\
& MF~\cite{restaino2016morphological}
  & 34.71 & 0.9018 & 1.64 & 1.70 & 0.7053 & 0.8900 & 0.1140 & 0.1322
  & 0.0154 & 0.1575 & 0.830 \\
\midrule
\multirow{4}{*}{VO} & TV~\cite{palsson2014tv}
  & 35.04 & 0.9149 & 1.91 & 1.74 & 0.6979 & 0.9072 & 0.0937 & \underline{0.1154}
  & 0.0096 & 0.1107 & 0.881 \\
& FE-HPM~\cite{vivone2015semiblind}
  & 35.39 & 0.9063 & 1.61 & 1.58 & 0.7059 & 0.8988 & 0.1107 & 0.1289
  & 0.0134 & 0.1666 & 0.822 \\
& PWMBF~\cite{palsson2015pwmbf}
  & 33.87 & 0.8966 & 1.96 & 1.83 & 0.6814 & 0.8631 & 0.1627 & 0.1622
  & 0.0196 & 0.1015 & 0.881 \\
& LRTCFPan~\cite{wu2023lrtcfpan}
  & 37.53 & 0.9335 & 1.30 & 1.27 & 0.6901 & 0.9351 & 0.0905 & 0.1213
  & \underline{0.0027} & 0.0906 & 0.907 \\
\midrule
\rowcolor{BlueGreen!5} & \textbf{PanBase} ($\mathbf{s}_{\mathrm{pan}}$)
  & \underline{38.31} & \underline{0.9446} & \underline{1.21} & \underline{1.17} & \underline{0.7352} & \underline{0.9503} & 0.0898 & 0.1168
  & \textbf{0.0000} & \underline{0.0350} & \underline{0.965} \\
\rowcolor{BlueGreen!5}\multirow{-2}{*}{\textbf{Ours}} & \textbf{PanBase} ($\mathbf{s}_{\mathrm{ms}}$)
  & \textbf{39.86} & \textbf{0.9584} & \textbf{1.09} & \textbf{0.94} & \textbf{0.8232} & \textbf{0.9605} & \textbf{0.0660} & \textbf{0.1000}
  & \textbf{0.0000} & \textbf{0.0294} & \textbf{0.971} \\
\bottomrule
\end{tabular}}
\end{table*}

\begin{table*}[t]
\small
\centering
\caption{\textbf{Classical pan-sharpening versus standalone PanBase on unseen WV2.}
All methods operate without neural networks. PanBase coefficients are calibrated once using HRMS training pairs.
CS/MRA/VO denote component substitution, multiresolution analysis, and variational optimization.
PanBase reuses WV3 coefficients with the WV2 sensor operator.
Bold and underline indicate the best and second-best values per metric.}
\vspace{-0.2cm}
\label{tab:h_classical_wv2}
\setlength{\tabcolsep}{6pt}
\renewcommand{\arraystretch}{1.1}
\resizebox{\textwidth}{!}{%
\begin{tabular}{cl cccccccc ccc}
\toprule
\multirow{2}{*}{\textbf{Type}}
& \multirow{2}{*}{\textbf{Method}}
& \multicolumn{8}{c}{\textbf{Reduced resolution (RR)}}
& \multicolumn{3}{c}{\textbf{Full resolution (FR)}} \\
\cmidrule(lr){3-10}\cmidrule(l){11-13}
& & PSNR$\uparrow$ & SSIM$\uparrow$ & SAM$\downarrow$ & ERGAS$\downarrow$
  & SCC$\uparrow$ & $Q2^{n}\uparrow$ & LPIPS$\downarrow$ & DISTS$\downarrow$
  & $D_{\lambda}\downarrow$ & $D_s\downarrow$ & HQNR$\uparrow$ \\
\midrule
-- & LRMS$\uparrow$ (interpolated)~\cite{aiazzi2002context}
  & 25.38 & 0.6160 & 6.53 & 6.77 & 0.1582 & 0.6465 & 0.6558 & 0.3293
  & 0.0236 & 0.0559 & 0.922 \\
\midrule
\multirow{4}{*}{CS} & BT-H~\cite{lolli2017haze}
  & 28.90 & 0.8319 & 5.90 & 4.41 & \underline{0.7322} & 0.8364 & 0.2284 & 0.1803
  & 0.0379 & 0.0866 & 0.879 \\
& BDSD-PC~\cite{vivone2019bdsdpc}
  & 29.04 & 0.8399 & 6.14 & 4.26 & 0.7261 & 0.8520 & 0.2016 & 0.1577
  & 0.0760 & 0.0396 & 0.888 \\
& GSA~\cite{aiazzi2007gsa}
  & 28.51 & 0.8159 & 6.25 & 4.50 & 0.7180 & 0.8358 & 0.2214 & 0.1697
  & 0.0376 & 0.0887 & 0.877 \\
& PRACS~\cite{choi2011pracs}
  & 27.24 & 0.7554 & 6.32 & 5.31 & 0.6860 & 0.7756 & 0.2804 & 0.2039
  & 0.0264 & \underline{0.0332} & 0.941 \\
\midrule
\multirow{4}{*}{MRA} & AWLP~\cite{otazu2005awlp}
  & 28.87 & 0.8305 & 6.07 & 4.44 & 0.7272 & 0.8390 & 0.2361 & 0.1872
  & 0.0160 & 0.0686 & 0.917 \\
& MTF-GLP-FS~\cite{vivone2018fullscale}
  & 28.59 & 0.8185 & 6.19 & 4.46 & 0.7183 & 0.8379 & 0.2249 & 0.1722
  & 0.0144 & 0.0762 & 0.911 \\
& MTF-GLP-HPM~\cite{aiazzi2006mtf,alparone2017intersensor}
  & 29.10 & 0.8409 & 6.06 & 4.23 & 0.7287 & 0.8507 & 0.2122 & 0.1688
  & 0.0181 & 0.0795 & 0.904 \\
& MF~\cite{restaino2016morphological}
  & 28.41 & 0.8189 & 6.13 & 4.61 & 0.7158 & 0.8333 & 0.2508 & 0.1948
  & 0.0228 & 0.0812 & 0.898 \\
\midrule
\multirow{4}{*}{VO} & TV~\cite{palsson2014tv}
  & 28.25 & 0.8150 & 6.03 & 4.71 & 0.7082 & 0.8260 & 0.2070 & 0.1561
  & 0.0237 & \textbf{0.0225} & \underline{0.954} \\
& FE-HPM~\cite{vivone2015semiblind}
  & 28.71 & 0.8234 & 6.10 & 4.43 & 0.7246 & 0.8382 & 0.2361 & 0.1841
  & 0.0189 & 0.0811 & 0.902 \\
& PWMBF~\cite{palsson2015pwmbf}
  & 27.60 & 0.7980 & 7.08 & 4.98 & 0.7200 & 0.8052 & 0.2355 & 0.1814
  & 0.0455 & 0.1220 & 0.839 \\
& LRTCFPan~\cite{wu2023lrtcfpan}
  & 29.38 & 0.8436 & 5.62 & 4.12 & 0.6983 & 0.8565 & 0.2293 & 0.1603
  & \underline{0.0075} & 0.0575 & 0.935 \\
\midrule
\rowcolor{BlueGreen!5} & \textbf{PanBase} ($\mathbf{s}_{\mathrm{pan}}$)
  & \underline{30.12} & \underline{0.8691} & \underline{5.49} & \underline{3.77} & 0.7319 & \underline{0.8741} & \underline{0.1323} & \underline{0.1263}
  & \textbf{0.0005} & 0.0404 & \textbf{0.959} \\
\rowcolor{BlueGreen!5}\multirow{-2}{*}{\textbf{Ours}} & \textbf{PanBase} ($\mathbf{s}_{\mathrm{ms}}$)
  & \textbf{30.34} & \textbf{0.8747} & \textbf{5.48} & \textbf{3.65} & \textbf{0.7503} & \textbf{0.8784} & \textbf{0.1203} & \textbf{0.1212}
  & \textbf{0.0005} & 0.0517 & 0.948 \\
\bottomrule
\end{tabular}}
\end{table*}

\begin{table*}[!t]
\small
\centering
\sethlcolor{BlueGreen!5}
\caption{\textbf{Ablation of PanBase construction in MS geometry.}
Here $\mathbf{d}$ denotes the PAN detail after alignment to the MS
geometry by the input-estimated translation $\tau$.
Network-free bases add components cumulatively, with coefficients
recalibrated per row and projection applied after detail injection.
Coeff.\ counts calibrated scalars for $C$ MS bands.
The \hl{highlighted row} is the complete $\mathbf{s}_{\mathrm{ms}}$.
PSNR/SAM use RR and HQNR uses FR.
WV2 reuses WV3 coefficients with its own sensor operator.}
\vspace{-0.2cm}
\label{tab:h_construction_ms}
\setlength{\tabcolsep}{3pt}
\renewcommand{\arraystretch}{1.1}
\resizebox{\textwidth}{!}{%
\begin{tabular}{l c ccc ccc ccc ccc}
\toprule
\multirow{2}{*}{\textbf{Configuration}}
& \multirow{2}{*}{\textbf{Coeff.}}
& \multicolumn{3}{c}{\textbf{WV3}}
& \multicolumn{3}{c}{\textbf{QB}}
& \multicolumn{3}{c}{\textbf{GF2}}
& \multicolumn{3}{c}{\textbf{WV2} (Unseen)} \\
\cmidrule(lr){3-5}
\cmidrule(lr){6-8}
\cmidrule(lr){9-11}
\cmidrule(l){12-14}
& & PSNR$\uparrow$ & SAM$\downarrow$ & HQNR$\uparrow$
  & PSNR$\uparrow$ & SAM$\downarrow$ & HQNR$\uparrow$
  & PSNR$\uparrow$ & SAM$\downarrow$ & HQNR$\uparrow$
  & PSNR$\uparrow$ & SAM$\downarrow$ & HQNR$\uparrow$ \\
\midrule

Interpolated LRMS ($\mathbf{x}_{\mathrm{ms},b}^{\mathrm{lr}}$)
& 0
& 27.12 & 5.80 & 0.909
& 28.06 & 8.44 & 0.831
& 32.08 & 1.82 & 0.905
& 25.38 & 6.53 & 0.922 \\
\midrule

\rowcolor{gray!5}\multicolumn{14}{l}{\textit{MS geometry ($\mathbf{d}$ shifted by $\tau$)}} \\
+ Projection ($\Pi_b(\cdot)$)
& 0
& 27.99 & 5.44 & 0.927
& 29.09 & 7.86 & 0.822
& 34.21 & 1.47 & 0.951
& 26.20 & 6.05 & 0.901 \\

+ Additive detail ($\mathbf{r}_b^1=\mathbf{d}$)
& $C$
& 32.95 & 5.21 & 0.923
& 35.30 & 6.97 & 0.710
& 39.01 & 1.20 & 0.970
& 29.65 & 5.62 & 0.949 \\

+ Multiplicative detail
($\mathbf{r}_b^2=\mathbf{m}_b\odot\mathbf{d}$)
& $2C$
& 33.55 & 4.86 & 0.918
& 35.83 & 6.20 & 0.705
& 39.27 & 1.12 & 0.969
& 30.14 & 5.56 & 0.949 \\

+ Local regression
($\mathbf{r}_b^3=\boldsymbol{\beta}_b\odot\mathbf{d}$)
& $3C$
& 33.81 & 4.63 & 0.916
& 35.94 & 6.11 & 0.704
& 39.59 & 1.10 & 0.967
& 30.30 & 5.52 & 0.947 \\

\rowcolor{BlueGreen!5}
+ Frequency bands
($\mathbf{r}_b^{i,j}=F_j(\mathbf{r}_b^i)$)
& $9C$
& 33.88 & 4.62 & 0.915
& 36.05 & 6.11 & 0.703
& 39.86 & 1.09 & 0.971
& 30.34 & 5.48 & 0.948 \\

\bottomrule
\end{tabular}}
\end{table*}

\begin{table*}[t]
\small
\centering
\caption{\textbf{Ablation of PanBase integration in MS geometry.}
With $\mathbf{s}_{\mathrm{ms}}$, we enable skip replacement,
input concatenation, or both under matched training.
Original retains the interpolated-LRMS skip and original inputs.
Mean change averages the six backbone--dataset pairs.}
\vspace{-0.2cm}
\label{tab:h_skip_input_ms}
\setlength{\tabcolsep}{7pt}
\renewcommand{\arraystretch}{1.1}
\resizebox{\textwidth}{!}{%
\begin{tabular}{ll cccc cccc}
\toprule
\multirow{2}{*}{\textbf{Backbone}}
& \multirow{2}{*}{\textbf{Dataset}}
& \multicolumn{4}{c}{\textbf{RR PSNR$\uparrow$ (dB)}}
& \multicolumn{4}{c}{\textbf{FR HQNR$\uparrow$}} \\
\cmidrule(lr){3-6}\cmidrule(l){7-10}
& & Original & Skip only & Input only & \cellcolor{BlueGreen!5}\textbf{Both (PanBase)}
  & Original & Skip only & Input only & \cellcolor{BlueGreen!5}\textbf{Both (PanBase)} \\
\midrule

\multirow{3}{*}{PanNet~\cite{yang2017pannet}}
& WV3
& 34.29
& 33.95\,{\scriptsize \hneg{$-$0.34}}
& 35.12\,{\scriptsize \hpos{+0.83}}
& 35.22\,{\scriptsize \hpos{+0.93}}
& 0.934
& 0.919\,{\scriptsize \hneg{$-$0.015}}
& 0.963\,{\scriptsize \hpos{+0.029}}
& 0.967\,{\scriptsize \hpos{+0.034}} \\

& QB
& 33.83
& 35.79\,{\scriptsize \hpos{+1.96}}
& 36.30\,{\scriptsize \hpos{+2.47}}
& 36.41\,{\scriptsize \hpos{+2.57}}
& 0.843
& 0.747\,{\scriptsize \hneg{$-$0.096}}
& 0.918\,{\scriptsize \hpos{+0.075}}
& 0.881\,{\scriptsize \hpos{+0.038}} \\

& GF2
& 36.87
& 40.03\,{\scriptsize \hpos{+3.15}}
& 39.79\,{\scriptsize \hpos{+2.92}}
& 40.06\,{\scriptsize \hpos{+3.18}}
& 0.915
& 0.970\,{\scriptsize \hpos{+0.055}}
& 0.966\,{\scriptsize \hpos{+0.051}}
& 0.963\,{\scriptsize \hpos{+0.047}} \\

\midrule
\multirow{3}{*}{PINO~\cite{liu2025pino}}
& WV3
& 36.42
& 37.35\,{\scriptsize \hpos{+0.93}}
& 36.72\,{\scriptsize \hpos{+0.30}}
& 37.55\,{\scriptsize \hpos{+1.13}}
& 0.959
& 0.961\,{\scriptsize \hpos{+0.003}}
& 0.964\,{\scriptsize \hpos{+0.005}}
& 0.954\,{\scriptsize \hneg{$-$0.005}} \\

& QB
& 37.37
& 38.18\,{\scriptsize \hpos{+0.82}}
& 37.96\,{\scriptsize \hpos{+0.59}}
& 38.03\,{\scriptsize \hpos{+0.66}}
& 0.925
& 0.936\,{\scriptsize \hpos{+0.010}}
& 0.934\,{\scriptsize \hpos{+0.009}}
& 0.945\,{\scriptsize \hpos{+0.020}} \\

& GF2
& 42.48
& 44.51\,{\scriptsize \hpos{+2.04}}
& 42.88\,{\scriptsize \hpos{+0.40}}
& 44.66\,{\scriptsize \hpos{+2.19}}
& 0.979
& 0.984\,{\scriptsize \hpos{+0.005}}
& 0.986\,{\scriptsize \hpos{+0.007}}
& 0.984\,{\scriptsize \hpos{+0.004}} \\
\midrule

\multicolumn{2}{l}{Mean change}
& -- & \hpos{+1.43} & \hpos{+1.25} & \hpos{+1.78} & -- & \hneg{$-$0.006} & \hpos{+0.029} & \hpos{+0.023} \\
\bottomrule
\end{tabular}}
\end{table*}

\begin{table*}[!t]
\small
\centering
\caption{\textbf{Ablation of alternative residual bases.}
Alternative bases share the same skip replacement,
input concatenation, and training.}
\vspace{-0.2cm}
\label{tab:h_classical_base_ms}
\setlength{\tabcolsep}{6pt}
\renewcommand{\arraystretch}{1.1}
\resizebox{\textwidth}{!}{%
\begin{tabular}{l cc cc cc cc}
\toprule
\multirow{2}{*}{\textbf{Base}}& \multicolumn{2}{c}{\textbf{PanNet}~\cite{yang2017pannet}, \textbf{WV3}}
& \multicolumn{2}{c}{\textbf{PanNet}~\cite{yang2017pannet}, \textbf{GF2}}
& \multicolumn{2}{c}{\textbf{LAGConv}~\cite{jin2022lagconv}, \textbf{WV3}}
& \multicolumn{2}{c}{\textbf{LAGConv}~\cite{jin2022lagconv}, \textbf{GF2}} \\
\cmidrule(lr){2-3}
\cmidrule(lr){4-5}
\cmidrule(lr){6-7}
\cmidrule(l){8-9}
& PSNR$\uparrow$ & HQNR$\uparrow$
& PSNR$\uparrow$ & HQNR$\uparrow$
& PSNR$\uparrow$ & HQNR$\uparrow$
& PSNR$\uparrow$ & HQNR$\uparrow$ \\
\midrule

LRMS$\uparrow$ (Original)
& 34.29 & 0.934
& 36.87 & 0.915
& 37.22 & 0.944
& 42.47 & 0.915 \\
\midrule

$\Pi$(LRMS$\uparrow$)
& 34.75\,{\scriptsize \hpos{+0.46}}
& 0.956\,{\scriptsize \hpos{+0.022}}
& 37.62\,{\scriptsize \hpos{+0.74}}
& 0.915\,{\scriptsize \hnul{-0.000}}
& \textbf{37.35}\,{\scriptsize \hpos{+0.13}}
& 0.956\,{\scriptsize \hpos{+0.012}}
& 42.90\,{\scriptsize \hpos{+0.43}}
& 0.918\,{\scriptsize \hpos{+0.004}} \\

GSA~\cite{aiazzi2007gsa} (CS)
& 34.30\,{\scriptsize \hpos{+0.01}}
& 0.915\,{\scriptsize \hneg{$-$0.019}}
& 35.69\,{\scriptsize \hneg{$-$1.19}}
& 0.839\,{\scriptsize \hneg{$-$0.076}}
& 37.14\,{\scriptsize \hneg{$-$0.07}}
& 0.955\,{\scriptsize \hpos{+0.011}}
& 42.01\,{\scriptsize \hneg{$-$0.46}}
& 0.908\,{\scriptsize \hneg{$-$0.006}} \\

MTF-GLP-HPM~\cite{aiazzi2006mtf,alparone2017intersensor} (MRA)
& 34.81\,{\scriptsize \hpos{+0.52}}
& 0.931\,{\scriptsize \hneg{$-$0.003}}
& 36.92\,{\scriptsize \hpos{+0.05}}
& 0.885\,{\scriptsize \hneg{$-$0.031}}
& 37.17\,{\scriptsize \hneg{$-$0.05}}
& 0.944\,{\scriptsize \hnul{-0.000}}
& 42.36\,{\scriptsize \hneg{$-$0.10}}
& 0.921\,{\scriptsize \hpos{+0.007}} \\

LRTCFPan~\cite{wu2023lrtcfpan} (VO)
& 34.81\,{\scriptsize \hpos{+0.52}}
& 0.942\,{\scriptsize \hpos{+0.009}}
& 37.80\,{\scriptsize \hpos{+0.93}}
& 0.904\,{\scriptsize \hneg{$-$0.011}}
& 37.24\,{\scriptsize \hpos{+0.02}}
& 0.956\,{\scriptsize \hpos{+0.013}}
& 42.24\,{\scriptsize \hneg{$-$0.23}}
& 0.948\,{\scriptsize \hpos{+0.034}} \\

\rowcolor{BlueGreen!5}
\textbf{PanBase} ($\mathbf{s}_{\mathrm{ms}}$)
& \textbf{35.22}\,{\scriptsize \hpos{+0.93}}
& \textbf{0.967}\,{\scriptsize \hpos{+0.034}}
& \textbf{40.06}\,{\scriptsize \hpos{+3.18}}
& \textbf{0.963}\,{\scriptsize \hpos{+0.047}}
& 37.29\,{\scriptsize \hpos{+0.07}}
& \textbf{0.962}\,{\scriptsize \hpos{+0.018}}
& \textbf{43.02}\,{\scriptsize \hpos{+0.55}}
& \textbf{0.969}\,{\scriptsize \hpos{+0.055}} \\

\bottomrule
\end{tabular}}
\end{table*}

\clearpage

{
    \small
    \bibliographystyle{ieeenat_fullname}
    \bibliography{main}
}

\end{document}